%% file: main.tex
\documentclass{article} 

\usepackage{macros}
\usepackage{arxiv}

\newtoggle{arxiv}
\toggletrue{arxiv}

\usepackage{csquotes} 
\usepackage{url}            
\usepackage{graphicx}
\usepackage[numbers,sort]{natbib}
\usepackage{subfiles}
\usepackage{tabularx}
\usepackage{etoolbox}
\usepackage{multirow}
\usepackage{enumitem}
\usepackage{amssymb}
\usepackage{amsmath}
\usepackage{amsthm}
\usepackage{xspace}
\usepackage{float}
\usepackage{wrapfig}
\usepackage{booktabs}       
\usepackage{amsfonts}       
\usepackage{nicefrac}       
\usepackage{microtype}      
\usepackage{color, soul}
\usepackage{pifont}
\usepackage{nth}
\usepackage{enumitem}
\usepackage{wrapfig}
\usepackage{xcolor}
\usepackage{subcaption}

\usepackage[framemethod=TikZ]{mdframed}

\usepackage{listings}
\definecolor{codegreen}{rgb}{0,0.6,0}
\definecolor{codegray}{rgb}{0.5,0.5,0.5}
\definecolor{codepurple}{rgb}{0.58,0,0.82}
\definecolor{backcolour}{rgb}{1,1,1}

\usepackage{listings}  
\usepackage{algorithm}
\algrenewcommand\algorithmicrequire{\textbf{Input:}}
\algrenewcommand\algorithmicensure{\textbf{Output:}}

\iftoggle{arxiv}{
    \setlength{\textwidth}{6.5in}
    \setlength{\textheight}{9in}
    \setlength{\oddsidemargin}{0in}
    \setlength{\evensidemargin}{0in}
    \setlength{\topmargin}{-0.5in}
    \newlength{\defbaselineskip}
    \setlength{\defbaselineskip}{\baselineskip}
    \setlength{\marginparwidth}{0.8in}
}

\newcolumntype{Y}{>{\hsize=.7\hsize}X}
\newcolumntype{Z}{>{\hsize=1.3\hsize}X}

\makeatletter
\def\@copyrightspace{\relax}
\makeatother

\makeatletter
\def\@myauthornotes{}
\def\myauthornote#1{%
  \if@ACM@anonymous\else
    \g@addto@macro\addresses{}%
    \g@addto@macro\@myauthornotes{%
      \stepcounter{footnote}\footnotetext{#1}}%
  \fi}
\makeatother

\iftoggle{arxiv}{
    \title{
        \name: Fast and \textit{Furious} AMD Kernels
    }
    \usepackage{authblk}
    \author[$\triangle$]{William Hu}
    \author[$\triangle$]{Drew Wadsworth}
    \author[$\dagger$]{Sean Siddens}
    \author[$\dagger$]{Stanley Winata}
    \author[$\ddagger$]{Daniel Y. Fu}
    \author[$\dagger$]{Ryan Swann}
    \author[$\dagger$]{Muhammad Osama}
    \author[$\triangle$]{Christopher Ré}
    \author[$\triangle$]{Simran Arora}
    \affil[$\triangle$]{Stanford University}\vspace{4pt}
    \affil[$\dagger$]{Advanced Micro Devices, Inc.}\vspace{4pt}
    \affil[$\ddagger$]{University of California, San Diego}\vspace{4pt}
    \affil[]{\{\texttt{willhu, simarora\}@stanford.edu}}
}

\definecolor{dkgreen}{rgb}{0,0.6,0}
\definecolor{gray}{rgb}{0.5,0.5,0.5}
\definecolor{light-gray}{gray}{0.95}
\definecolor{mauve}{rgb}{0.58,0,0.82}
\definecolor{backcolour}{rgb}{0.95,0.95,0.92}

\usepackage{graphicx}
\usepackage{subcaption}
\usepackage{wrapfig,lipsum,booktabs}
\usepackage{threeparttable}
\usepackage{makecell}
\usepackage{url}
\usepackage{algorithm}
\usepackage{algpseudocode}
\usepackage{cleveref}
\usepackage{changepage}
\usepackage{booktabs}

\usepackage{epstopdf}
\usepackage{graphicx}
\usepackage{textcomp}       
\usepackage{scalerel}       
\newcommand{\name}{\textsc{HipKittens}}
\newcommand{\shortname}{\textsc{HK}}
\newcommand{\tk}{\textsc{ThunderKittens}}
\newcommand{\shorttk}{\textsc{TK}}
\usepackage{hyperref}

\usepackage{listings}
\usepackage{color}
\definecolor{dkgreen}{rgb}{0,0.6,0}
\definecolor{gray}{rgb}{0.5,0.5,0.5}
\definecolor{mauve}{rgb}{0.58,0,0.82}
\lstdefinelanguage{CUDACPP}{
  language=C++,
  morekeywords={__global__, __host__, __device__, __shared__, blockIdx, blockDim, threadIdx, gridDim},
  morecomment=[l][\color{magenta}]{\#},
}
\lstdefinestyle{pythonstyle}{
  language=Python,
  frame=tb,
  aboveskip=3mm,
  belowskip=3mm,
  showstringspaces=false,
  columns=flexible,
  basicstyle={\ttfamily\footnotesize},
  numbers=none,
  numberstyle=\footnotesize\color{gray},
  keywordstyle=\color[rgb]{0.13,0.29,0.53},
  commentstyle=\color[rgb]{0.13,0.55,0.13},
  stringstyle=\color[rgb]{0.31,0.60,0.02},
  breaklines=true,
  breakatwhitespace=true,
  tabsize=4
}

\input{math_commands.tex}

\usepackage{enumitem}

\newcommand{\ShowNotes}{}
\input{macros}

\begin{document}

\maketitle

\input{sections/0-abstract}

\input{sections/1-introduction}
\input{sections/2-background}
\input{sections/3-hip-kittens}

\input{sections/4-experiments}

\input{sections/5-conclusion}

\section{Acknowledgements}
We thank the Hazy Research Lab and Stanford AI Lab for providing feedback on this work. 
We gratefully acknowledge the support of NIH under No. U54EB020405 (Mobilize), NSF under Nos. CCF2247015 (Hardware-Aware), CCF1763315 (Beyond Sparsity), CCF1563078 (Volume to Velocity), and 1937301 (RTML); US DEVCOM ARL under Nos. W911NF-23-2-0184 (Long-context) and W911NF-21-2-0251 (Interactive Human-AI Teaming); ONR under Nos. N000142312633 (Deep Signal Processing); Stanford HAI under No. 247183; NXP, Xilinx, LETI-CEA, Intel, IBM, Microsoft, NEC, Toshiba, TSMC, ARM, Hitachi, BASF, Accenture, Ericsson, Qualcomm, Analog Devices, Google Cloud, Salesforce, Total, the HAI-GCP Cloud Credits for Research program,  the Stanford Data Science Initiative (SDSI), and members of the Stanford DAWN project: Meta, Google, and VMWare. The U.S. Government is authorized to reproduce and distribute reprints for Governmental purposes notwithstanding any copyright notation thereon. Any opinions, findings, and conclusions or recommendations expressed in this material are those of the authors and do not necessarily reflect the views, policies, or endorsements, either expressed or implied, of NIH, ONR, or the U.S. Government. 

\section{Contributions} WH, DW, and SA designed and implemented {\name}, kernels, micro experiments, and baselines. SS designed the cache reuse strategy. WH and SA wrote the manuscript. SW, DF, RS, MO, CR advised and SA supervised the project.

\bibliography{references}
\bibliographystyle{references}

\clearpage
The appendix is organized as follows:
\begin{enumerate}
    \item Appendix~\ref{app:extended_related_work} contains a discussion of extended related work.
    \item Appendix~\ref{app:extended_results} provides extended results and ablations.
    \item Appendix~\ref{app:implementation} discusses library implementation details.
    \item Appendix~\ref{app:kernel_listings} provides sample kernel listings written using {\shortname}.
    \item Appendix~\ref{app:case_study} provides a case study of FP6 kernels on AMD.
\end{enumerate}
\input{appendix/extended_related_work}

\input{appendix/evaluations}

\input{appendix/implementations}

\input{appendix/kernels}

\clearpage

\input{appendix/case_study}

\end{document}

%% file: math_commands.tex
\usepackage{amsmath,amsfonts,bm}

\def\eqref#1{equation~\ref{#1}}

\def\1{\bm{1}}

\DeclareMathAlphabet{\mathsfit}{\encodingdefault}{\sfdefault}{m}{sl}
\SetMathAlphabet{\mathsfit}{bold}{\encodingdefault}{\sfdefault}{bx}{n}



%% file: macros.tex
\ifdefined\ShowNotes
  \newcommand{\colornote}[3]{{\color{#1}\bf{#2 #3}\normalfont}}
\else
  \newcommand{\colornote}[3]{}
\fi

\definecolor{darkred}{rgb}{0.7,0.1,0.1}
\definecolor{darkgreen}{rgb}{0.1,0.5,0.1}
\definecolor{cyan}{rgb}{0.7,0.0,0.7}
\definecolor{dblue}{rgb}{0.2,0.2,0.8}
\definecolor{maroon}{rgb}{0.76,.13,.28}
\definecolor{burntorange}{rgb}{0.81,.33,0}
\definecolor{royalpurple}{rgb}{0.47,.31,0.66}

\ifdefined\ShowNotes
  
\else
  
\fi

%% file: sections/0-abstract.tex
\begin{abstract}

AMD GPUs offer state-of-the-art compute and memory bandwidth; however, peak performance AMD kernels are written in raw assembly. To address the difficulty of mapping AI algorithms to hardware, recent work proposes C++ embedded and PyTorch-inspired domain-specific languages like {\tk} ({\shorttk}) to simplify high performance AI kernel development on NVIDIA hardware. 
We explore the extent to which such primitives --- for explicit tile-based programming with optimized memory accesses and fine-grained asynchronous execution across workers --- are NVIDIA-specific or general. We provide the first detailed study of the programming primitives that lead to performant AMD AI kernels, and we encapsulate these insights in the {\name} (\shortname) programming framework. We find that tile-based abstractions used in prior DSLs generalize to AMD GPUs, however we need to rethink the algorithms that instantiate these abstractions for AMD. 
We validate the {\shortname} primitives across CDNA3 and CDNA4 AMD platforms. In evaluations, {\shortname} kernels compete with AMD's hand-optimized assembly kernels for GEMMs and attention, and consistently outperform compiler baselines. Moreover, assembly is difficult to scale to the breadth of AI workloads; reflecting this, in some settings {\shortname} outperforms all available kernel baselines by $1.2-2.4\times$ (e.g., $d=64$ attention, GQA backwards, memory-bound kernels). These findings help pave the way for a single, tile-based software layer for high-performance AI kernels that translates across GPU vendors. {\name} is released at: \url{https://github.com/HazyResearch/HipKittens}.
\end{abstract}

%% file: sections/1-introduction.tex
\section{Introduction}
\label{sec:introduction}

While AI has largely used a single hardware vendor in the past~\cite{krizhevsky2012imagenet, llama3modelcard, openai2025gptoss120bgptoss20bmodel}, 
AMD GPU hardware now offers state-of-the-art peak compute and memory bandwidth (Table \ref{fig:hardware-overview}). However, the lack of mature software support creates a hardware lottery (``CUDA moat'')~\cite{hooker2020lottery, semianalysis2024moat}. Peak-performance AMD kernels are \textit{written in raw assembly} by a handful of experts (i.e., the AITER library ~\cite{amd2025aiter}), which is slow to scale to the breadth of AI workloads. For instance, AITER and PyTorch Llama GQA backwards achieve just 30\% and 24\% of SoTA performance respectively on AMD MI355X GPUs (Section~\ref{sec:sec4_results}).

\vspace{1mm}
Developing NVIDIA kernels also required painstaking effort a few years ago. For instance, using low level CUDA/CUTLASS, it took \textit{two years} between the H100 GPU's release and the release of peak performance open-source attention kernels~\cite{dao2024flashattention3}. Compilers like Triton~\cite{triton} are simpler to use, but sacrifice performance and struggle to quickly support new hardware features~\cite{spector2025tk, triton2025gluon}. AI-designed kernels are showing early promise~\cite{baronio2025kevinmultiturnrlgenerating, lange2025robustagenticcudakernel}, but current models also struggle to use new hardware features~\cite{ouyang2025kernelbench} and are susceptible to reward hacking ~\cite{baronio2025kevinmultiturnrlgenerating}. Recently, lightweight C++ embedded DSLs like  {\tk} ({\shorttk}) (and successors like CuTe DSL~\cite{nvidia2025cute4} and Gluon~\cite{zhou2025linearlayoutsrobustcode}) consider simplifying development by encoding kernel design in terms of a small set of opinionated primitives that give the developer full control:
\begin{enumerate}[itemsep=2pt,topsep=2pt,leftmargin=*]
    \item \textbf{Tiles.} The basic data type is a \textit{tile} with optimized memory access patterns. {\shorttk} exposes lightweight PyTorch-inspired bulk compute operators (\verb|mma|, \verb|exp|, etc.) over tiles, wrapping PTX. Tiles help developers explicitly manage data at each level of the GPU memory hierarchy.
    \item \textbf{Overlapping.} A few basic kernel patterns help developers achieve high occupancy, or schedule workers (\textit{waves} on AMD, \textit{warps} on NVIDIA) onto the different hardware execution units. Modern NVIDIA kernels have consolidated around wave specialization (producer-consumer) scheduling patterns~\cite{dao2024flashattention3, spector2025tk, spector2025look, 25-comet, wang2025tilelang}.
    \item \textbf{Grid scheduling.} By assigning work to thread blocks in the appropriate order, a developer can maximize the reuse of non-programmable cache memory.
\end{enumerate}

\input{figures/main/title}

\textit{Our work asks whether entirely new programming primitives are needed to simplify AMD kernel development, or whether existing primitives suffice.} Ideally, we would have a simple framework that helps developers write a breadth of high performance kernels.
Our exploration led to {\name} ({\shortname}), a minimal collection of C++ embedded programming primitives for AMD:

\begin{enumerate}[itemsep=2pt,topsep=2pt,leftmargin=*]
    \item \textbf{Optimized access patterns for programmable GPU memory.} 
    Careful register memory management is critical for peak performance kernels. {\shortname} retains the tile data structures of prior DSLs to help developers manage memory \cite{spector2025tk}. However, optimizing the tiles for AMD introduces new challenges.

    Compilers such as Triton and HIPCC routinely impact the kernel developer's ability to carefully schedule register allocations and lifetimes (Sec.~\ref{sec:hip-warp}). For instance, HIPCC prevents the HIP developer from using certain types of registers (AGPRs) as input operands to matrix instructions.\footnote{AMD CDNA includes 512 registers per SIMD, which are evenly partitioned across waves co-located on the SIMD. For kernels with a single SIMD per wave, the hardware splits the registers into 256 vector general-purpose registers (VGPRs) and 256 accumulator registers (AGPRs).} We thus introduce a way for the developer to \textit{bypass} the compiler altogether, and explicitly pin the registers belonging to each tile. 

    As for memory access patterns, different NVIDIA matrix instruction shapes are all built from the same underlying \textit{core matrix} structure, which makes it easy to use a single tile swizzle strategy for all shapes as in {\shorttk} and Linear Layouts~\cite{zhou2025linearlayoutsrobustcode}. AMD matrix instructions lack this compositional structure, leading to an explosion in the number of tile layouts. Further, the shared memory bank structure and order in which threads in a wave run, differs based on the memory instruction on AMD (Sec.~\ref{sec:hip-warp}). {\shortname} handles this complexity for the developer when tiles are created.

    \item \textbf{Schedules to overlap compute and memory.} Ideally we would have simple, reusable patterns for scheduling the compute and memory within kernels that generalize across AI workloads. The wave specialization pattern dominates NVIDIA kernels and DSLs; \textit{producer waves} handle memory operations while \textit{consumers} execute bulk compute operations over large tiles. However, we find that this pattern underperforms on AMD CDNA3 and CDNA4 GPUs due to architectural differences---AMD's static register allocation means that producer waves consume registers without contributing to computation, which limits the output tile size that gets computed per thread block and thus the kernel's arithmetic intensity. On the MI355X, wave specialization achieves just 80\% of peak BF16 GEMM performance (Tab.~\ref{tab:producer_consumer_ablations}).

    We identify two alternate scheduling patterns that consistently achieve peak AMD performance: \textsc{8-wave ping-pong} and \textsc{4-wave interleave}, which tradeoff programmability with performance (Tab.~\ref{fig:pingpong}). The \textsc{8-wave} pattern employs two waves per SIMD that alternate between compute and memory roles, each executing large bulk tile operations (Fig.~\ref{fig:main}), while the \textsc{4-wave} pattern employs one wave per SIMD and finely interleaves instructions over small tile sizes. Remarkably, the simple \textsc{8-wave} pattern is sufficient to match AMD's hand-optimized assembly kernels across BF16 GEMM, FP8 GEMM, and attention forward, and outperforms baselines by $1.8\times$ on GQA non-causal backward.

    \item \textbf{Optimized access patterns for non-programmable GPU memory.} Chiplet architectures are becoming the dominant path to GPU scaling---NVIDIA Blackwell uses 2 chips, AMD MI355X uses 8---but existing frameworks ignore their hierarchical cache structures, leaving performance untapped. Each AMD CDNA4 chiplet contains 32 processors with private L2 cache, while all chiplets share a last-level cache (LLC) between L2 and global memory. These hierarchical caches have orthogonal preferences for how work is parallelized across thread blocks. 
    Table~\ref{tab:l2_table_14592} demonstrates an instance where a naive row-major ordering for allocating work to thread blocks achieves only a 36\% L2 hit rate for BF16 GEMMs. 
    We show that optimizing solely for L2 reuse (e.g., to 79\% hit rate) degrades LLC performance and overall bandwidth. 
    {\shortname} introduces an algorithm that models both cache levels when scheduling thread blocks. This improves performance by 19\% over the naive row-major baseline (Tab.~\ref{tab:l2_table_14592}).  
\end{enumerate}
\vspace{-4mm}
\paragraph{Evaluation.} We validate {\name} on AMD CDNA3 MI325X and CDNA4 MI355X GPUs. On the most widely used and optimized workloads in AI, {\shortname} competes with or outperforms all AMD baselines (BF16 FP8 GEMM, GQA/MHA Attention forward and backward, RoPE, LayerNorm). {\shortname} outperforms all available AMD baselines on average, including AMD's \textit{hand-optimized kernels that have been written in raw assembly}. However, assembly is not a scalable method for kernel development and leaves many important AI workloads unsupported --- in such settings (e.g., some attention shapes, GQA backwards, memory bound kernels), {\shortname} outperforms the available AMD baselines by $1.2-10\times$. Furthermore, {\shortname} consistently outperforms compilers approaches (e.g., by up to $3\times$ the Triton BF16 GEMM and $2\times$ the Mojo MHA forwards).

We contribute (1) principles for writing high performance AMD kernels, (2) {\shortname}, a collection of opinionated C++ programming primitives for the AI community, and (3) a suite of performant AMD kernels. We further show that the tile primitives proposed in the {\shorttk} DSL transfer to AMD, providing evidence that one unified and performant programming model is possible across AI accelerators. Scaling kernel support across \textit{multiple silicon platforms} is key to unlocking the ``compute capacity needed to achieve AI's full potential''~\cite{openai2025amd}. We hope this work helps open AI's hardware landscape.

\input{tables/specs}

%% file: figures/main/title.tex
\begin{figure*}[t]
    \vspace{-4mm}
    \includegraphics[width=\linewidth]{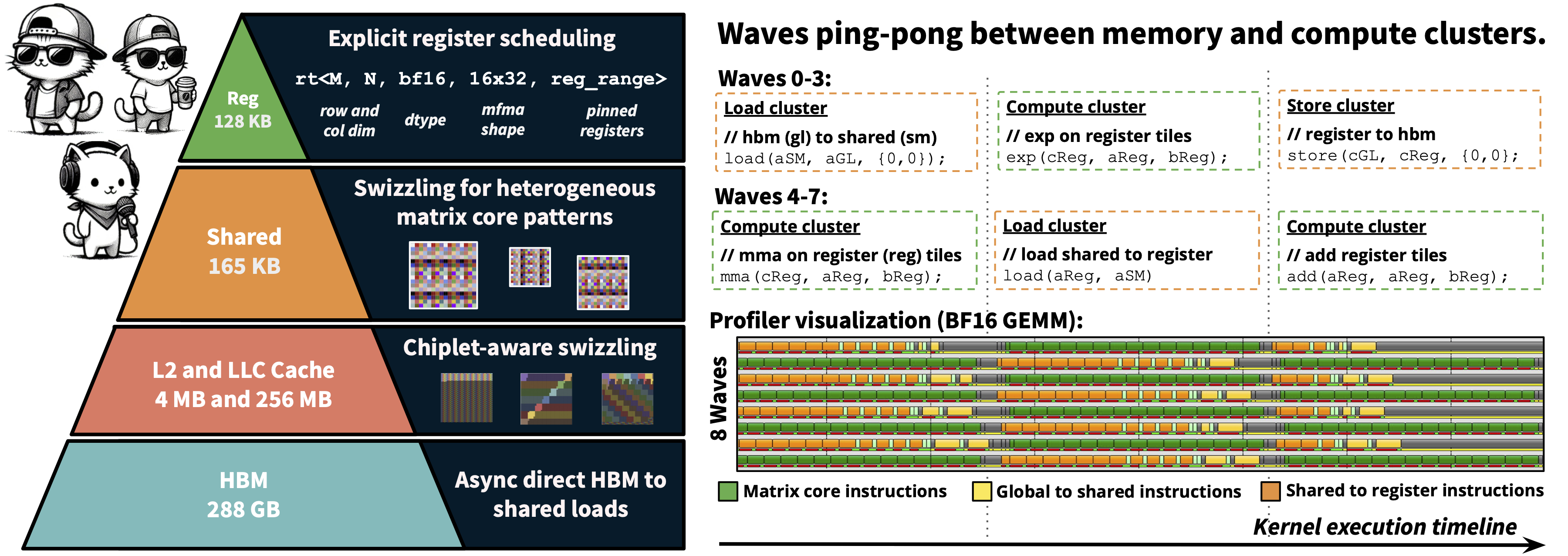}
    \vspace{-4mm}
    \caption{We study whether existing tile based programming primitives suffice for AMD kernels, or whether entirely new primitives are needed. Our study led to {\name}: a minimal and opinionated set of primitives for fast and \textit{furious} AMD kernels. {\shortname} introduces a general  \textsc{8-wave ping-pong schedule} to overlap compute and memory, programmer controlled register allocation, and efficient shared memory and chiplet-aware swizzling algorithms to enable a suite of high performance AMD AI kernels.}
    \label{fig:main}
    \vspace{-4mm}
\end{figure*}

%% file: tables/specs.tex
\begin{figure*}[t!]
\centering
\begin{minipage}[c]{0.64\textwidth}
    \vspace{-4mm}
    \centering
    \small
    \begin{sc}
    \begin{tabular}{@{}lcc@{}}
    \toprule
    Spec &  NVIDIA B200 SXM5 & AMD MI355X OAM \\
    \midrule
    BF16 matrix / tensor & 2.2 PFLOPs & 2.5 PFLOPs \\
    MXFP8 matrix / tensor & 4.5 PFLOPs & 5.0 PFLOPs \\
    MXFP6 matrix / tensor & 4.5 PFLOPs & 10.1 PFLOPs \\
    MXFP4 matrix / tensor & 9.0 PFLOPs & 10.1 PFLOPs \\
    \midrule
    Memory capacity & 180 GB & 288 GB \\
    Memory bandwidth & 8.0 TB/s & 8.0 TB/s \\
    \bottomrule
    \end{tabular}
    \end{sc}
\end{minipage}%
\hfill
\begin{minipage}[c]{0.33\textwidth}
    \centering
    \includegraphics[width=0.95\linewidth]{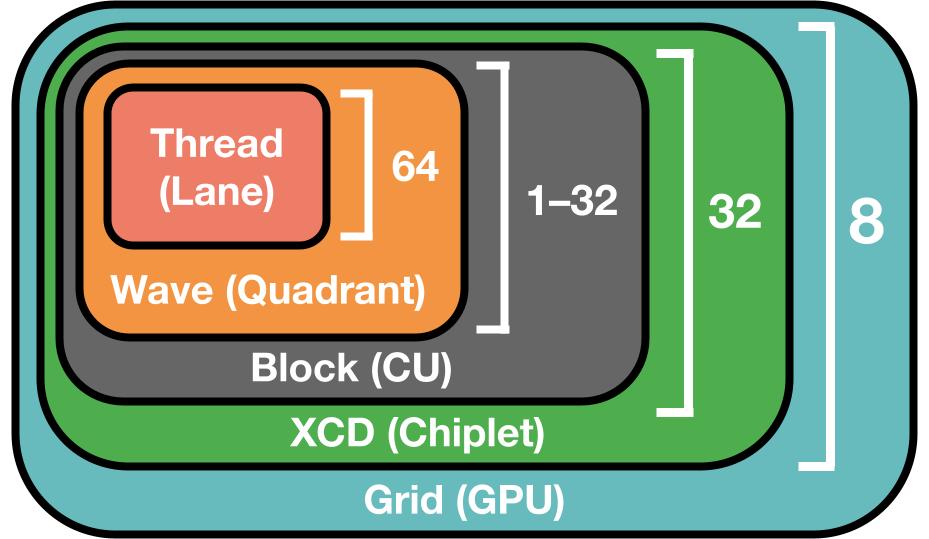}
\end{minipage}
\caption{\textbf{Hardware overview.} (Left) Peak memory and compute speeds for the latest generation GPU platforms~\cite{amd2025mi355x, 25-blackwell}. (Right) Diagram of the AMD GPU software and hardware hierarchy.}
\vspace{-4mm}
\label{fig:hardware-overview}
\end{figure*}

%% file: sections/2-background.tex
\vspace{-2mm}
\section{Background}
\label{sec2_preliminaries}

In this section, we provide background on AMD GPU hardware in Section~\ref{sec2:background} and discuss related work in Section~\ref{sec2:related_work}. Section~\ref{app:extended_related_work} provides an extended discussion of related work. 
\vspace{-2mm}
\subsection{GPU fundamentals}
\label{sec2:background}

GPU kernels are small programs that load data, perform work on it, and write the results back out to memory. We generally adopt AMD terminology in this work and provide a mapping between AMD and NVIDIA terminology in Appendix~\ref{sec:terminology}.
\begin{enumerate}[itemsep=0.1pt,topsep=0pt,leftmargin=*]
    \item \textbf{Compute hierarchy.} Kernels are executed by tens of thousands of threads across hundreds of processors, called ``compute units'' (CUs). CUs organize their hardware resources in 4 ``single instruction, multiple data'' (SIMD) units. Threads are arranged in a hierarchy: threads are the smallest units of execution, ``waves'', groups of 64 threads, execute in lockstep on individual SIMDs, and ``thread blocks'', groups of waves, are jointly scheduled on the CUs. AMD MI355X GPUs contain 256 CUs organized into 8 accelerator complex dies (XCDs) of 32 CUs in a chiplet layout.

    \item \textbf{Memory hierarchy.}  
    Memory is organized in a hierarchy: small amounts of quick-access and large amounts of slow-access memory. A single SIMD contains 512 32-bit vector registers (for a total of 512KB per CU). Each CU has an L1 cache and shared memory that can be  accessed by multiple waves in the same thread block. Each XCD shares a non-programmable 4MB L2 cache. All CUs share a large, slow global memory (HBM) and a last level cache (LLC) sits between L2 and HBM. 

    \item \textbf{Occupancy.} Threads execute instructions on physical execution units (ALU, FMA, matrix cores), which are specialized for different types of compute. Instructions performed by these units each have a fixed issue latency and limited amount of bandwidth. Different waves can occupy different units simultaneously to avoid saturating any single unit. Each unit imposes different constraints on the memory layouts, or the mapping of logical data elements to physical thread ownership~\cite{amd2025matrix}.
\end{enumerate}

\vspace{-2mm}
\paragraph{Software overview.} Developers can write kernels at different levels of the software stack. Raw assembly provides maximal control over register usage and instruction selection and ordering. CUDA / HIP C++ gets compiled (via NVCC, HIPCC) to assembly, and the compiler may introduce its own instruction reordering and register lifetime tracking. LLVM accepts compiler hints,
which let the developer guide the compiler's behavior. Some compilers expose high level interfaces on top of C++ (e.g., Python~\cite{paszke2019pytorchimperativestylehighperformance}, Triton~\cite{triton}).

\vspace{-2mm}
\subsection{Related work}
\label{sec2:related_work}
Currently, peak-performance AMD kernels finely interleave compute and memory instruction issues in raw assembly (see the AITER and Composable Kernel libraries)~\cite{amd2025aiter, amd2025ck}. In contrast, to simplify and accelerate the kernel development process, the AI community recently adopted bulk programming operators over optimized \textit{tile} primitives for kernel development, as proposed in {\tk}~\cite{spector2025tk} \footnote{\url{https://github.com/HazyResearch/ThunderKittens} (May 2024)} and successors (e.g., CuTe DSL~\cite{nvidia2025cute4}\footnote{\url{https://docs.nvidia.com/cutlass/media/docs/pythonDSL/cute_dsl.html} (Sept 2025)}, Gluon~\cite{zhou2025linearlayoutsrobustcode}\footnote{\url{https://github.com/triton-lang/triton/tree/main/python/tutorials/gluon} (June 2025)}). However, these existing C++ based DSLs only run on NVIDIA GPUs, wrapping PTX and CUDA. Compiler libraries such as Triton, TileLang, and Mojo are built on top of LLVM/MLIR~\cite{lattner2020mlir, lattner2020llvm, llvm2025user} and can compile for AMD GPUs. However, these works have neither provided reusable principles or primitives for AMD, nor released comprehensive suites of high performance AMD kernels. For instance, Mojo's MHA kernel suffers from expensive bank conflicts and achieves just 50\% of the peak kernels' performance on the MI355X.\footnote{Measured using \texttt{rocprofv3 --pmc SQ\_LDS\_BANK\_CONFLICT,SQ\_INSTS\_LDS --output-format csv --output-file profiles\_3 -d out -- mojo bench\_mha.mojo} at \url{https://github.com/modular/modular/tree/main/max/kernels/benchmarks/gpu} in the nightly Modular build on a MI355X GPU on 11/06/2025.} 
{\name} provides the first systematic set of AMD AI kernel primitives towards opening up the hardware landscape.

\label{sec:background-brittle}

%% file: sections/3-hip-kittens.tex
\vspace{-2mm}
\section{HipKittens}
\label{sec3_method}
This section describes {\name} (\shortname), a framework for programming AI kernels on AMD GPUs. 
{\shortname} builds on the {\tk} framework~\cite{spector2025tk}, which uses C++ embedded tile-based programming primitives to simplify high-performance and flexible AI kernel development (discussed in \Cref{sec:preliminaries}). We describe {\shortname}'s principles for optimizing the access patterns of programmable GPU memory in~\Cref{sec:hip-warp}, maximizing occupancy in~\Cref{sec:hip-block}, and optimizing the access patterns of non-programmable cache memory in~\Cref{sec:hip-grid}.

\vspace{-2mm}
\subsection{Tile programming interface} 
\label{sec:preliminaries}
Like existing kernel frameworks, {\shortname} adopts tiles as the basic data structure and provides a suite of optimized operators over tiles. The tile design and suite of operators is heavily inspired by PyTorch and NumPy~\cite{paszke2019pytorchimperativestylehighperformance, numpy}, given their familiarity to the AI community. 
\begin{itemize}[itemsep=0.1pt,topsep=0pt,leftmargin=*]
    \item \textbf{Memory.} The developer can initialize tiles in register or shared memory. A tile is parametrized by a \verb|dtype| (FP32, BF16, FP16, FP8, FP6), \verb|rows|, \verb|columns| and a \verb|layout| (row or column major).
    Tile rows and columns are restricted to be a multiple of the matrix core shape. {\shortname} provides operators to \verb|load| and \verb|store| tiles across different levels of the GPU memory hierarchy.
    \item \textbf{Compute.} {\shortname} provides a suite of bulk compute operators over tiles, inspired by the set of operators in PyTorch (e.g., \verb|mma|, \verb|exp|, \verb|add|). The functions are lightweight and do not add overhead as they directly wrap raw AMD CDNA assembly/HIP (NVIDIA PTX/CUDA for {\shorttk}). 
\end{itemize}
\noindent Given these familiar programming primitives,  
{\shortname} automatically optimizes the memory access patterns for tiles. Memory management on AMD GPUs raises key challenges at each level of the hierarchy, discussed next.

\subsection{Optimizing programmable memory access}
\label{sec:hip-warp}

We now discuss the specifics of {\name} tiles.

\vspace{-2mm}
\subsubsection{Developer-controlled register scheduling}
Careful register management is critical for high performance. However, compilers either prevent (e.g., Triton) or impede (e.g., in the HIPCC compiler) the developer's ability to maximally control register allocations. 

For instance, in kernels with a single wave per SIMD, the AMD hardware splits the SIMD's 512 registers into 256 vector general-purpose registers (VGPRs) and 256 accumulator registers (AGPRs). However, while the hardware does support using AGPRs as input to matrix core instructions, HIPCC does not. For workloads that involve both matrix and vector operations (e.g., attention backwards), kernels compiled via HIPCC would need to generate redundant \verb|v_accvgpr_read| instructions that move data from AGPRs to VGPRs prior to issuing matrix instructions.

\input{figures/programmable_memory/mi355x_attn_bkwd}

\vspace{-2mm}
\paragraph {\textsc{Explicit register scheduling}.} 
The compiler constraints motivate a feature in {\shortname} that gives developers the ability to fully control register scheduling. The developer pins the registers belonging to each tile, rather than letting HIPCC manage the registers. By bypassing the compiler, the developer can use AGPRs as inputs to matrix instructions, resulting in our SoTA-level backwards attention kernel (Tab.~\ref{table:attn_bkwd_micro}). The interface for programming with pinned register tiles exactly matches that of using standard compiler-managed register tiles. We leave both options so developers can choose the level of control they want.

\vspace{-2mm}
\subsubsection{Tiles for heterogeneous matrix core shapes}
AI kernels use different matrix core instruction shapes (MxNxK), depending on the workload properties, in order to carefully manage register pressure. However, it is challenging to use multiple shapes on AMD GPUs. 

\input{figures/programmable_memory/matrix_layouts}
\vspace{-2mm}
\paragraph{Matrix layout complexity. } Recall that GPU matrix instructions impose rules as to which thread owns each data element in its registers. Further, shared memory accesses result in bank conflicts if multiple threads in a wave attempt to access the same bank simultaneously. Waves (and NVIDIA warps) execute shared memory accesses in \textit{phases}; i.e., a \textit{subset} of threads in a wave accesses shared memory concurrently~\cite{nvidia2023smemforum}.

The complexity of AMD matrix layouts relative to NVIDIA layouts impacts the access patterns at each level of the GPU memory hierarchy. First, NVIDIA matrix instructions use a \textit{regular} pattern (Fig.~\ref{fig:nvidia-core-matrix}); all shapes are composed from an underlying $16\times16$ \textit{core matrix} building block that is stamped out multiple times depending on the overall matrix instruction shape. Thus, prior frameworks like {\shorttk}~\cite{spector2025tk} and Linear layouts~\cite{zhou2025linearlayoutsrobustcode}, can use a unified swizzling strategy that generalizes across matrix shapes. Meanwhile, each AMD matrix instruction uses an entirely different layout without a similar underlying structure.
Second, NVIDIA instructions sequentially assign threads to phases (e.g., threads 0-7 in phase one, 8-15 in phase two). Meanwhile on AMD, the phases are non-sequential and differ based on the memory instruction.\footnote{For e.g., the threads in a wave execute the \texttt{ds\_read\_b128} instruction in 4 phases and load data from 64 shared memory banks, each 4-bytes wide, while a \texttt{ds\_read\_b96} executes over 8 phases and loads from 32 banks. The phases are undocumented in the CDNA ISA so we create a solver to determine them, and document the phases in Tab.~\ref{tab:phase-bank}.}

\input{figures/programmable_memory/swizzle}

\vspace{-2mm}
\paragraph{Optimized tile memory.} We discuss how {\shortname} abstracts away this complexity from kernel developers: 
\begin{enumerate}[itemsep=2pt,topsep=0pt,leftmargin=*]
    \item \textbf{Register.} By default, register tiles in {\shortname} use the smallest MFMA instruction since this provides maximal scheduling control as highlighted in~\Cref{sec:hip-block}. However, for the edge case kernels that use alternate sizes, {\shortname} lets the developer parameterize the desired register tile by the MFMA instruction shape.
    \item \textbf{Shared.} On AMD GPUs, it is not possible to use a single swizzle pattern for all layouts (a simple counter-example is provided in Appendix \ref{app:proof-by-contradiction}). While we could implement unique swizzle patterns for every matrix layout, this adds code complexity. Instead, we identify the layouts that commonly co-occur and support swizzle patterns that are bank conflict free for these instances.  \Cref{fig:swizzle_plot} shows one such swizzle that is bank conflict free for both the $16\times32$ row layout and column layout load.
    \item \textbf{Global.} AMD GPUs support direct asynchronous HBM to shared memory loads. Like TMA, these loads bypass the register file. The instruction takes as input per-thread addresses in HBM from which each thread will read data. While DSLs like {\shorttk} directly swizzle the shared memory addresses, swizzling shared memory on AMD is instead accomplished by swizzling on the HBM addresses.
\end{enumerate}

\vspace{-2mm}
\subsection{Overlapping compute and memory utilization}
\label{sec:hip-block}
We study the principles for scheduling instructions within AMD AI kernels and identify two high-performance patterns that lead to peak utilization across diverse workloads.

\vspace{-3mm}
\paragraph{Current approaches and their limitations.} 
State-of-the-art AI kernels and DSLs have consolidated around \emph{wave specialization}---a pattern where specialized producer waves handle memory movement while consumer waves handle computation. This approach dominates in NVIDIA implementations including FlashAttention-3~\cite{dao2024flashattention3}, COMET for MoE~\cite{25-comet}, and GEMMs~\cite{bikshandi2023developing}, and kernel DSLs like {\shorttk}~\cite{spector2025tk} and TileLang~\cite{wang2025tilelang}. In this paradigm, waves occupy specific hardware units for long periods of time, so they can issue bulk operations over large tile primitives. This tile-based programming makes the code size compact and readable.

However, wave specialization struggles to generalize to modern AMD devices due to fundamental architectural differences. Instead, state-of-the-art AMD kernels (AITER~\cite{amd2025aiter}, CK~\cite{amd2025ck}) resort to raw assembly to finely interleave instruction issues---an approach orthogonal to tile-based programming. While it might seem that AMD requires bespoke schedules for each AI workload, we identify simple general principles that achieve high performance across diverse applications.

\vspace{-2mm}
\subsubsection{Wave specialization underperforms on AMD}
NVIDIA kernels implement wave specialization using dedicated memory access hardware (\textit{tma}), asynchronous matrix multiplies which accept operands directly from shared or tensor memory (\textit{wgmma}, \textit{tcgnen05}), deep pipelines enabled by large shared memory per processor (B200 has 40\% larger SRAM than AMD MI355X per processor), register reallocation (where the register-efficiency of TMA lets producers give their registers to consumers), and hardware synchronization primitives (\textit{mbarriers}).
AMD lacks these architectural features, changing the kernel design space.

\input{figures/wave_specialization/figure}

To evaluate how these differences impact performance, we vary the synchronization mechanism, pipeline depth, and producer-consumer ratios (Tab.~\ref{tab:producer_consumer_ablations}). Our experiments reveal two principles. We need to maximize the output tile size computed per thread block to increase the arithmetic intensity (operations per byte moved), and we need to maximize the pipeline depth to hide the latency of memory loads. 

Peak performance open-source {\shorttk} and CUTLASS profiler-selected GEMMs use wave specialization and an output tile size of $256\times256$ on the B200.~\footnote{The profiler sweeps and tunes the suite of CUTLASS GEMMs, selecting the best one for the shape and dtype.}i 

Our best AMD GEMM achieves comparable performance when computing a $256\times256$ output tile per thread block only when using \textit{no wave specialization} (i.e., zero producers) and degrades as the number of producers increases (Tab.~\ref{tab:producer_consumer_ablations}).
This is because AMD hardware statically divides registers across all waves~\cite{amd2025hardware}, meaning producers consume registers without contributing to output computation. This limits the usable output tile size when using wave specialization.  

\vspace{-2mm}
\paragraph{Tradeoffs.} NVIDIA's larger shared memory enables the use of deep pipelines while using large matrix instruction shapes (e.g., $256\times256\times16$). However, AMD's smaller tensor core shapes (e.g., $16\times16\times32$) provide an alternative path to establish deep pipelines by using finer-granularity load and compute stages. 

NVIDIA's matrix multiply instructions, which accept operands from shared or tensor memory, helps alleviate register pressure, and it may be surprising that AMD can match performance without this. However, AMD devices have a $2\times$ larger register file to compensate.

We also validate that using shared memory atomics instead of \textit{mbarriers} adds negligible overhead; we find the $192\times256$ producer consumer kernel, which uses atomics, performs similarly to our non-wave-specialized kernel emphasizing that the output tile shape is the dominant factor impacting performance (Tab.~\ref{tab:producer_consumer_ablations}). 

\vspace{-2mm}
\subsubsection{Performant scheduling patterns for AMD AI kernels}

AMD GPUs have four SIMD units per CU, and waves scheduled on the same SIMD can overlap compute and memory instructions. We identify two scheduling patterns that consistently achieve peak performance across AI workloads by exploiting this parallelism differently:

\input{figures/pingpong/figure}

\begin{enumerate}[itemsep=2pt,topsep=0pt,leftmargin=*]
    \item \textbf{\textsc{8-wave ping-pong} (balanced workloads).} 
    This pattern employs eight waves per thread block---two resident per SIMD. The waves are split into two groups of four, with each group containing one wave per SIMD. Within each SIMD, the two waves alternate the type of work each does: one issues only compute instructions while the other issues only memory instructions, and then they swap roles, flipping back and forth between compute and memory as shown in Figure~\ref{fig:main}. A conditional barrier controls the alternation.
    
    This pattern excels when compute and memory durations are roughly balanced. A SIMD's compute wave executes matrix fused multiply add (MFMA) instructions while its paired memory wave prefetches the next data, hiding memory effectively.
    
    \item \textbf{\textsc{4-wave interleave} (imbalanced workloads).} 
    This pattern places exactly one wave on each of the processor's four SIMDs. Each wave issues both compute and memory instructions in a carefully staggered sequence to maximize occupancy of the hardware units.
    
    This fine-grained pattern better saturates both MFMA and LDS pipelines when workloads are imbalanced (either compute-heavy or memory-heavy). The wave per SIMD can adapt its instruction mix dynamically.
\end{enumerate}

These schedules tradeoff programmability and performance. {\shortname} lets developers use tile-based primitives to implement either of these patterns, albeit at different tile \textit{granularities}. \textsc{8-wave ping-pong} allows for large tile primitives similar to the ones used in \textit{wave specialization}. On the other hand, \textsc{4-wave interleave} requires developers to program with small base tile primitives, extending the code size due to finer grained instruction issues.
This tradeoff is captured in~\Cref{fig:pingpong}. Surprisingly, we find that \textsc{8-wave} is \textit{sufficient} to match or outperform AMD's raw assembly kernels across BF16 GEMM, FP8 GEMM, and attention forwards workloads. On GQA non-causal attention backwards, our \textsc{8-wave} kernel outperforms the baselines (PyTorch SDPA, CK, and AITER) by $1.8\times$, and our \textsc{4-wave} kernel delivers an even larger $2.3\times$ speedup.

\subsection{Optimizing the access patterns of non-programmable GPU memory}

Modern GPUs---AMD and NVIDIA---are moving towards chiplet, rather than monolithic, architectures (e.g., Blackwell is comprised of two chips).
This results in a disaggregated cache hierarchy, where distinct \textit{clusters} of processors are attached to distinct slices of the GPU cache (see Figure~\ref{fig:hardware-overview}). Here, we explore principles for disaggregated cache scheduling and introduce {\shortname}'s algorithm for cache reuse.

\label{sec:hip-grid}

\vspace{-2mm}
\paragraph{Cost model.} AMD devices use two types of caches -- L2 and LLC -- where cache misses have a worst case miss penalty of 300ns for the L2 cache and 500ns for the LLC cache. AMD devices assign clusters of 32 (CDNA4) or 38 (CDNA3) compute units to a cluster (accelerated complex die, or XCD), and include $8$ clusters per GPU. The hardware scheduler assigns thread blocks to XCDs in round-robin order. 
The grid schedule, or order of work assigned to thread blocks, impacts the cache reuse and achieved bandwidth:

\begin{equation}
\begin{split}
\text{Bandwidth}
= \text{LLC Bandwidth } \times \text{LLC Hit \%} \\
+ \text{L2 Bandwidth } \times \text{L2 Hit \%}
\end{split}
\label{eq:cache_cost}
\end{equation}

\input{figures/cache_micros/l2_xcd}

In a GEMM kernel ($D = AB + C$), each thread block computes a distinct tile of the output matrix $D$. When thread blocks are scheduled in naïve row-major order, cache reuse is suboptimal ($\approx$55\%) because blocks that share the same L2 cache often load different, non-overlapping tiles of $A$ and $B$. Thus, their memory accesses fail to exploit spatial locality, leading to redundant data movement. This behavior is illustrated in Fig.~\ref{fig:default} and Tab.~\ref{tab:l2_table} (Row 1).
To mitigate this, we use two key principles to improve cache reuse:

\vspace{2mm}
\begin{enumerate}[itemsep=0.1pt,topsep=0pt,leftmargin=*]
    \item \textbf{L2 Reuse.} Thread blocks mapped to the same XCD (and thus sharing an L2 cache) should cover a rectangular region of the output matrix---an “L2 tile.” This layout ensures that consecutive blocks reuse both the same rows of $A$ and the same columns of $B$. However, optimizing purely for L2 locality can cause each XCD to fetch disjoint portions of $A$ and $B$, leading to redundant loads at the next cache level.
    \item \textbf{LLC Reuse.} To further improve reuse at the last-level cache (LLC), we must coordinate accesses across XCDs. Ideally, the combined access footprint of all XCDs---the “LLC tile”---should overlap in both $A$ and $B$. In other words, multiple XCDs should work on nearby or identical regions of the input matrices, so that shared data remains resident in the LLC.
\end{enumerate}
By jointly optimizing these two principles, we can raise both L2 and LLC hit rates, leading to higher effective bandwidth (Figure \ref{fig:w5}, Table \ref{tab:l2_table}, Row 3).
For instance, Table \ref{tab:l2_table_14592} shows that an L2/LLC-aware schedule achieves up to 15\% higher performance than the default grid order. The benefit is particularly pronounced when the output matrix width (in tiles) is coprime with the number of XCDs—for example, 57 tiles across 8 XCDs on an AMD MI355X---since the default schedule causes worst case reuse patterns (Tab.~\ref{tab:l2_table_14592}).

\vspace{-2mm}
\paragraph{{\name} chiplet swizzling algorithm.} 

\begin{algorithm*}[t]
\caption{\texttt{XCD swizzle for cache reuse on GEMMs}}
\begin{algorithmic}[1]\label{algo:xcdswizzle}
\Require grid block indices $(b.x,b.y,b.z)$; grid dimensions $(g.x,g.y,g.z)$;
number of XCDs $\texttt{nXCD}$;
problem sizes $M,N$ with tile sizes $\texttt{BLOCK}_M,\texttt{BLOCK}_N$;
window height $W$, chunk size $C$

\Ensure remapped block indices $(b.x', b.y', b.z)$

\State $\texttt{blocks} \gets g.x \times g.y$ \Comment{blocks per batch (a single $b.z$ slice)}
\State $xy \gets b.x + g.x \times b.y$ \Comment{flatten $(b.x,b.y)$ within the batch}

\State $\texttt{blocks\_per\_cycle} \gets \texttt{nXCD} \times C$
\State $\texttt{limit} \gets \left\lfloor \dfrac{\texttt{blocks}}{\texttt{blocks\_per\_cycle}} \right\rfloor \times \texttt{blocks\_per\_cycle}$ \Comment{largest full $(\texttt{nXCD}\!\times\!C)$-aligned prefix}
\If{$xy > \texttt{limit}$}
  \State $xy \gets xy$ \Comment{tail region: leave order unchanged}
\Else
  \State $\texttt{xcd} \gets xy \bmod \texttt{nXCD}$ \Comment{which XCD this block belongs to (round-robin)}
  \State $\texttt{local} \gets \left\lfloor \dfrac{xy}{\texttt{nXCD}} \right\rfloor$ \Comment{local index after de-interleaving by XCD}
  \State $\texttt{chunk\_idx} \gets \left\lfloor \dfrac{\texttt{local}}{C} \right\rfloor$ \State $\texttt{pos} \gets \texttt{local} \bmod C$
  \State $xy \gets \texttt{chunk\_idx} \times \texttt{blocks\_per\_cycle} + \texttt{xcd} \times C + \texttt{pos}$
\EndIf

\State $\texttt{num\_rows} \gets \frac{M}{\texttt{BLOCK}_M}$ \Comment{tile rows along $M$}
\State $\texttt{num\_cols} \gets \frac{N}{\texttt{BLOCK}_N}$ \Comment{tile cols along $N$}
\State $\texttt{tid\_per\_group} \gets W \times \texttt{num\_cols}$ \Comment{one window (height $W$) across all columns}
\State $\texttt{group\_id} \gets \frac{xy}{\texttt{tid\_per\_group}}$ \Comment{which window of rows}
\State $\texttt{first\_row} \gets \texttt{group\_id} \times W$
\State $\texttt{win\_h} \gets \min(\texttt{num\_rows} - \texttt{first\_row},\, W)$ \Comment{tail-safe window height}
\State $\ell \gets xy \bmod \texttt{tid\_per\_group}$ \Comment{local index within the window}
\State $\texttt{row} \gets \texttt{first\_row} + (\ell \bmod \texttt{win\_h})$ \Comment{fast index: go \emph{down} within the column}
\State $\texttt{col} \gets \frac{\ell}{\texttt{win\_h}}$ \Comment{slow index: move to next column after $\texttt{win\_h}$ rows}
\State \Return $(\texttt{row}, \texttt{col}, b.z)$ \Comment{logical tile coordinates (+ batch)}
\end{algorithmic}
\label{alg:xcd-swizzle}
\end{algorithm*}

To make cache-aware scheduling accessible to developers, {\name} provides a simple and tunable strategy for maximizing cache reuse across a wide range of GEMM problem sizes.
\Cref{alg:xcd-swizzle} implements this strategy in two steps:
\vspace{2mm}
\begin{enumerate}[itemsep=0.1pt,topsep=0pt,leftmargin=*]
    \item \textbf{XCD grouping.} Flatten the 2D-grid into a linear sequence and remap block ID's such that chunks of $C$ consecutive IDs are resident on the same XCD. This reduces cross-chiplet traffic.
    \item \textbf{Hierarchical windowed traversal.} Instead of processing the grid row by row, we process it in vertical windows of height $W$. This has the effect of ``folding'' the input block ID space into rectangular tiles, optimizing L2 cache reuse.
\end{enumerate}
\vspace{2mm}

The two parameters, $W$ and $C$, control the trade-off between L2 and LLC reuse.
Since L2 bandwidth is roughly $3\times$ higher than LLC bandwidth, $W$ should be chosen to maximize L2 hit rate.
On AMD MI355X, each XCD contains 32 CUs, and empirical results show that L2 tiles of shape $8\times4$ or $4\times8$ achieve the best hardware utilization.
Tuning the chunk size $C$ further improves LLC efficiency by coordinating access patterns across XCDs so that they operate on similar rows of the input matrix.

%% file: figures/programmable_memory/mi355x_attn_bkwd.tex
\begin{table}[t!]
    \vspace{-4mm}
    \begin{center}
    \begin{small}
    \begin{sc}
    \begin{tabular}{lccc}
        \toprule
        Method & Seq. Length & TFLOPS \\
        \midrule
        \textsc{HK} & 4096 & 855 \\
        \textsc{HK with Pinned Registers} & 4096 & \textbf{1024} \\
        AMD Assembly (AITER) & 4096 & 1018 \\
        \midrule
        \textsc{HK} & 8192 & 909 \\
        \textsc{HK with Pinned Registers} & 8192 & 1091 \\
        AMD Assembly (AITER) & 8192 & \textbf{1169} \\
        \bottomrule
    \end{tabular}
    \vspace{-4mm}
    \end{sc}
    \end{small}
    \end{center}
    \caption{\textbf{{\textsc{Explicit register scheduling} enables increased developer control}.} A 4-wave MHA non-causal backwards kernel implemented in HIP underperforms AMD's raw assembly kernel (AITER) due to compiler limitations. We match AITER by bypassing the compiler and pinning register tiles to explicit registers. We use batch size $16$, heads $16$ and head dim $128$.}
    \label{table:attn_bkwd_micro}
    \vspace{-4mm}
\end{table}

%% file: figures/programmable_memory/matrix_layouts.tex
\begin{figure}[h!]
\vspace{-4mm}
  \captionsetup[subfigure]{
    font=small,
    labelformat=parens,
    labelsep=space,
    justification=centering,
    singlelinecheck=false,
    skip=2pt,
    width=\linewidth
  }
  \captionsetup[figure]{font=small}

  \centering
  \begin{subfigure}[t]{0.2\linewidth}
    \includegraphics[width=\linewidth,height=0.25\textheight,keepaspectratio]{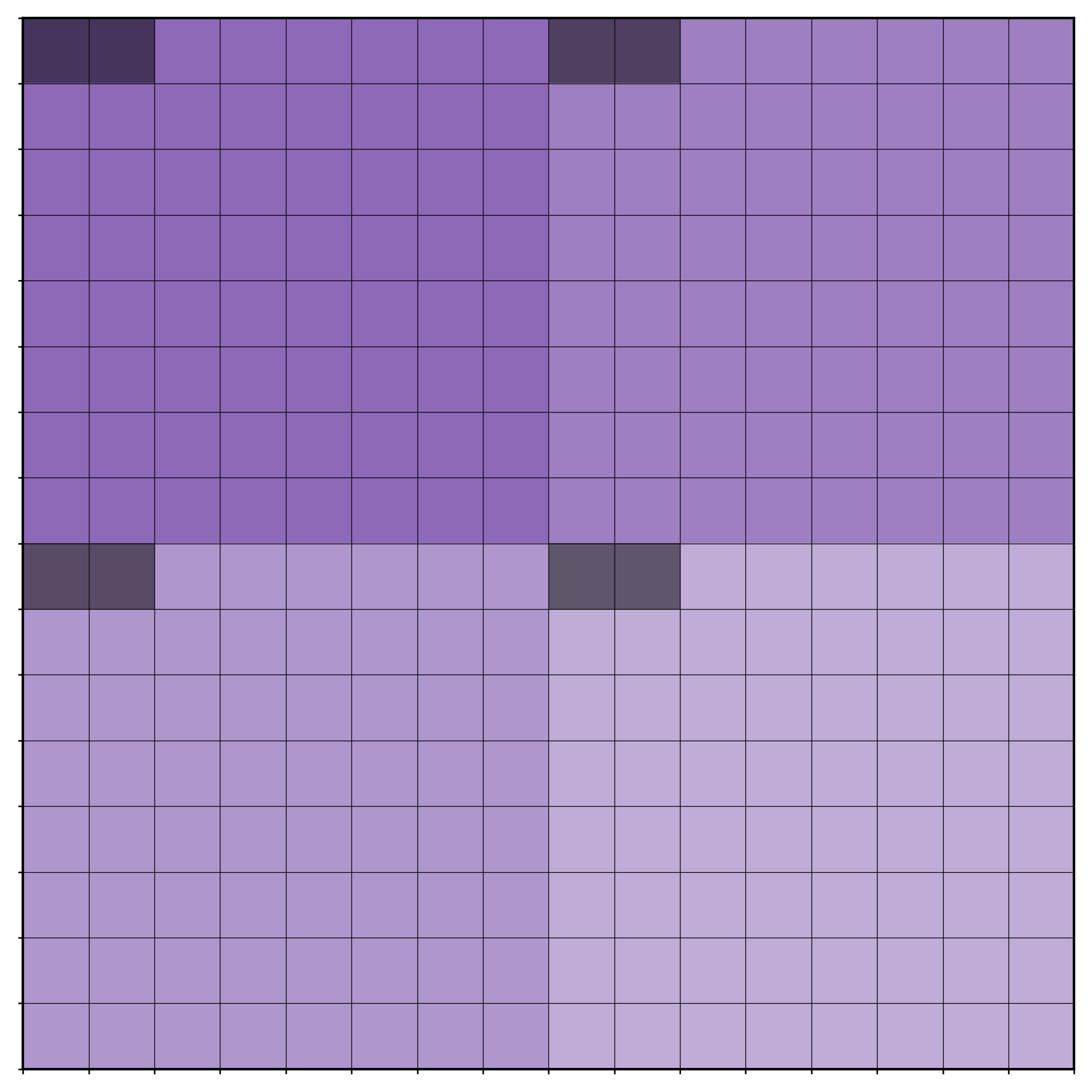}
    \subcaption{NVIDIA core matrices}\label{fig:nvidia-core-matrix}
  \end{subfigure}
  \hspace{0.03\linewidth}
  \begin{subfigure}[t]{0.4\linewidth}
    \includegraphics[width=\linewidth,height=0.25\textheight,keepaspectratio]{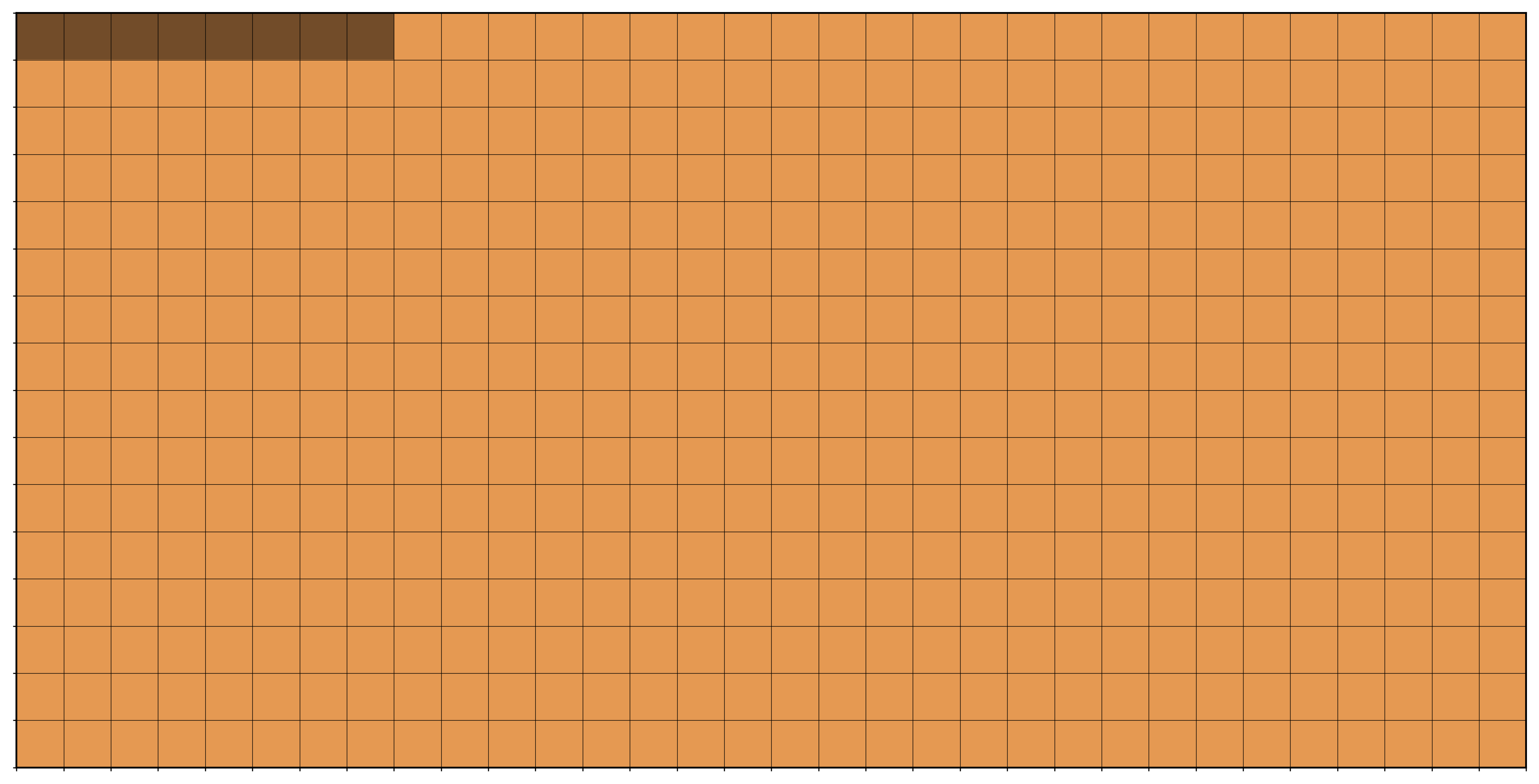}
    \subcaption{AMD 16\(\times\)16\(\times\)32 MFMA A/B matrix}\label{fig:amd-16x32}
  \end{subfigure}
  \hspace{0.03\linewidth}
  \begin{subfigure}[t]{0.2\linewidth}
    \includegraphics[width=\linewidth,height=0.25\textheight,keepaspectratio]{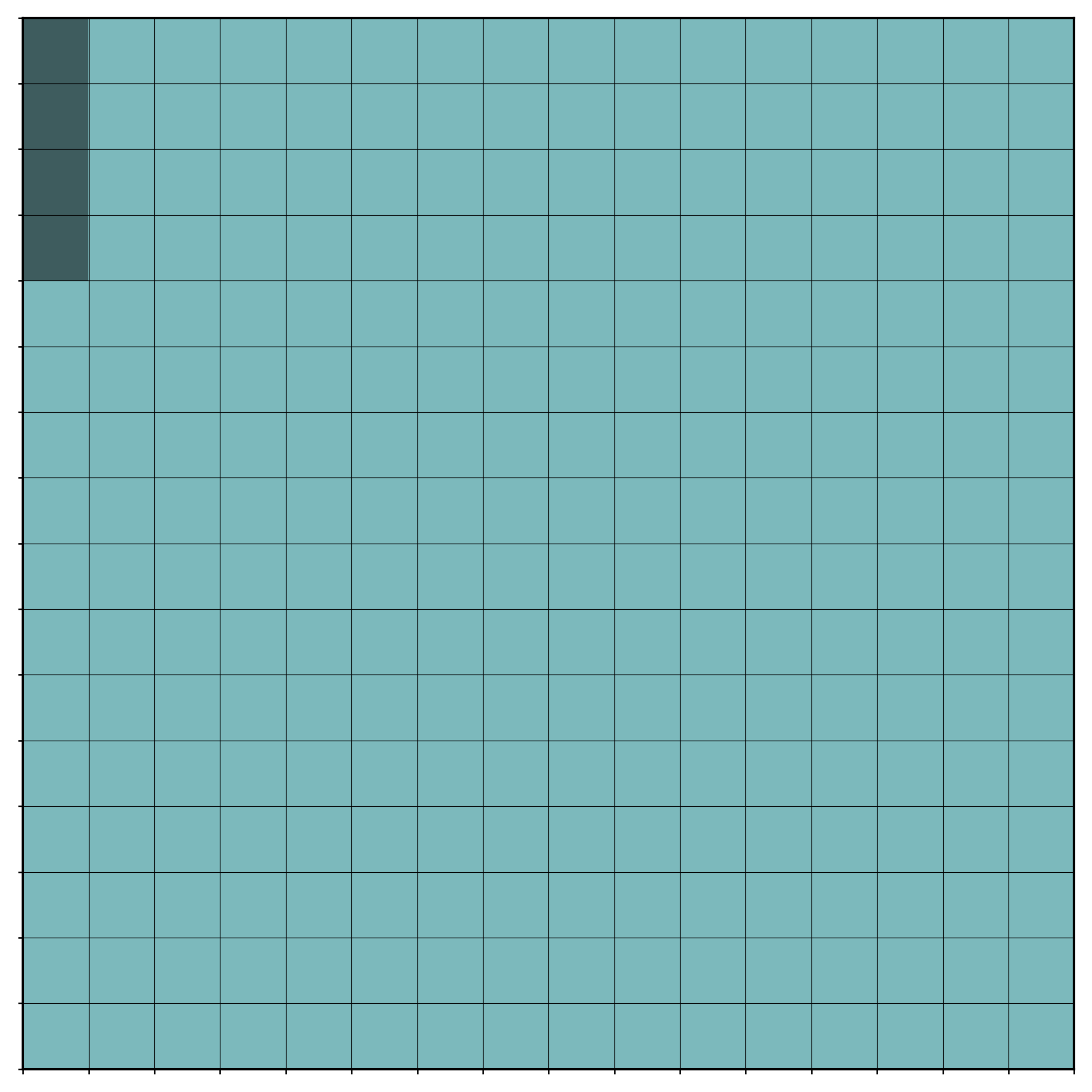}
    \subcaption{AMD 16\(\times\)16\(\times\)32 C/D matrix}\label{fig:amd-16x16}
  \end{subfigure}
  \caption{Matrix layouts on NVIDIA and AMD GPUs. The shaded cells in each matrix represent elements owned by thread 0.}
  \label{fig:core-matrix-comparison}
  \vspace{-4mm}
\end{figure}

%% file: figures/programmable_memory/swizzle.tex
\begin{figure*}[h!]
    \includegraphics[width=\linewidth]{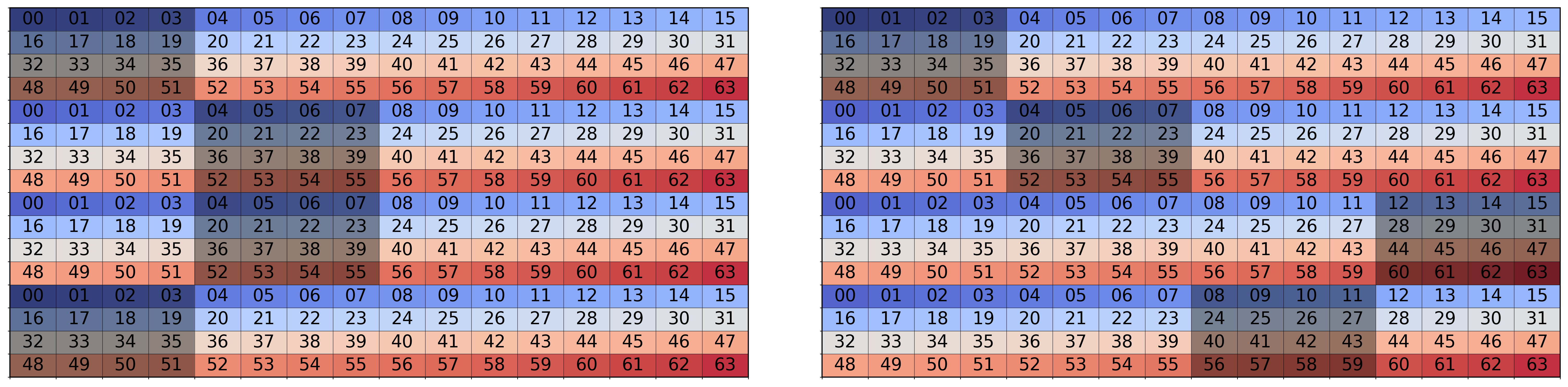}
    \vspace{-4mm}
    \caption{\textbf{Swizzle pattern for a 16x32 tile of BF16s.} Shared memory on AMD CDNA4 GPUs have different banking behavior depending on the instruction. \texttt{ds\_read\_b128} accesses shared memory through 64 banks, each 32-bits wide, and correspond the individual cells and numbers in the figure. The shaded cells represent banks that are accessed by the first phase of a \texttt{ds\_read\_b128} instruction for a 16x32 row layout register tile. On the left is an unswizzled layout suffering from 2-way bank conflicts. On the right is a swizzled layout with no bank conflicts. The swizzle applied here swaps the first 8 columns with the last 8 starting from the 8th row. This swizzling strategy simultaneously enables bank-conflict free accesses from column-major reads using \texttt{ds\_read\_b64\_tr\_b16}.  Details can be found in \ref{app:layouts}.}
    \label{fig:swizzle_plot}
\end{figure*}

%% file: figures/wave_specialization/figure.tex
\begin{table}[t]
\centering
\small
\begin{sc}
\begin{tabular}{lccccc}
\toprule
\# P / \# C & MFMA Shape &  Output &  TFLOPS \\
\midrule
{\shortname} 4 / 8 & $16\times16\times32$ & $128\times256$ & 893 \\
{\shortname} 4 / 12 & $16\times16\times32$ & $192\times256$ & 1278 \\
\midrule
{\shortname} 0 / 8 & $16\times16\times32$ & $192\times256$ & 1281 \\
{\shortname} 0 / 8 & $16\times16\times32$ & $256\times256$ & \textbf{1610} \\
\toprule
{\shorttk} & $256\times256\times16$ & $256\times256$ & 1538 \\
CUTLASS & $256\times256\times16$ & $256\times256$ & 1570 \\
\bottomrule
\end{tabular}
\end{sc}
\caption{\textbf{Producer consumer comparisons.} We report results for a series of producer consumer BF16 GEMM kernels of shape $M=N=K=8192$. We denote the number of producers and consumers as P and C respectively. We denote the underlying matrix instruction size, output tile size computed per thread block, and TFLOPS measured (500 iterations of warmup, 100 iterations of measurement on inputs from $\mathcal{N}(0, 1)$). AMD kernels run on an MI355X and NVIDIA kernels (TK, CUTLASS) on a B200. 
}
\label{tab:producer_consumer_ablations}
\vspace{-4mm}
\end{table}

%% file: figures/pingpong/figure.tex
\begin{table}[t]
    \vspace{-4mm}
    \begin{center}
    \begin{small}
    \begin{sc}
    \begin{tabular}{lccc}
        \toprule
        Kernel &  Pattern &  LoC & TFLOPS \\
        \midrule
        FP8 GEMM & 8-wave & 48 & 3222 \\
        FP8 GEMM & 4-wave &  183 & 3327 \\
         \midrule
        MHA backwards & 8-wave & 331 & 894 \\
        MHA backwards & 4-wave & 989 & 1091 \\
        \bottomrule
    \end{tabular}
    \vspace{-10pt}
    \end{sc}
    \end{small}
    \end{center}
    \caption{\textbf{Scheduling patterns for AMD.} We identify two primary paradigms---\textsc{8-wave} and \textsc{4-wave}---that generalize across workloads. Both patterns can leverage {\shortname}'s tile primitives. We report the hot loop code size and TFLOPs, showing how these patterns trade off programmability and performance.}
    \label{fig:pingpong}
    \vspace{-4mm}
\end{table}

%% file: figures/cache_micros/l2_xcd.tex
\begin{figure}[t]
  \centering
  \begin{subfigure}[t]{0.3\linewidth}
    \includegraphics[width=\linewidth,height=0.25\textheight,keepaspectratio]{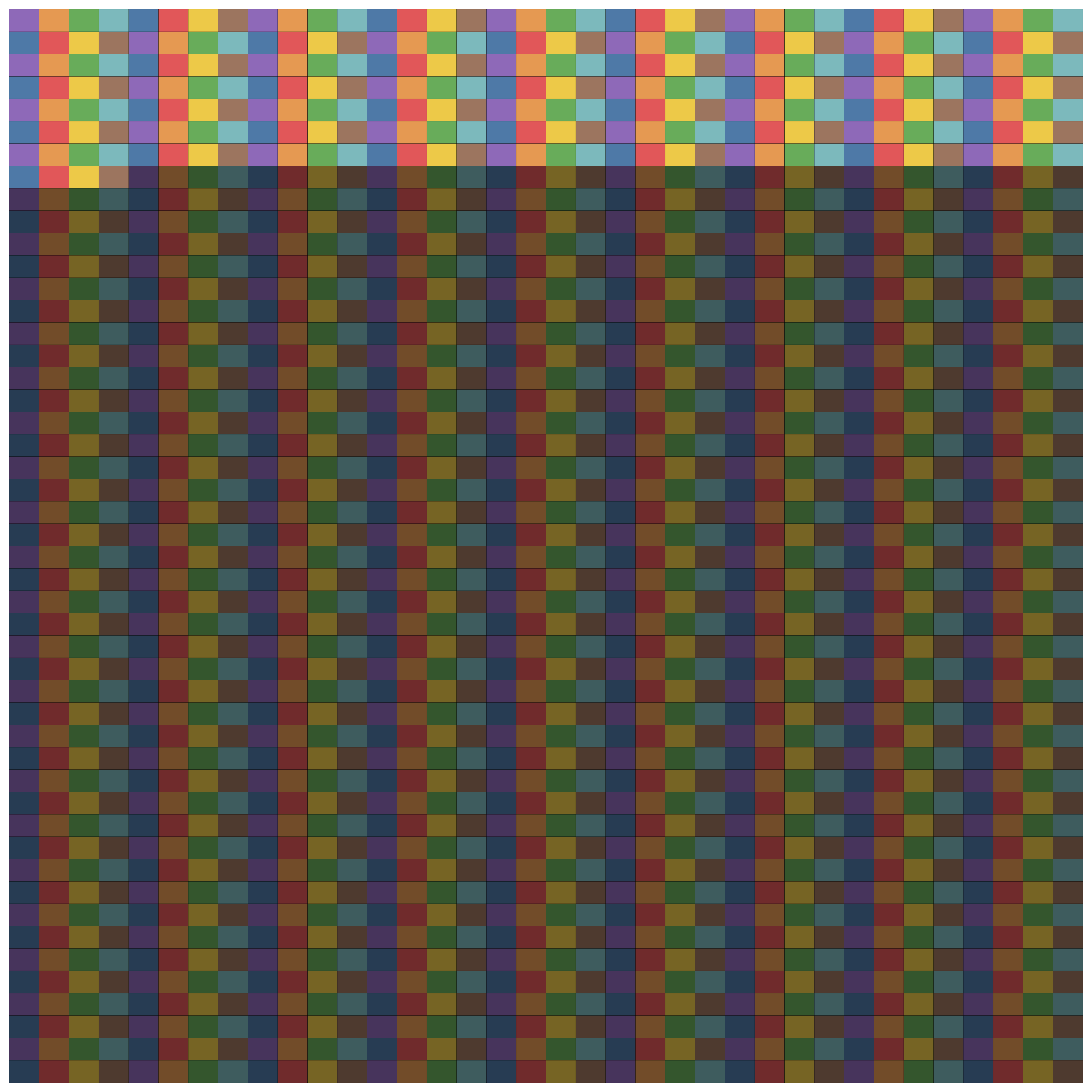}
    \subcaption{Row-major}\label{fig:default}
  \end{subfigure}\hfill
  \begin{subfigure}[t]{0.3\linewidth}
    \includegraphics[width=\linewidth,height=0.25\textheight,keepaspectratio]{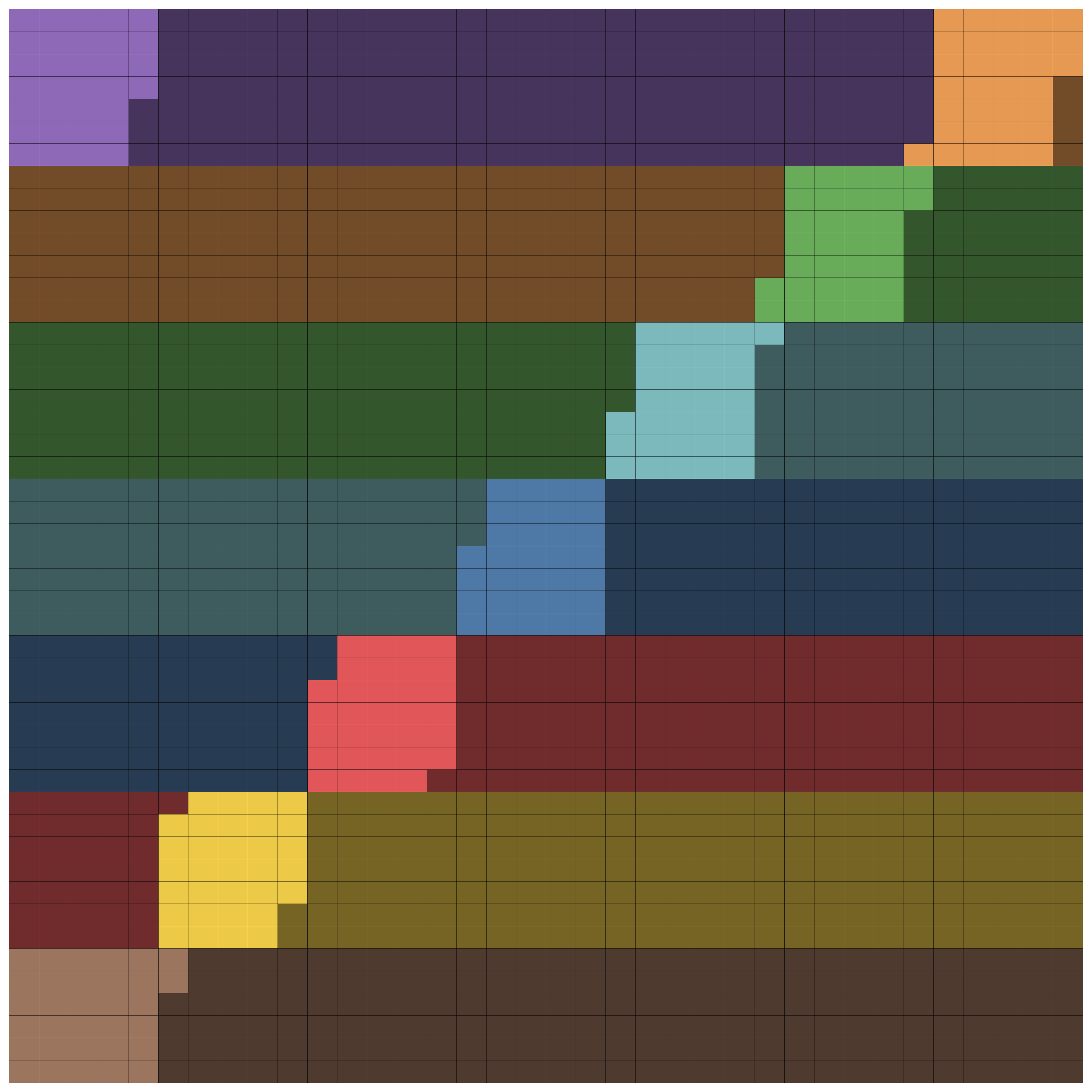}
    \subcaption{XCD (W7, C216)}\label{fig:w7}
  \end{subfigure}\hfill
  \begin{subfigure}[t]{0.3\linewidth}\centering
    \includegraphics[width=\linewidth,height=0.25\textheight,keepaspectratio]{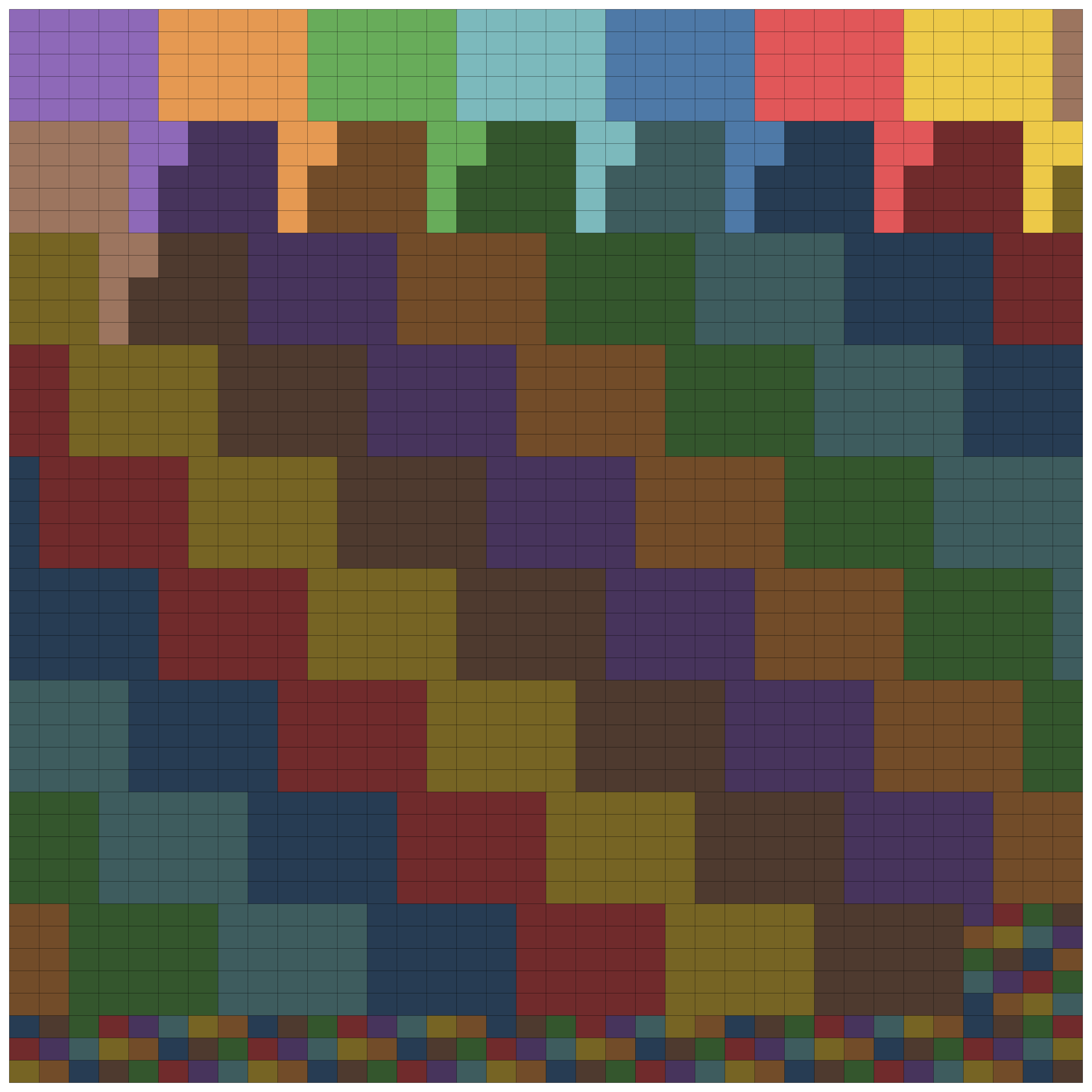}
    \subcaption{XCD (W5, C25)}\label{fig:w5}
  \end{subfigure}
\end{figure}

\begin{table}[t]
\centering
\small 
\begin{sc}
\setlength{\tabcolsep}{4pt} 
\begin{tabular}{lcccc}
  \toprule
  Block Order & L2 \% & LLC \%  & Mem. BW  & TFLOPS \\
  \midrule
  \multicolumn{5}{c}{\textbf{Matrix Multiply (M=N=K=9216, MT 192x256x64)}} \\
  \midrule
  Row-major          & 55\% & 95\%  & 15.1 TB/s & 1113 \\
  XCD ($W 7/C 216$)  & 79\% & 24\%  & 14.9 TB/s &  991 \\
  XCD ($W 5/ C 25$)   & 75\% & 93\%  & 18.3 TB/s & 1145 \\
  \toprule
  \multicolumn{5}{c}{\textbf{Matrix Multiply (M=N=K=14592, MT 192x256x64)}} \\
  \midrule
  Row-major     & 36\% & 76\%  & 10.7 TB/s & 900 \\
  XCD ($W 8/ C 542$)  & 79\% & 7\%  & 13.9 TB/s &  980 \\
  XCD $W 8/C 64$  & 78\% & 55\%  & 16.6 TB/s & 1068 \\
  \bottomrule
\end{tabular}
\caption{\textbf{Chiplet swizzling for cache reuse. } Visualization of three different grid schedules for the output matrix of a $M=N=K=9216$ BF16 GEMM. The color represents the XCD assignment for the first set of thread blocks scheduled across the GPU (256 CUs). Schedule \ref{fig:default} (Table Row 1) assigns blocks to the grid according to block ID. Schedules \ref{fig:w7} (Table Row 2) and \ref{fig:w5} (Table Row 3) apply \Cref{alg:xcd-swizzle} with different window and chunk size parameters. Table \ref{tab:l2_table} shows how these schedules trade off L2 and LLC reuse to gain performance. \Cref{fig:14592_default} provides the corresponding visualization for for the $14592$ shape.}
  \label{fig:three-grids}
\label{tab:l2_table}
\label{tab:l2_table_14592}
\vspace{-2mm}
\end{sc}
\end{table}

%% file: sections/4-experiments.tex
\input{figures/results/main_gemm_results}

\input{figures/results/main_attention_results}

\vspace{-2mm}
\section{Experiments}
\label{sec:sec4_results}

In this section, we validate that {\shortname} enables peak performance kernels, using simple and reusable tile-based primitives, across a breadth of AI operations.

\vspace{-2mm}
\paragraph{Baselines.} We compare to the best performing baseline kernels across PyTorch (compiled and SDPA), AITER~\cite{amd2025aiter}, Composable Kernel~\cite{amd2025ck}, ROCm Libary Triton~\cite{amd2025rocmlib}, and HipBLASLT~\cite{amd2025rocmlib}. We evaluate on both MI325 CDNA3 and MI355 CDNA4. We benchmark {\shortname} kernels in Python scripts using Python bindings (ecept FP8 where AMD PyTorch support remains experimental). For each kernel, we use 500 warmup runs and report the average TFLOPs/s performance over 100 runs over randomly generated input tensors from the standard normal distribution. All kernels are benchmarked in AMD's recently released beta Docker using ROCm 7.0 (\url{rocm/7.0-preview:rocm7.0_preview_pytorch_training_mi35x_beta}).

\label{sec:results_kernels}

\input{figures/results/main_attention_bkwd}
\input{figures/results/main_mem_bound_results}

{\shortname} provides a comprehensive suite of peak-performance AMD AI kernels, written using reusable tile-based abstractions. We also include code listings in Appendix~\ref{app:kernel_listings} and additional results in Appendix~\ref{app:extended_results}:
\begin{enumerate}[itemsep=0.1pt,topsep=0pt,leftmargin=*]
    \item \textbf{BF16 and FP8 GEMM.} {\shortname} competes with the AMD baseline kernels that are written in assembly (AITER, HipBLASLTt/PyTorch). {\shortname} outperforms the Triton compiler by $1.3-3.0\times$. Further, we obtain these results using a single 8-wave kernel schedule that generalizes across the evaluated problem shapes.
    
    \item \textbf{Attention forwards.} We evaluate multi-head attention (MHA) and group-query attention (GQA) kernels in causal and non-causal settings, and for head dimensions $64$ and $128$. {\shortname} outperforms all available AMD baselines on average, including the AITER kernels, which are written in hand-optimized raw assembly by AMD engineers.
    {\shortname} is $1.0-2.1\times$ faster than AITER, $1.3-4.5\times$ faster than PyTorch (SDPA), $1.0-1.4\times$ CK, and $1.2-4.5\times$ Triton kernels in~\Cref{fig:attention}.

    {\shortname}’s attention forward kernel uses an \textsc{8-wave ping-pong}. Within compute clusters, each wavefront interleaves online-softmax vector ops (max/subtract/exp2/accumulate) with MFMA instructions. 
    Despite substantial scheduling and hardware differences between MI355X and NVIDIA B200, the kernel is competitive with FlashAttention-3 under comparable settings \cite{dao2024flashattention3}.

    \item \textbf{Attention backwards.} Our GQA causal and non-causal backwards attention kernels outperform the baselines by $1.8-2.5\times$ across settings (Fig.~\ref{fig:attention_bkwd}). Our MHA kernels compete with the strongest available baselines, which are written in assembly (Fig.~\ref{fig:app_attention_mha_bwd}).

    Attention backwards is a notoriously register heavy workload. Our efficient {\shortname} kernel uses multiple MFMA instruction shapes ($16\times16\times32$ and $32\times32\times16$), different shared memory access patterns (e.g., both row and column layout loads to registers from the same shared tile), and explicit register pinning.

    \item \textbf{Memory bound results.} We consider a fused dropout-residual-layernorm kernel (from prenorm Transformer architectures) and a rotary positional encoding kernel in \Cref{fig:mem_bound}. {\shortname} outperforms both AITER and PyTorch compiled kernels by $1.1-2.2\times$ across settings.
\end{enumerate}

The inconsistent performance of AMD libraries and the difficulty of scaling assembly-engineered kernels (e.g., reinforced by head dim. 64 attention and GQA non-causal backwards) reflects the value of having a simple set of kernel programming abstractions to accelerate AMD kernel development. 
Finally, to validate kernel stability, we use our kernels to pretrain Llama 1B~\cite{llama3modelcard} and BERT 110M~\cite{devlin2020transformers} on the Slim Pajama corpus, matching the perplexity of models trained using PyTorch and AITER after 10B tokens of training.

%% file: figures/results/main_gemm_results.tex
\begin{figure}[t]
\centering
\includegraphics[width=0.48\linewidth]{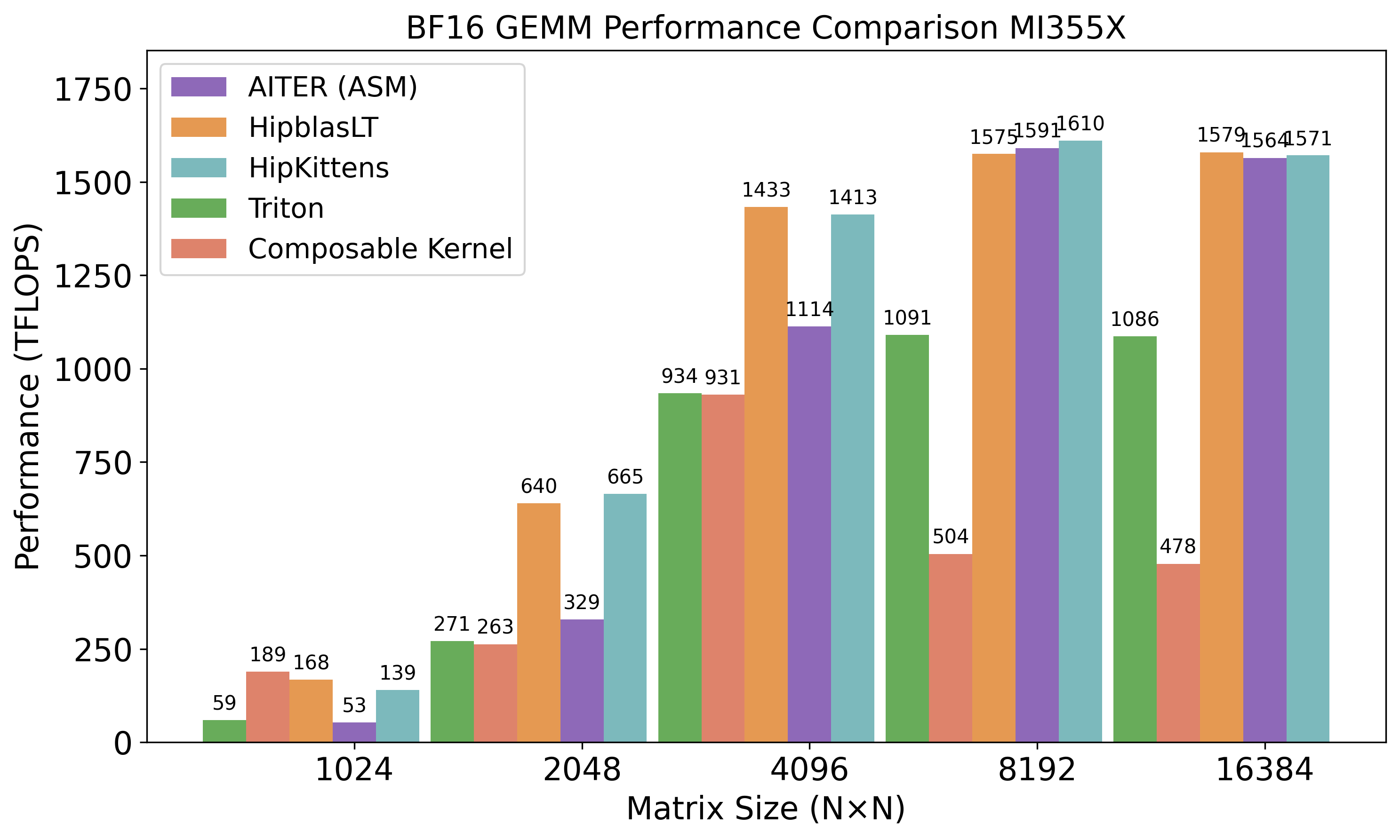}
\hfill
\includegraphics[width=0.48\linewidth]{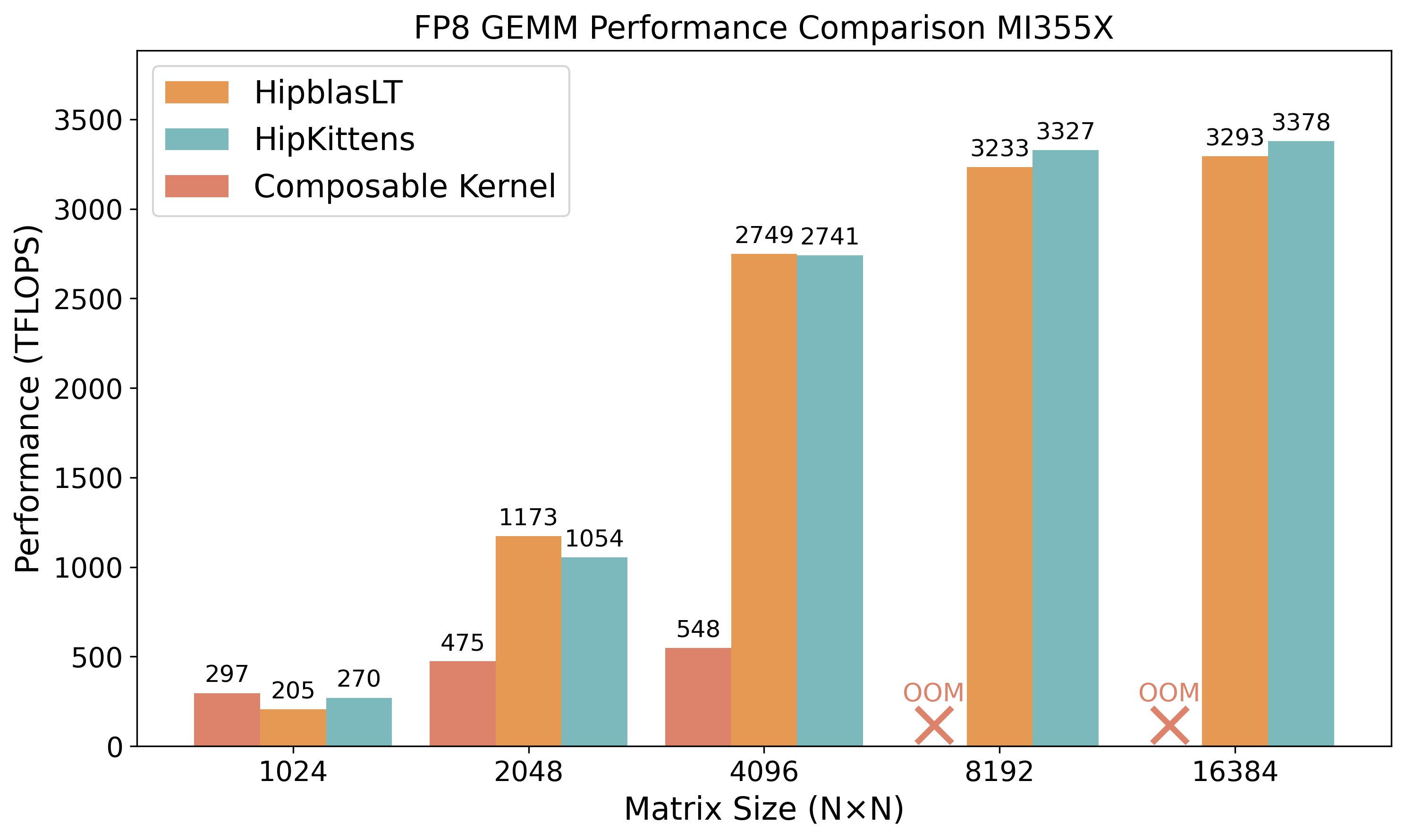}
\vspace{-2mm}
\caption{\textbf{GEMM.} We compare {\shortname} BF16 and FP8 GEMMs to the strongest available baselines.}
\vspace{-2mm}
\label{fig:gemm}
\end{figure}

%% file: figures/results/main_attention_results.tex
\begin{figure}[h!]
\centering
\includegraphics[width=0.48\linewidth]{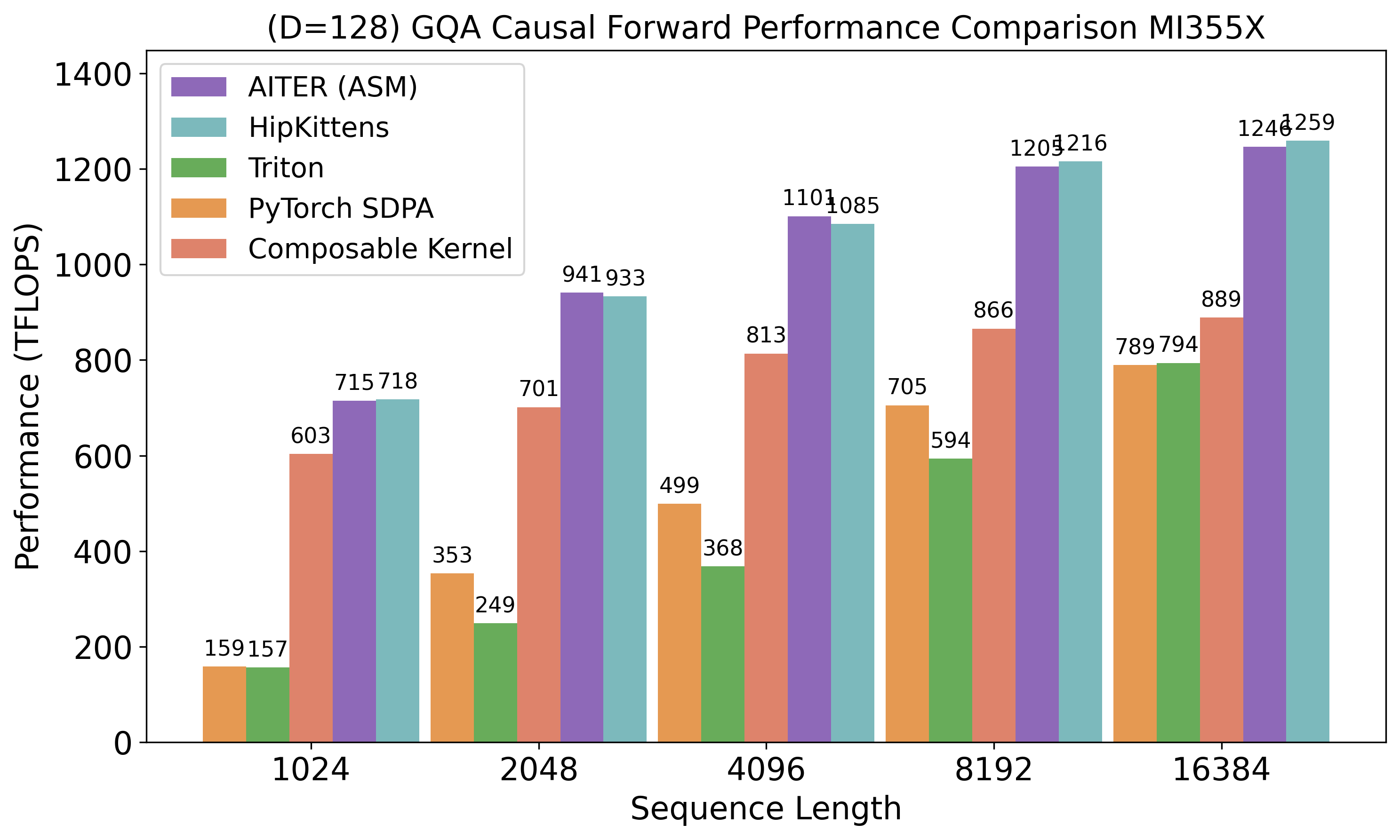}
\hfill
\includegraphics[width=0.48\linewidth]{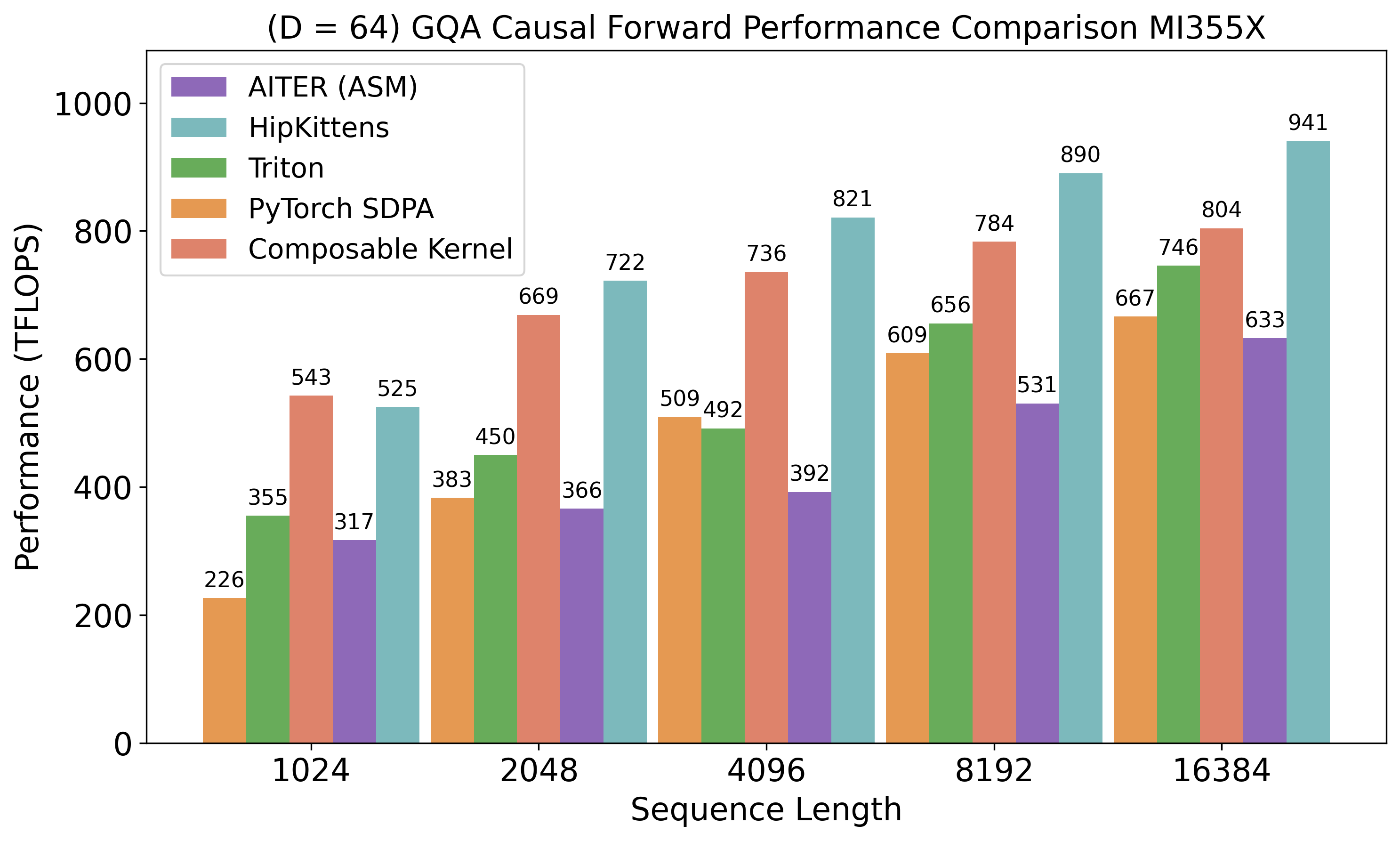}
\includegraphics[width=0.48\linewidth]{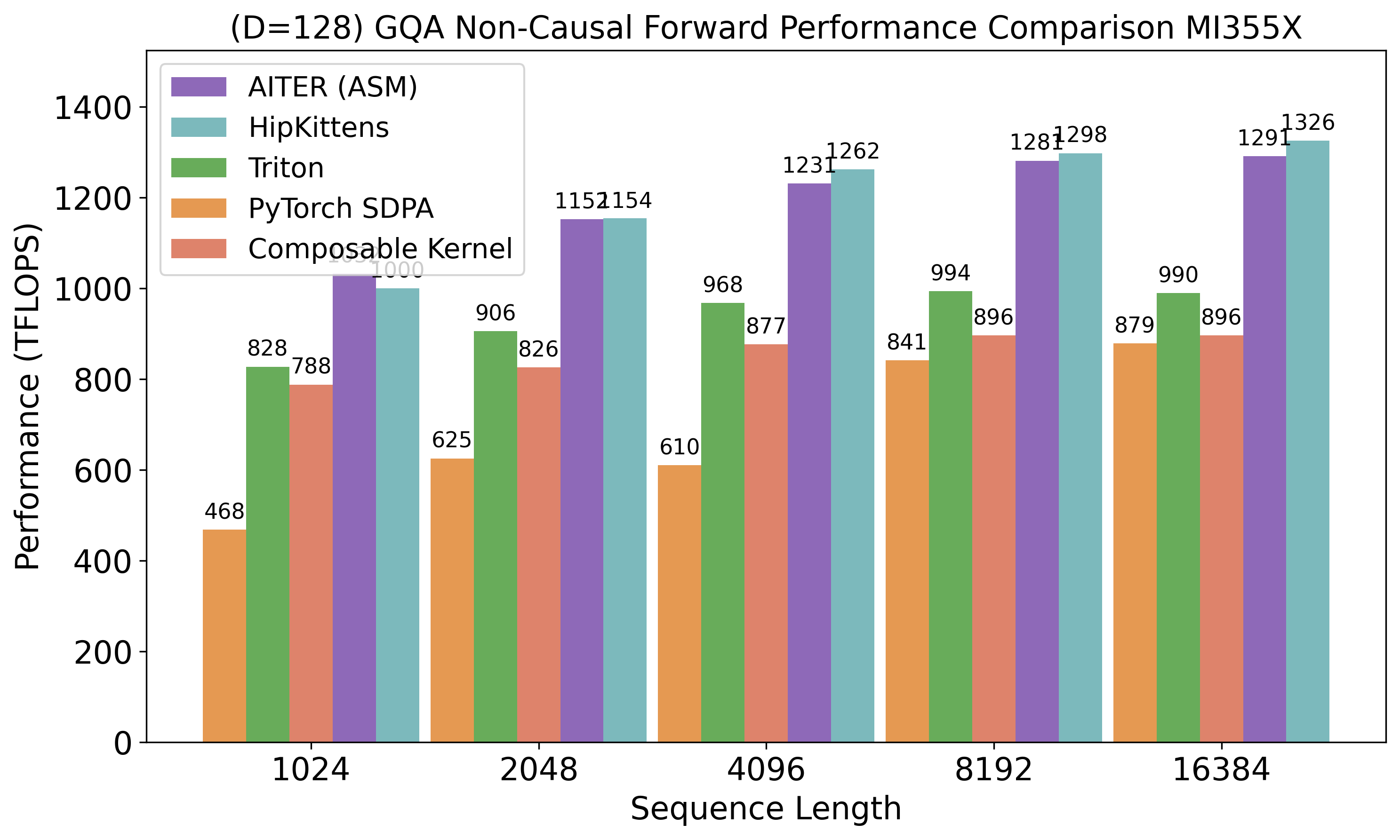}
\hfill
\includegraphics[width=0.48\linewidth]{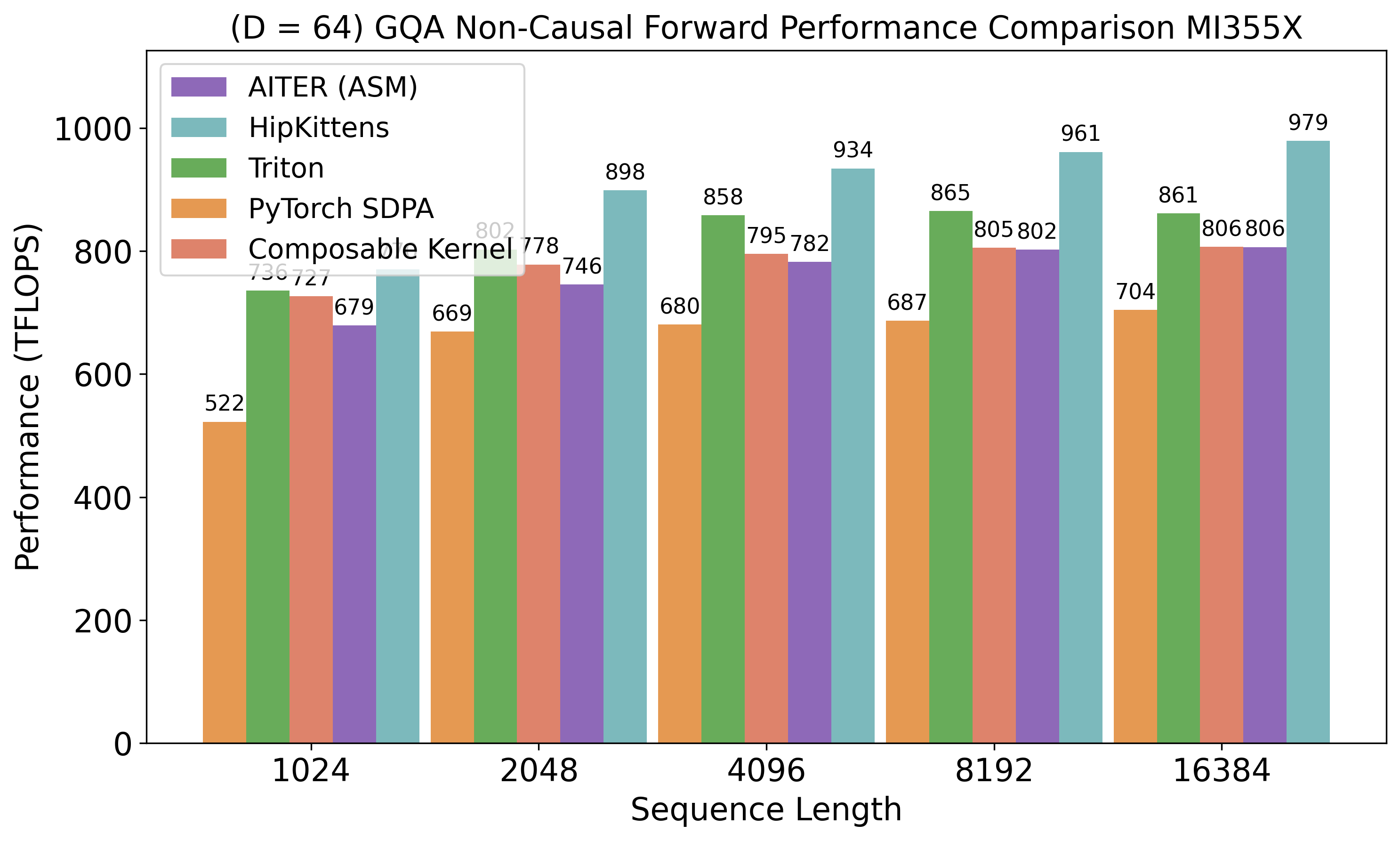}
\vspace{-2mm}
\caption{\textbf{Attention forwards.} We compare {\name} GQA and MHA (\Cref{fig:app_attention_mha_fwd}) to the strongest available baselines. We use batch $16$, query heads $64$, key value heads $8$,  head dim $64$ and $128$.}
\vspace{-2mm}
\label{fig:attention}
\end{figure}

%% file: figures/results/main_attention_bkwd.tex
\begin{figure}[t]
\centering
\includegraphics[width=0.48\linewidth]{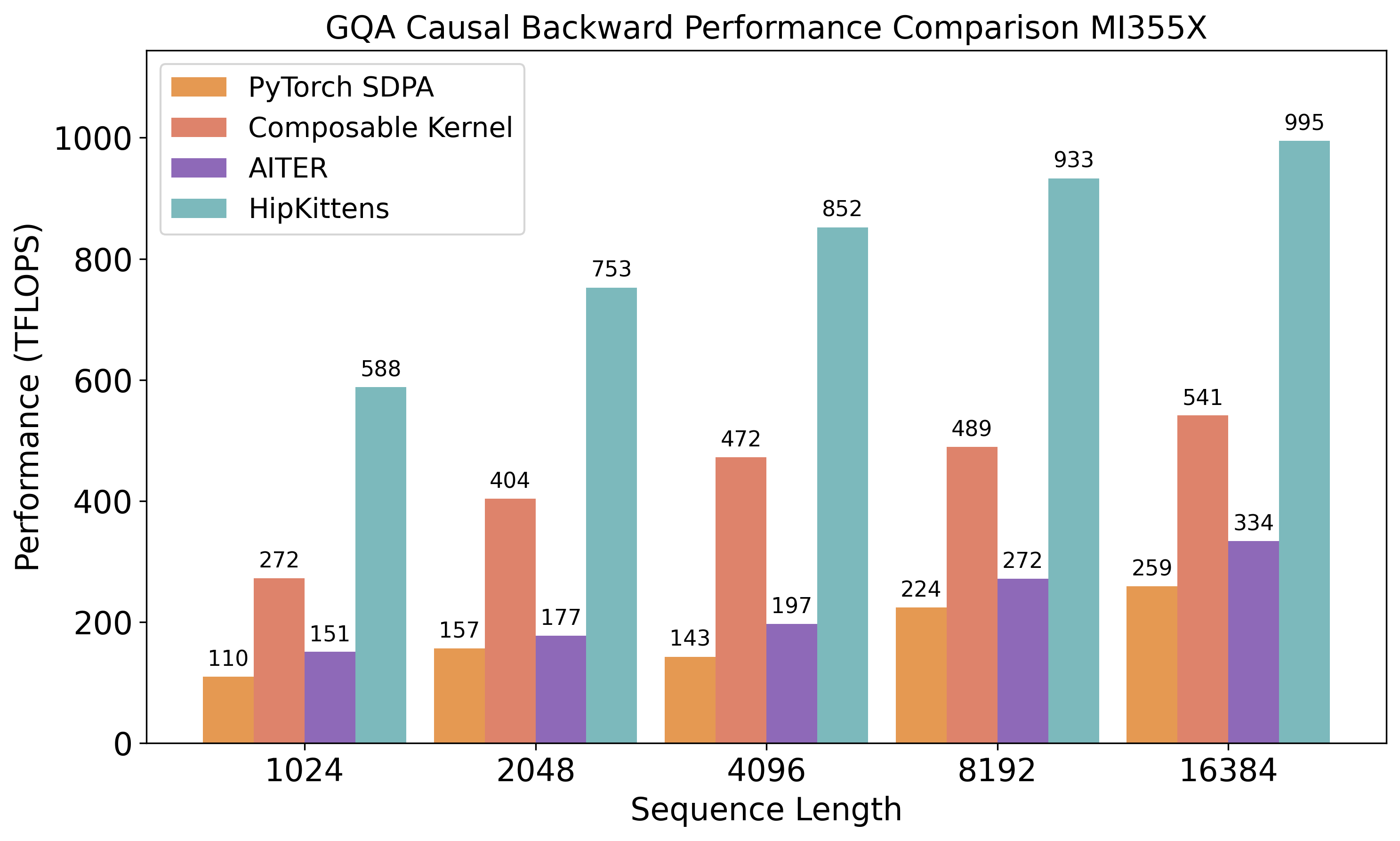}
\hfill
\includegraphics[width=0.48\linewidth]{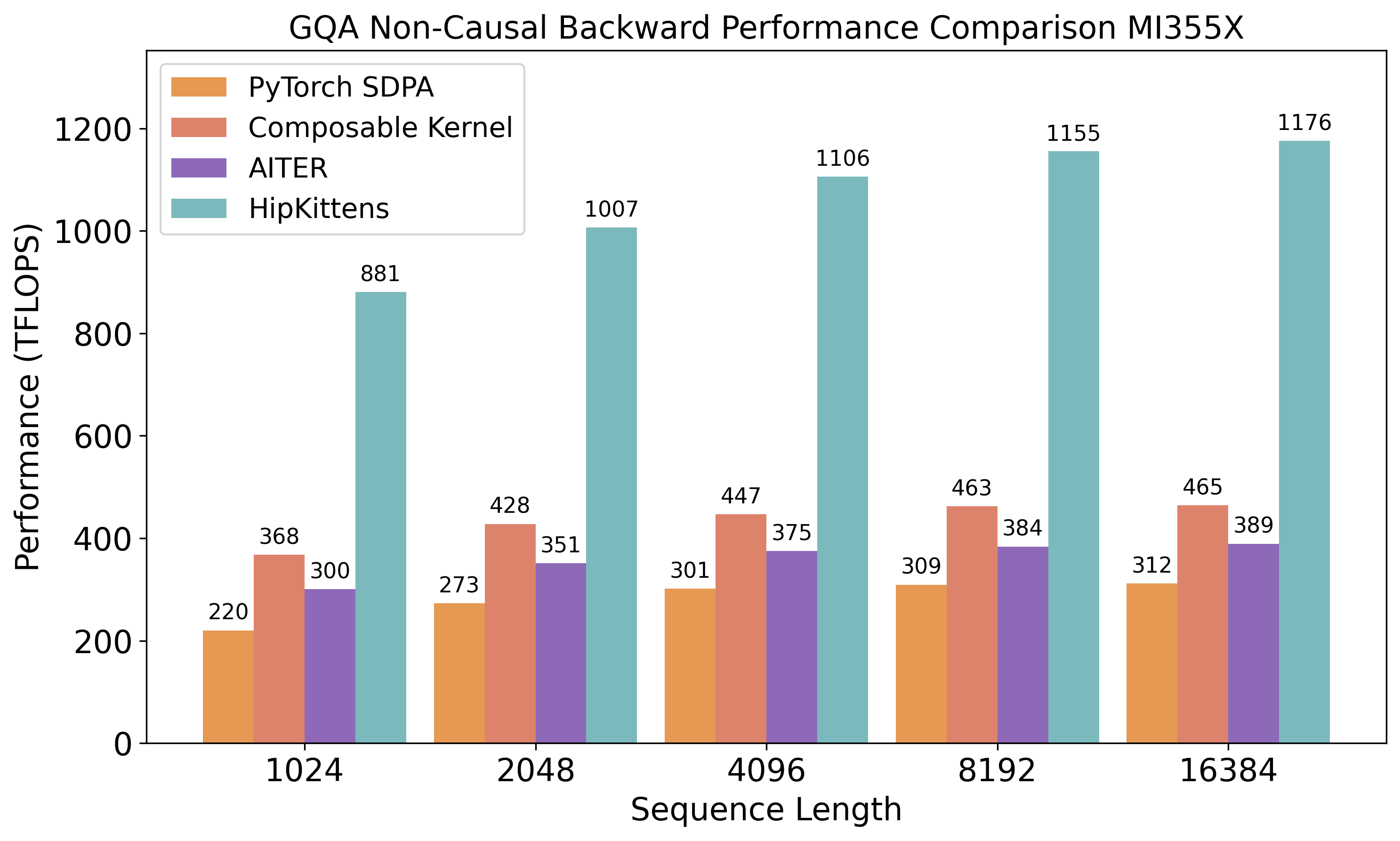}
\vspace{-2mm}
\caption{\textbf{Attention backwards.} We compare {\name} GQA and MHA (\Cref{fig:app_attention_mha_bwd}) to the strongest available baselines. We use batch $16$, query heads $64$, key value heads $8$, and head dim $128$. }
\label{fig:attention_bkwd}
\vspace{-2mm}
\end{figure}

%% file: figures/results/main_mem_bound_results.tex
\begin{figure}[t]
\centering
\includegraphics[width=0.48\linewidth]{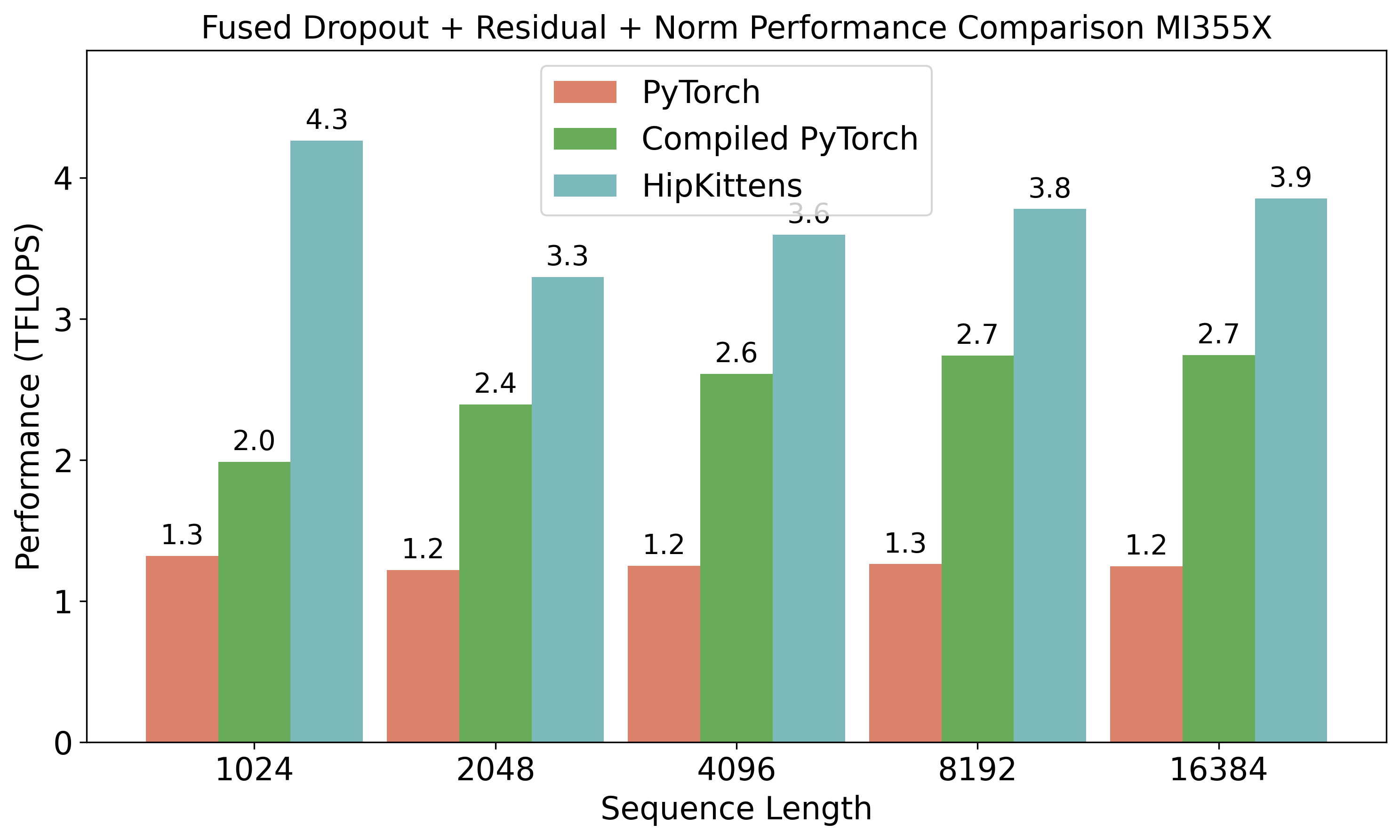}
\hfill
\includegraphics[width=0.48\linewidth]{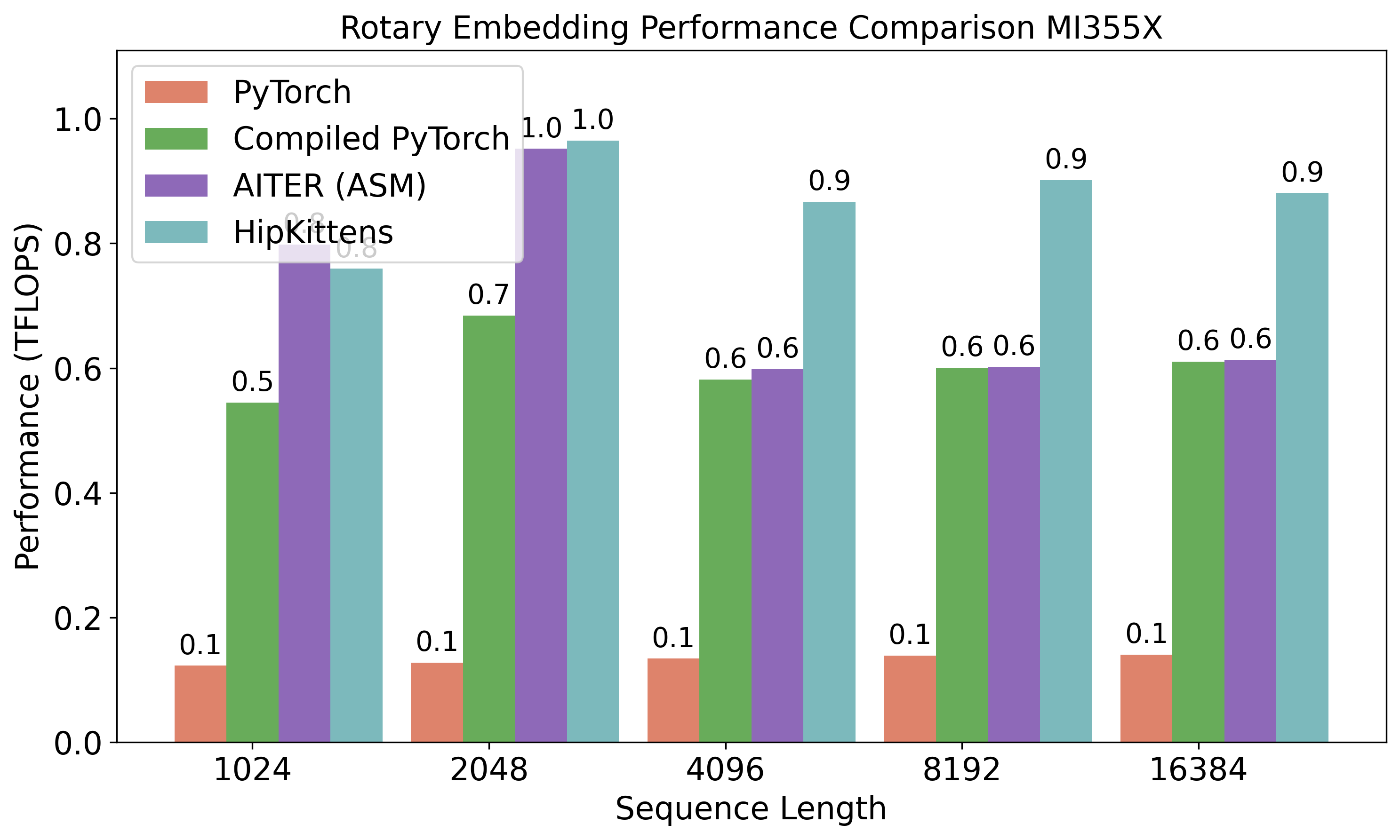}
\caption{\textbf{Memory bound.} We compare {\name} fused dropout-residual-layernorm and rotary kernels to the strongest available baselines at batch $16$, heads $16$, and head dim $128$.}
\vspace{-2mm}
\label{fig:mem_bound}
\end{figure}

%% file: sections/5-conclusion.tex
\vspace{-3mm}
\section{Discussion and conclusion}
\label{sec:conclusion}

Ideally, AI systems can leverage the full diversity of modern hardware. AMD CDNA4 GPUs offer state-of-the-art compute and memory bandwidth, but the ``CUDA moat'' limits adoption. While prior systems such as Triton aim for multi-silicon portability, our study shows that these compilers (and sometimes even C++ compilers) often fail to enable peak AMD performance. 

This work provides the first systematic analysis of the principles that enable high-performance AMD AI kernels and introduces {\name}, a minimal set of C++ embedded programming primitives that capture those principles. 
Though the abstractions and front-end \textit{interface}---tiles and PyTorch-inspired bulk operations over tiles---remains the same across NVIDIA and AMD, the instantiation of those abstractions---in terms of schedules, memory movement, and cache optimizations---differ due to fundamental hardware differences. We evaluate the ideas presented in {\shortname} by implementing a suite of representative AI workloads and find that we can achieve peak performance across them. 
By codifying the principles for AMD kernels into composable, open abstractions, these findings move the community closer to the long-standing vision of a universal software stack that performs well across diverse hardware platforms.

%% file: appendix/extended_related_work.tex
\appendix

\section{Terminology}
\label{sec:terminology}

Although both NVIDIA and AMD GPUs follow a broadly similar SIMT (single-instruction, multiple-thread) execution model, 
their hardware and software stacks use distinct terminology. The table above summarizes the key correspondences.

\begin{table}
\centering
\scriptsize
\begin{sc}
\begin{tabular}{lcc}
\toprule
\textbf{Concept} & \textbf{NVIDIA Term} & \textbf{AMD Term} \\
\midrule
Execution unit & Warp (32 threads) & Wave (64 threads) \\ 
\midrule
Thread block & Thread Block (CTA) & Workgroup \\
\midrule
Processor & Streaming Multiprocessor & Compute Unit \\
\toprule
Shared memory & Shared Memory (SMEM) & Local Data Share (LDS) \\
\midrule
Registers & Registers & Accumulator / Vector Registers (AGPRs/VGPRs) \\
\midrule
Global memory & HBM & HBM \\
\midrule
Cache & L2 (GPU-wide) & L2 Cache (chiplet-wide) + LLC (GPU-wide) \\
\toprule
Matrix compute & Tensor core & Matrix core \\
\midrule
Matrix instruction & WGMMA / WMMA / TCGEN05 & MFMA  \\
\midrule
Async memory ops & TMA & Buffer load to lds \\
\toprule
Compiler / toolchain & CUDA, NVCC & HIP, HIPCC \\
\bottomrule
\end{tabular}
\label{tab:terms}
\end{sc}
\end{table}

\clearpage
\section{Extended related work}
\label{app:extended_related_work}

\subsection{Libraries and frameworks for AI kernels} 
\paragraph{High-performance AMD kernel libraries} AMD provides AITER~\cite{amd2025aiter}, a library of high-performance kernels, but its fastest implementations are written in raw assembly. While effective, these kernels do not expose reusable abstractions and are brittle to extend across workloads. CUDA and HIP expose kernels at the level of individual threads, whereas ML workloads are built from larger reusable computational patterns that benefit from coarser abstractions. Several libraries and compilers have attempted to bridge this gap. Composable Kernel (like CUTLASS) uses deeply nested C++ templates, creating a level of complexity that makes it difficult to use and extend~\cite{nvidia2017cutlass, amd2025ck}.

\vspace{-2mm}
\paragraph{Compilers}  Compiler-based systems such as TVM and Triton~\cite{tvm, triton, helion2025} target a broader ML audience with higher-level Python-like DSLs. While more familiar to researchers, these frameworks both sacrifice fine-grained control over registers and synchronization and have been slow to support new hardware features (see \Cref{app:ecosystem}). As a result, on AMD, developers often resort to inline assembly in Triton kernels to recover performance~\cite{kim2022sayselephantscantrun}. Even on NVIDIA, where there has been relatively more investment into the compiler systems, C++-based DSLs provide up to $10\times$ performance improvements~\cite{spector2025tk}. 

There are also more recent compilers:
\begin{enumerate}
    \item \textbf{TileLang and Mojo} can compile to AMD via LLVM IR, but they lack abstractions for AMD’s architectural constraints (e.g., flexible tile sizing under register pressure, thread block scheduling, cache-aware grid ordering). Their evaluations on AMD are limited: TileLang reports a single attention kernel at 257 TFLOPs on AMD MI300X GPU. TileLang also depends on backend calls to CUTLASS/CK, while Mojo relies on compiler hints rather than reusable abstractions. Mojo's attention kernel for the MI300X also shows bank conflicts. Neither framework systematically supports AMD thus far. 
    \item \textbf{Linear Layouts} formalizes MMA/WGMMA/MFMA layouts as linear maps and implements automatic, optimized conversions between them in Triton’s backend (via warp shuffles and swizzled shared memory), with measured speedups. However, it does not define abstractions for thread-block or grid-level scheduling, and its evaluations do not demonstrate kernels that mix tensor core shapes or discuss the different phase orderings of memory instructions (as in \Cref{sec:hip-warp}).
\end{enumerate}

\paragraph{Programming frameworks for peak performance}  Recently, a wave of DSLs has proposed C++ embedded tile-based primitives for AI kernels including {\tk} (\shorttk)~\cite{spector2025tk} and successors (TileLang~\cite{wang2025tilelang}, CuTe~\cite{nvidia2024cute}, Linear Layouts~\cite{zhou2025linearlayoutsrobustcode}). These approaches demonstrate that small, opinionated sets of abstractions can yield both simplicity and high performance. However, none provides a comprehensive set of AMD kernel abstractions:  {\tk} and CuTe support and validate only on NVIDIA hardware, with kernel templates such as producer–consumer scheduling tied to NVIDIA-specific features. In contrast, our work identifies a minimal and principled set of abstractions that suffice for performant AMD kernels across warp, block, and grid levels. We demonstrate their sufficiency by implementing end-to-end kernels across diverse workloads, including attention backward, which requires mixing tensor core shapes.

\subsection{AMD software ecosystem is brittle} 
\label{app:ecosystem}
Many AMD libraries are forks of NVIDIA libraries. Given the hardware differences, this risks sub-optimal performance. We also find that despite the investment into AMD software, the current ecosystem is brittle, motivating {\shortname}:
\begin{itemize}[itemsep=0.1pt,topsep=0pt,leftmargin=*]
\item \textbf{PyTorch kernels:} The built-in scaled dot product attention (SDPA) backend achieves just 259 TFLOPS on AMD MI355X GPUs for Llama GQA backwards (as of October 2025 on ROCm 7.0.0). Attention is a workhorse operation in modern AI workloads, and this performance gap highlights limited backend maturity.
\item \textbf{Assembly kernels:} AITER~\cite{amd2025aiter} includes high-performance kernels, but its fastest implementations are written directly in raw assembly. This approach is difficult to scale across the breadth of AI workloads; we can see the lack of full-fledged kernel support for GQA backwards on the AMD MI355X where AITER achieves just 272/384 TFLOPS at a sequence length of 8192 for causal and non-causal respectively. 
\item \textbf{Triton kernels:} The Triton compiler on AMD struggles with register lifetime tracking and lowering memory accesses to the most performant intrinsics. For example, it may fail to reclaim registers~\href{https://github.com/ROCm/tritonBLAS/blob/98c7d7d43ce831b30240389687f0417ff4b8fcc4/include/tritonblas/internal/streamk_matmul.py#L206-L265}{(code
 snippet)} or lower vectorized loads~\href{https://github.com/ROCm/iris/blob/ec2dd919536ff56ce731020ac879a32430983291/iris/iris.py#L1456-L1462}{(code
 snippet)}. 
 Torch-compiled kernels can deliver competitive performance on memory-bound workloads (\Cref{fig:mem_bound}), but the optimizations are black-box and may miss optimal intrinsics. For example, a compiled LayerNorm kernel on Llama-like dimensions exhibits a 23\% lower L2 hit rate than our {\shortname} kernel.
 The time between new CDNA/PTX features and their integration into compilers is also slow; for instance as of September 2025, buffer loads are not the default for Triton loads/stores on AMD (\href{https://github.com/triton-lang/triton/pull/8013}{code snippet}).
\end{itemize}

 This ecosystem motivates {\shortname}, which is designed to help simplify and accelerate high performance AMD kernel development.
\vspace{2mm}

%% file: appendix/evaluations.tex
\clearpage
\section{Extended analysis}
\label{app:extended_results}

This section provides details on our setup and supplemental results. 

\paragraph{Setup details.} AMD provides multiple docker containers at \url{https://hub.docker.com/u/rocm}. We use the recent AMD provided Docker containers to benchmark kernels:  \url{docker.io/rocm/7.0-preview:rocm7.0\_preview\_pytorch\_training
\_mi35x\_beta} on MI350X/MI355X and  \url{docker.io/rocm/pytorch} on the MI300X/MI325X. A sample command to launch the container is provided below:

\begin{figure}[h!]
\begin{lstlisting}[language=CUDACPP, basicstyle=\scriptsize\ttfamily]
podman run -it \
    --ipc=host \
    --network=host \
    --privileged \
    --cap-add=CAP_SYS_ADMIN \
    --cap-add=SYS_PTRACE \
    --security-opt seccomp=unconfined \
    --device=/dev/kfd \
    --device=/dev/dri \
    -v $(pwd):/workdir/ \
    -e USE_FASTSAFETENSOR=1 \
    -e SAFETENSORS_FAST_GPU=1 \
    rocm/7.0-preview:rocm7.0_preview_pytorch_training_mi35x_beta \
    bash
\end{lstlisting}
\end{figure}

\vspace{-2mm}
\paragraph{Baselines.} For the AITER baselines, we use~\Cref{lst_install_aiter}. If AITER does not automatically come with the Docker, we install from source. For the Composable Kernel baselines, we use the installation process and kernels indicated in~\Cref{lst_install_ck}. For the PyTorch baselines, we use~\Cref{lst_install_torch}. For the HipBLASLT baselines, we use the  command from~\Cref{lst_install_hipblaslt} within the AMD provided Dockers.

\begin{figure}[h!] 
\begin{lstlisting}[language=CUDACPP, basicstyle=\scriptsize\ttfamily]
// Attention
import aiter
out_aiter, softmax_lse = aiter.flash_attn_func(Q_aiter, K_aiter, V_aiter, causal=causal, return_lse=True, deterministic=False)
out_aiter.backward(dO_aiter)
// GEMM
from aiter.tuned_gemm import tgemm
C_aiter = tgemm.mm(A, B, None, None, None)
\end{lstlisting}
\caption{AITER benchmarking.}
\label{lst_install_aiter}
\end{figure}

\begin{figure}[t!] 
\begin{lstlisting}[language=CUDACPP, basicstyle=\scriptsize\ttfamily]
out_pt = torch.nn.functional.scaled_dot_product_attention(q_pt, k_pt, v_pt, attn_mask=None, dropout_p=0.0, is_causal=causal)
out_pt.backward(dO_bhnd)
\end{lstlisting}
\caption{PyTorch benchmarking.}
\label{lst_install_torch}
\end{figure}

\begin{figure}[t!] 
\begin{lstlisting}[language=CUDACPP, basicstyle=\scriptsize\ttfamily]
// Build
[~] git clone https://github.com/rocm/composable_kernel
[~] cd composable_kernel
[~/composable_kernel] mkdir build && cd build
[~/composable_kernel/build] ../script/cmake-ck-dev.sh .. gfx950 -G Ninja
[~/composable_kernel/build] ninja tile_example_gemm_basic
[~/composable_kernel/build] ninja tile_example_fmha_fwd
[~/composable_kernel/build] ninja tile_example_fmha_bwd

// Run
./bin/tile_example_gemm_basic -prec=fp8 -m=1024 -n=1024 -k=1024 -warmup=500 -repeat=100 -v=1 
./bin/tile_example_fmha_fwd -prec=bf16 -b=16 -h=16 -d=128 -s=1024 -mask=1 -warmup=500 -repeat=100 -kname=1
./bin/tile_example_fmha_bwd -prec=bf16 -b=16 -h=16 -d=128 -s=1024 -mask=1 -warmup=500 -repeat=100 -kname=1
\end{lstlisting}
\caption{CK benchmarking. For each dimension of the GEMM, we report the best performance across the CK {tile example streamk gemm basic}, {tile example gemm basic}, and {tile example gemm universal}.}
\label{lst_install_ck}
\end{figure}

\begin{figure}[t!] 
\begin{lstlisting}[language=CUDACPP, basicstyle=\scriptsize\ttfamily]
// bf16
hipblaslt-bench --batch_count 1 --a_type bf16_r --b_type bf16_r --c_type bf16_r --d_type bf16_r --rotating 512 --iters 100 --cold_iters 500 -m 1024 -n 1024 -k 1024

// fp8
hipblaslt-bench --api_method c --stride_a 0 --stride_b 0 --stride_c 0 --stride_d 0 --alpha 1.000000 --beta 0.000000 --transA T --transB N --batch_count 1 --scaleA 1 --scaleB 1 --a_type f8_r --b_type f8_r --c_type bf16_r --d_type bf16_r --scale_type f32_r --bias_type f32_r --compute_type f32_r --rotating 4 --iters 100 --cold_iters 500 -m 8192 -n 8192 -k 8192 --lda 8192 --ldb 8192 --ldc 8192 --ldd 8192 --initialization norm_dist

// fp6. after insstalling from source
./clients/hipblaslt-bench --api_method c -m 1024 -n 1024 -k 1024 --alpha 1 --beta 0 --transA T --transB N --batch_count 1 --scaleA 3 --scaleB 3 --a_type f6_r --b_type f6_r --c_type f16_r --d_type f16_r --compute_type f32_r --rotating 0 --cold_iters 500 --iters 100

\end{lstlisting}
\caption{HipBLASLT benchmarking.}
\label{lst_install_hipblaslt}
\end{figure}

\clearpage
\subsection{{\name} kernels}
We include the remaining kernel plots from~\Cref{sec:sec4_results}. 
We include BF16 GEMM results for the MI325X and MI350X GPUs in~\Cref{fig:app_matmul}. 
We include MHA results on the MI355X GPUs in~\Cref{fig:app_attention_mha_fwd} for the forwards pass and \Cref{fig:app_attention_mha_bwd} for the backwards pass.

\input{figures/results/appendix_gemm_results}

\input{figures/results/appendix_attention_results}

\input{figures/results/appendix_d64_attention}

\clearpage
\subsection{Grid schedules.}

In \Cref{sec:hip-grid}, we include~\Cref{tab:l2_table_14592} to discuss the chiplet swizzling strategy to optimize L2 and LLC reuse. In \Cref{fig:14592-schedules} we provide the corresponding grid order visualization for the $14592$ dimension GEMM.
\input{figures/cache_micros/14592_schedules}

\clearpage
\subsection{{\tk} performance}
We benchmark {\shorttk}~\cite{spector2025tk} and CuBLASLT kernels on inputs drawn from $\mathcal{N}(0, 1)$ with 500 iterations of warmup and 100 iterations of measured runs, remaining consistent with our protocol on the AMD GPUs. Results are shown in~\Cref{fig:tk_perf}. 
\footnote{We evaluate both on NVIDIA H100 and NVIDIA B200 GPUs. We generally observe that the NVIDIA kernel performance degrades as the number of iterations increases and we note that \citet{spector2025tk} reports using fewer iterations.} We include this to highlight how the {\shorttk} philosophy, which we extend in this work, leads to performant NVIDIA kernels.

\input{figures/results/blackwell}

%% file: figures/results/appendix_gemm_results.tex
\begin{figure}[h!]
\centering
\includegraphics[width=\linewidth,height=0.25\textheight,keepaspectratio]{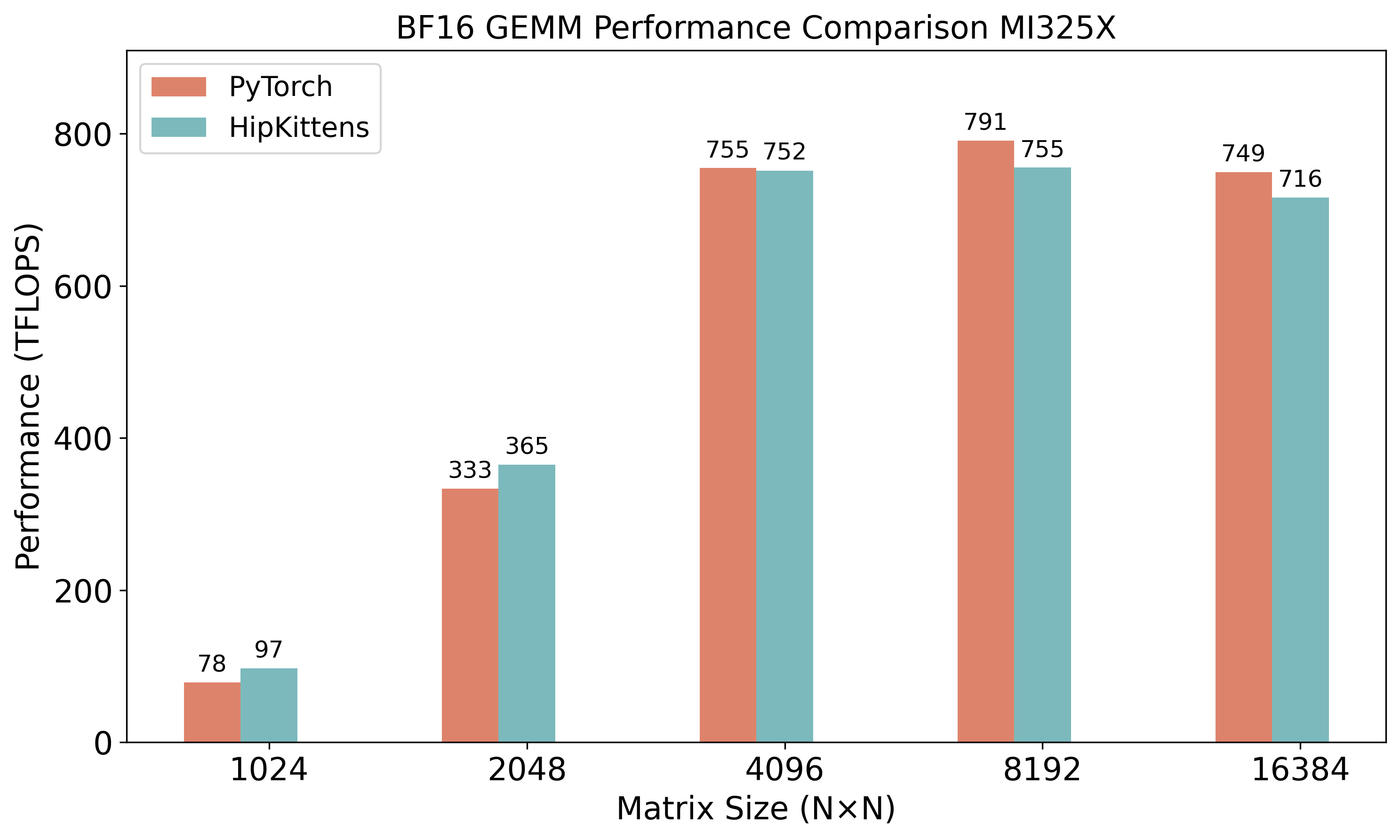}
\includegraphics[width=\linewidth,height=0.25\textheight,keepaspectratio]{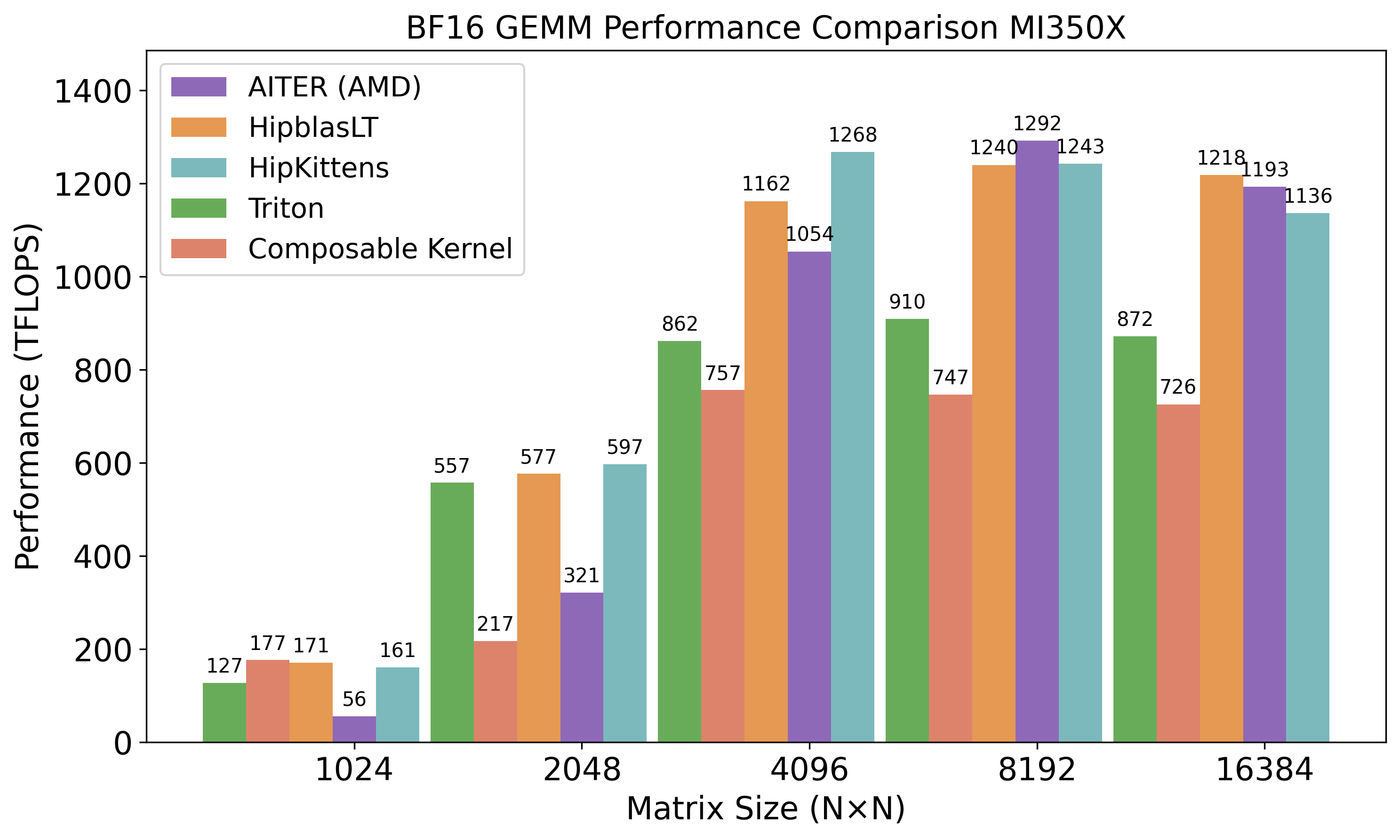}
\caption{\textbf{BF16 GEMM.} We compare {\name} to the strongest available baselines on the MI325X and MI350X. For these kernels, we use dimensions $M=N=K$. }
\label{fig:app_matmul}
\end{figure}

%% file: figures/results/appendix_attention_results.tex
\begin{figure}[h!]
\centering
\includegraphics[width=\linewidth,height=0.25\textheight,keepaspectratio]{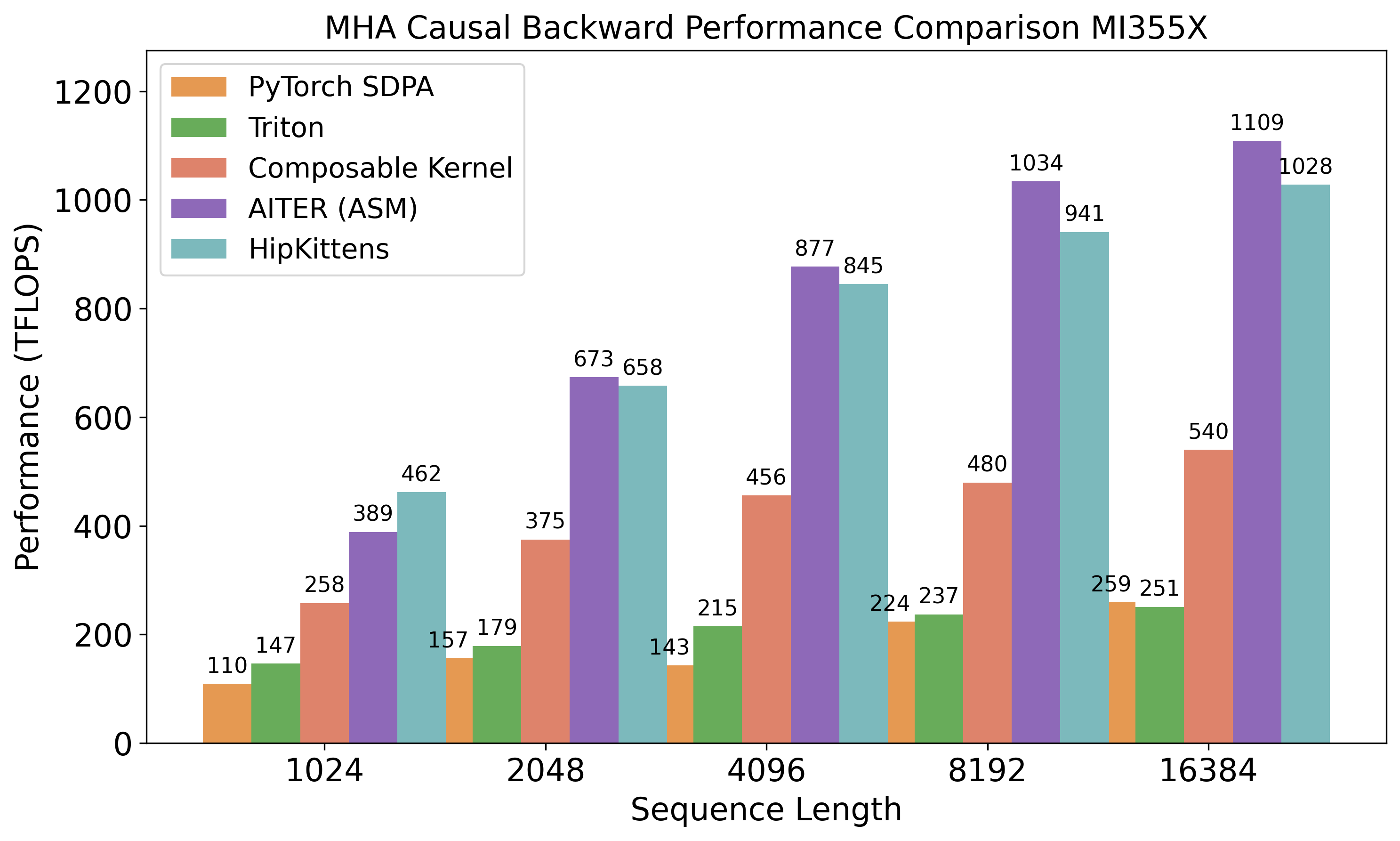}
\includegraphics[width=\linewidth,height=0.25\textheight,keepaspectratio]{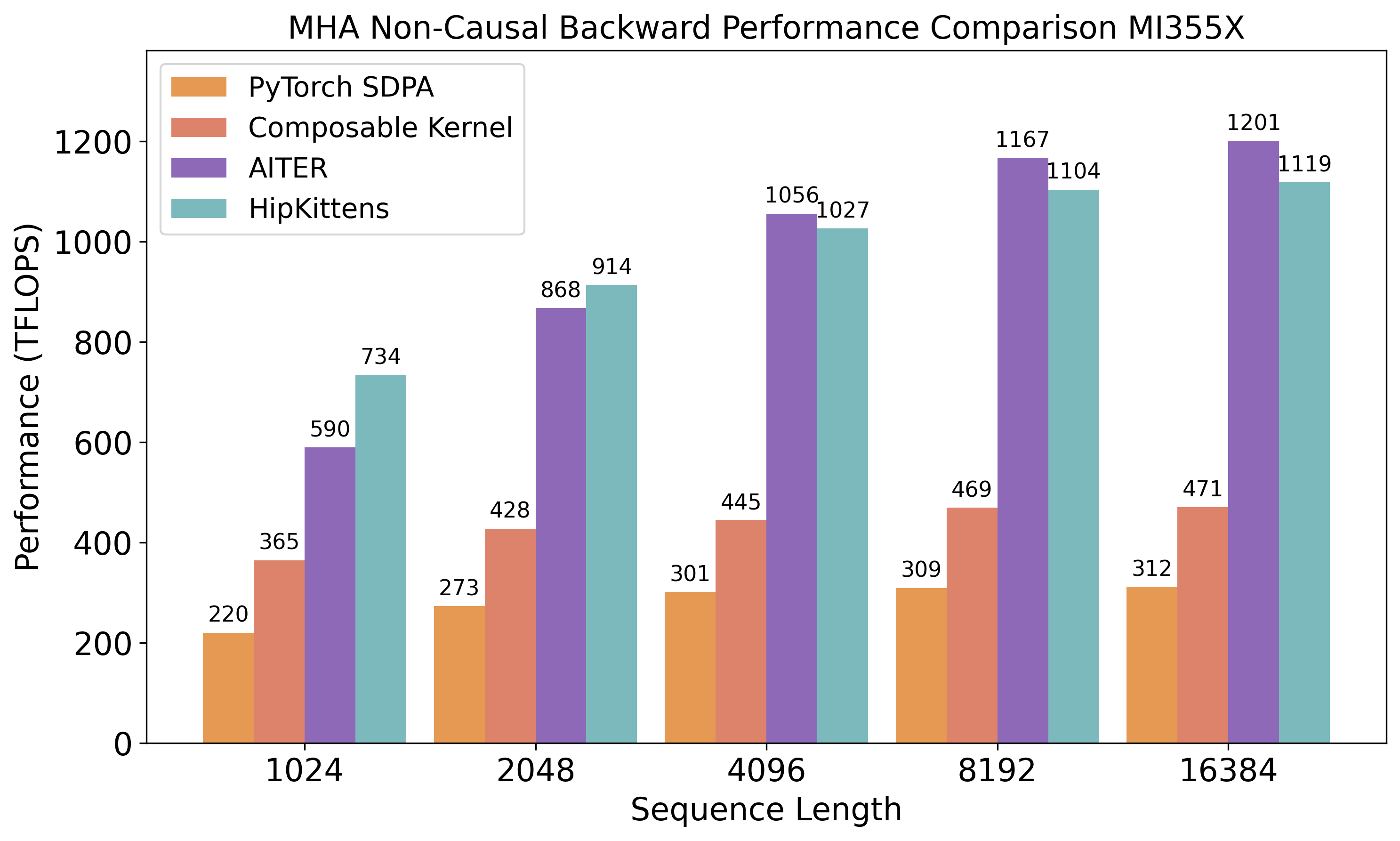}
\caption{\textbf{Attention backwards.} MI355X results for causal and non-causal attention. We use batch size $16$, heads $16$, head dim $128$.}
\label{fig:app_attention_mha_bwd}
\end{figure}

\begin{figure}[h!]
\centering
\includegraphics[width=\linewidth,height=0.25\textheight,keepaspectratio]{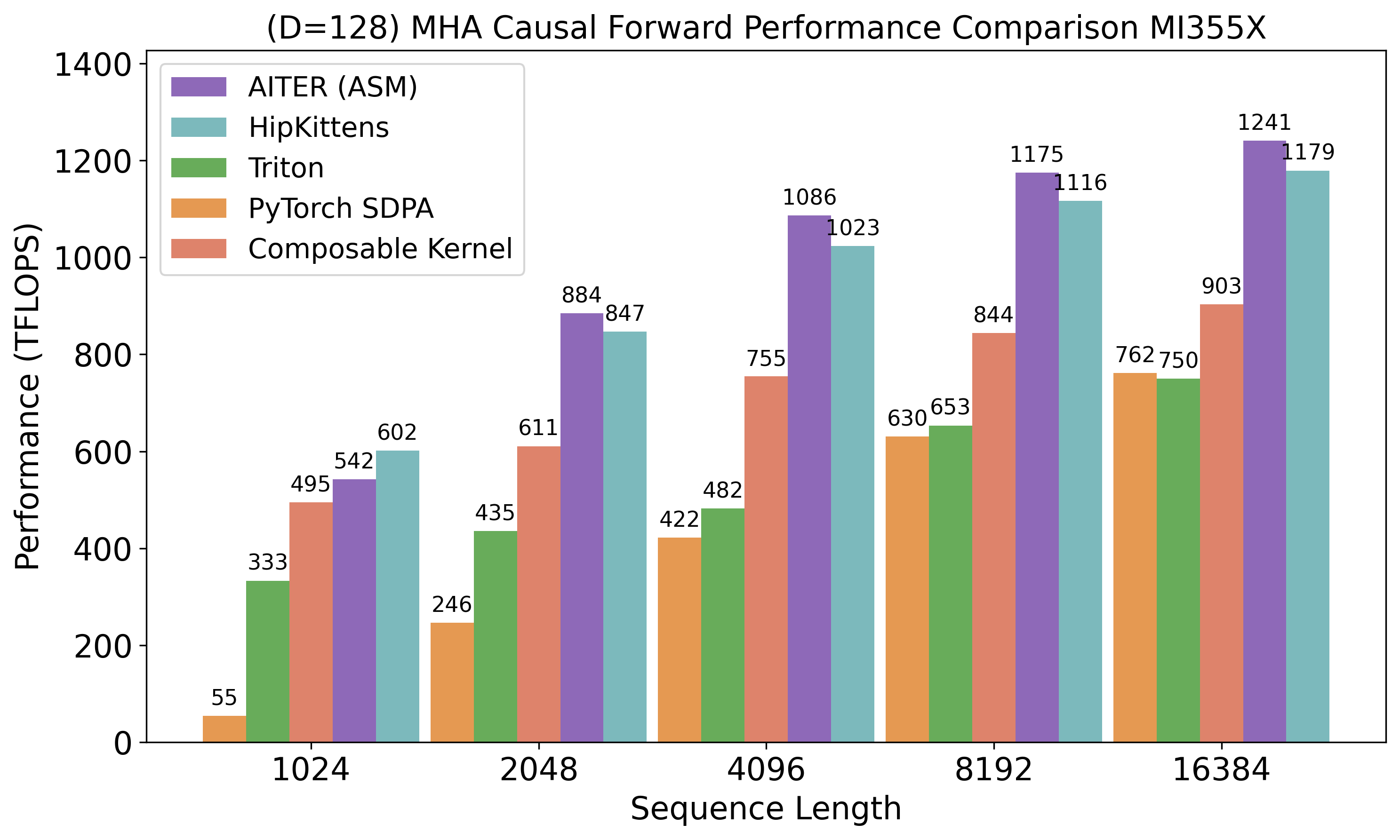}
\includegraphics[width=\linewidth,height=0.25\textheight,keepaspectratio]{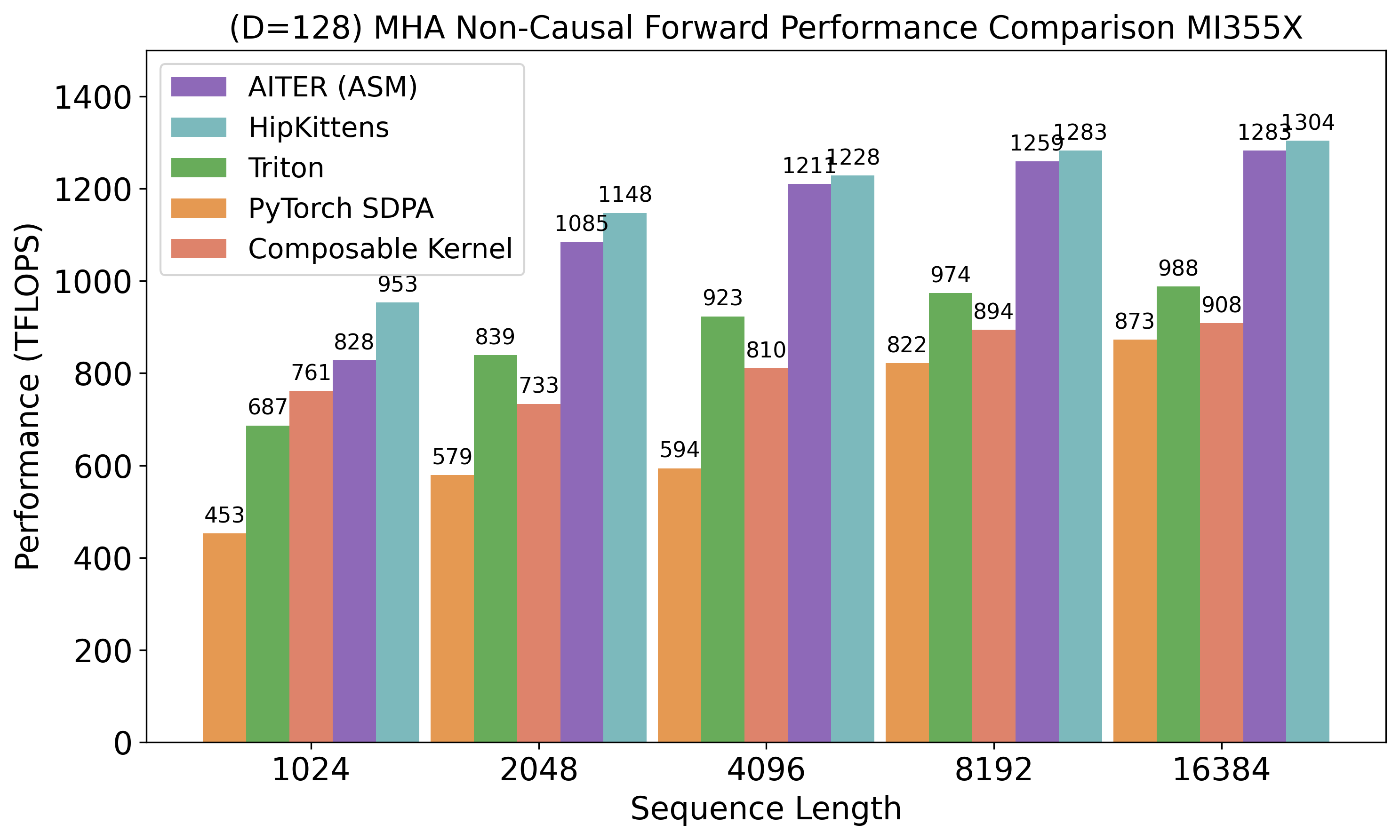}
\caption{\textbf{Attention forwards.}  MI355X results for causal and non-causal attention. We use batch size $16$, heads $16$, head dim $128$.}
\label{fig:app_attention_mha_fwd}
\end{figure}

%% file: figures/results/appendix_d64_attention.tex
\begin{figure}[h!]
\centering
\includegraphics[width=\linewidth,height=0.25\textheight,keepaspectratio]{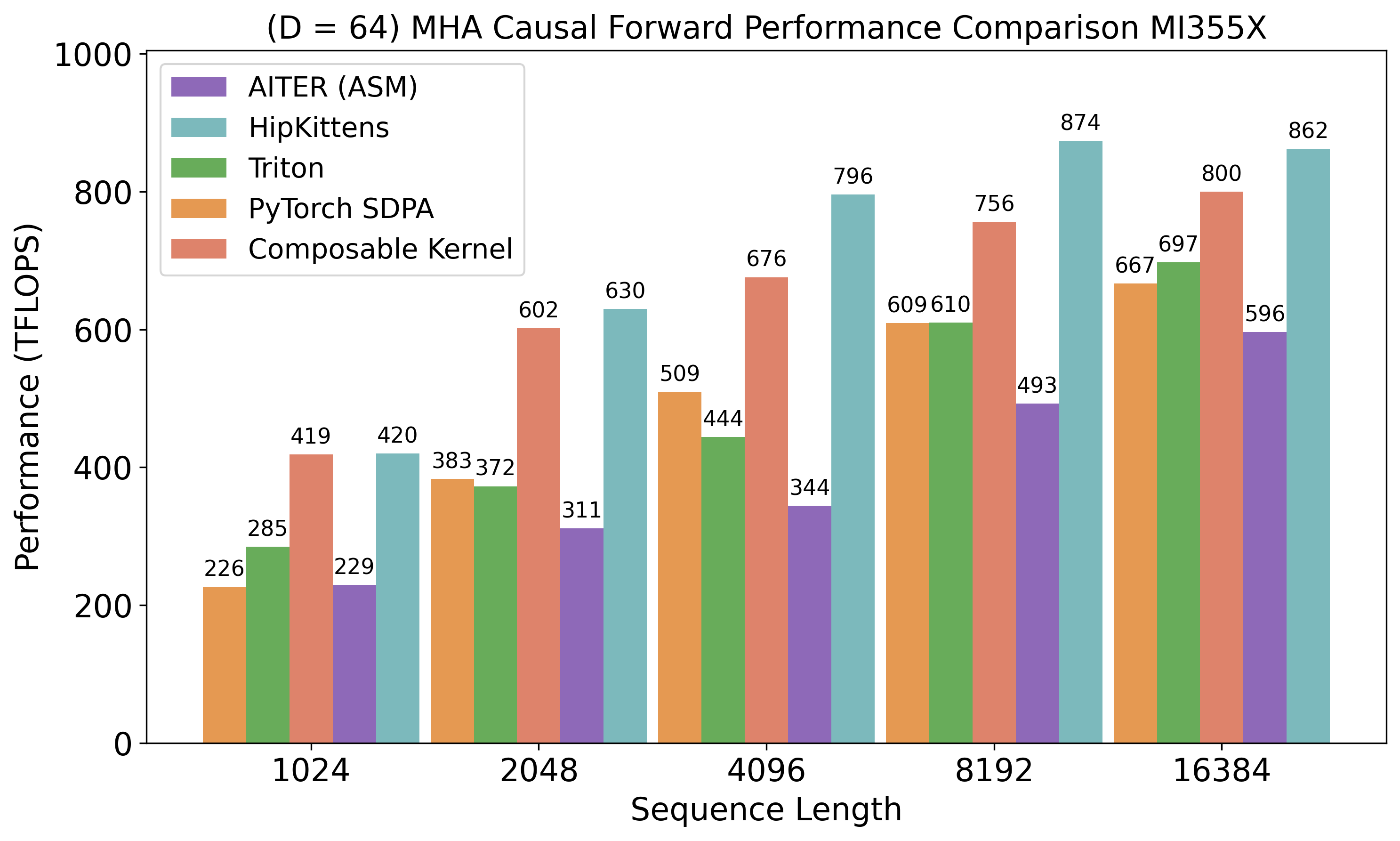}
\includegraphics[width=\linewidth,height=0.25\textheight,keepaspectratio]{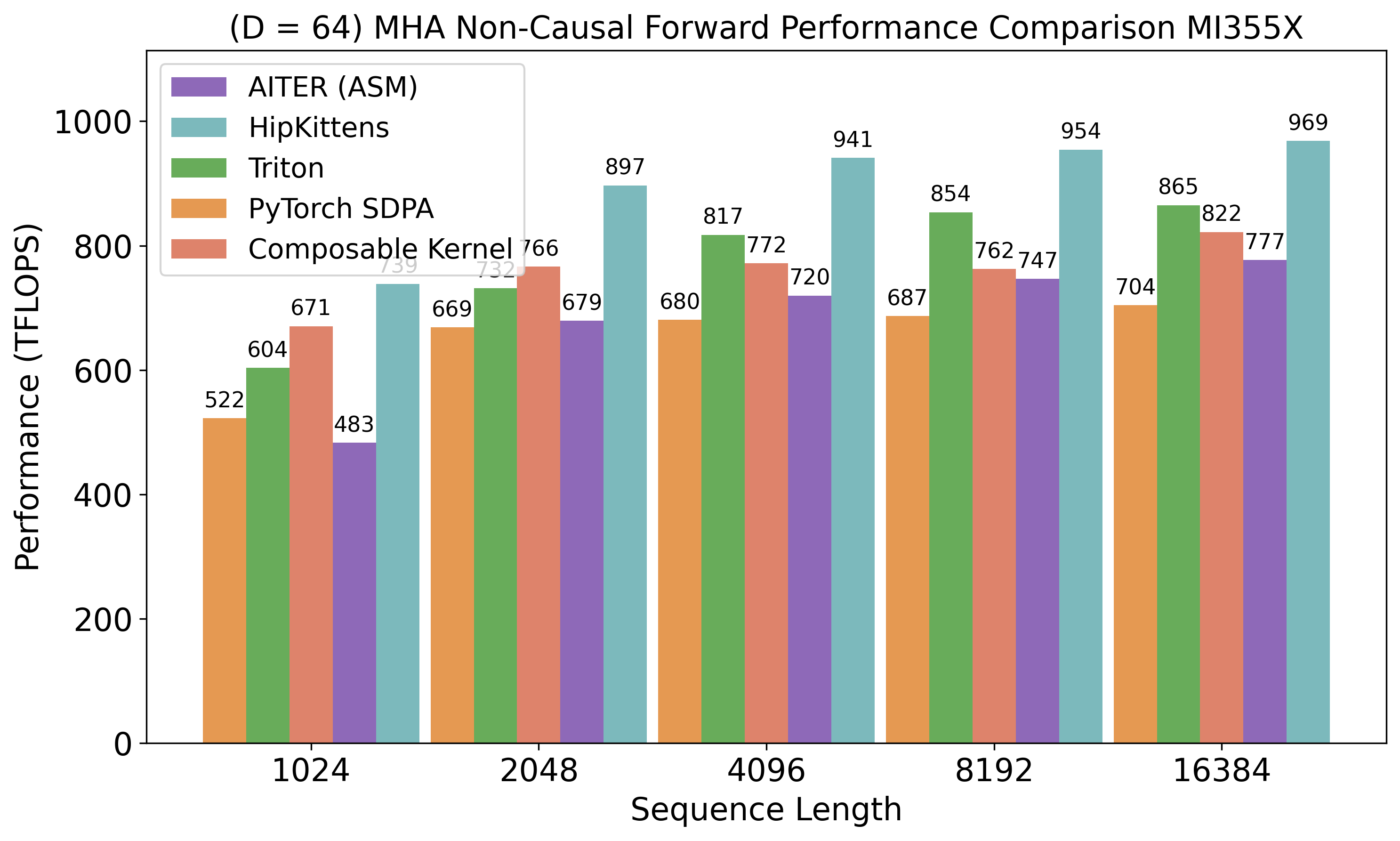}
\caption{\textbf{Attention forwards.}  MI355X results for causal and non-causal attention. We use batch size $16$, heads $16$, head dim $64$.}
\label{fig:app_attention_mha_d64_fwd}
\end{figure}

%% file: figures/cache_micros/14592_schedules.tex
\begin{figure}[h!]
  \centering
  \begin{subfigure}[t]{0.32\linewidth}\centering
    \includegraphics[width=\linewidth,height=0.25\textheight,keepaspectratio]{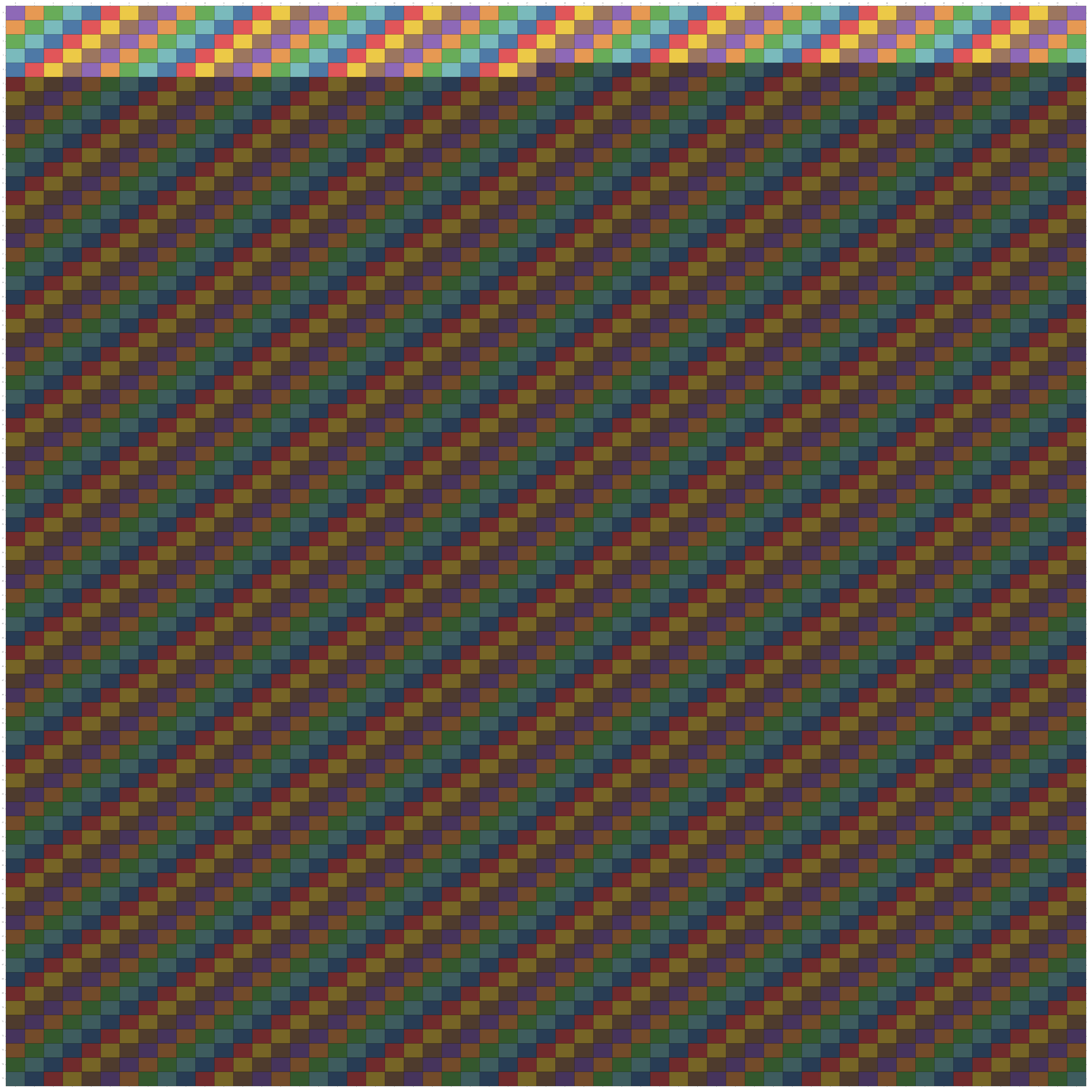}
    \subcaption{}\label{fig:14592_default}
  \end{subfigure}\hfill
  \begin{subfigure}[t]{0.32\linewidth}\centering
    \includegraphics[width=\linewidth,height=0.25\textheight,keepaspectratio]{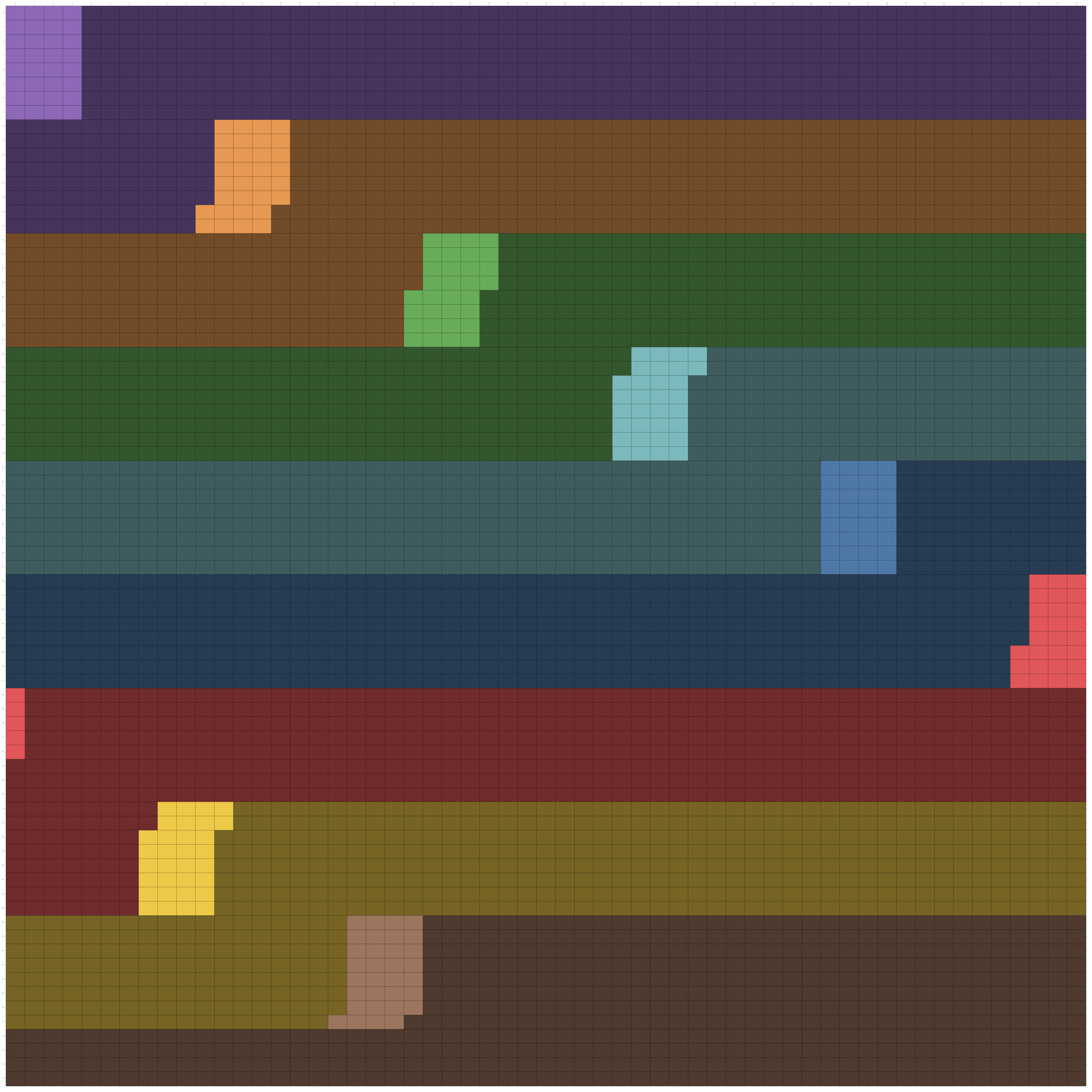}
    \subcaption{}\label{fig:14592_w8_c0}
  \end{subfigure}\hfill
  \begin{subfigure}[t]{0.32\linewidth}\centering
    \includegraphics[width=\linewidth,height=0.25\textheight,keepaspectratio]{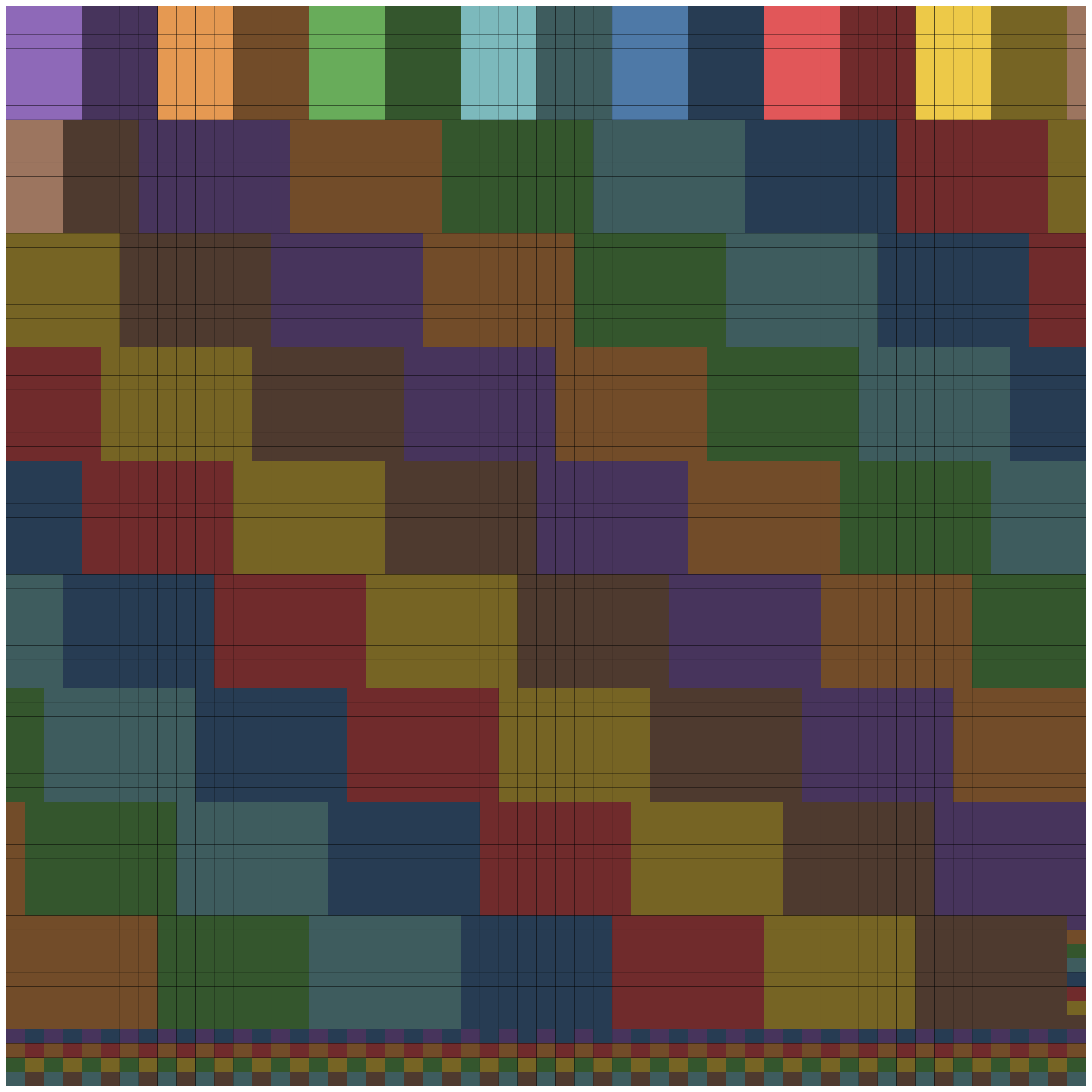}
    \subcaption{}\label{fig:14592_w8_c64}
  \end{subfigure}
\caption{Visualization of three different grid schedules for the output D matrix of a M=N=K=14592 BF16 GEMM. Color represents XCD assignment. Highlighted is the first timestep of thread blocks scheduled across the device (256 CUs). Schedule \ref{fig:14592_default} assigns blocks to the grid according to block ID. Schedules \ref{fig:14592_w8_c0} and \ref{fig:14592_w8_c64} apply algorithm \ref{alg:xcd-swizzle} with different chunk sizes and the same window size. Table \ref{tab:l2_table_14592} showcases the performance for each of these schedules. This GEMM setting is especially sensitive to these optimizations due to the default schedule resulting in worst case L2 reuse and the large memory footprint making LLC reuse even more important.}

  \label{fig:14592-schedules}
\end{figure}

%% file: figures/results/blackwell.tex
\begin{figure}[h!]
\centering
\includegraphics[width=0.48\linewidth]{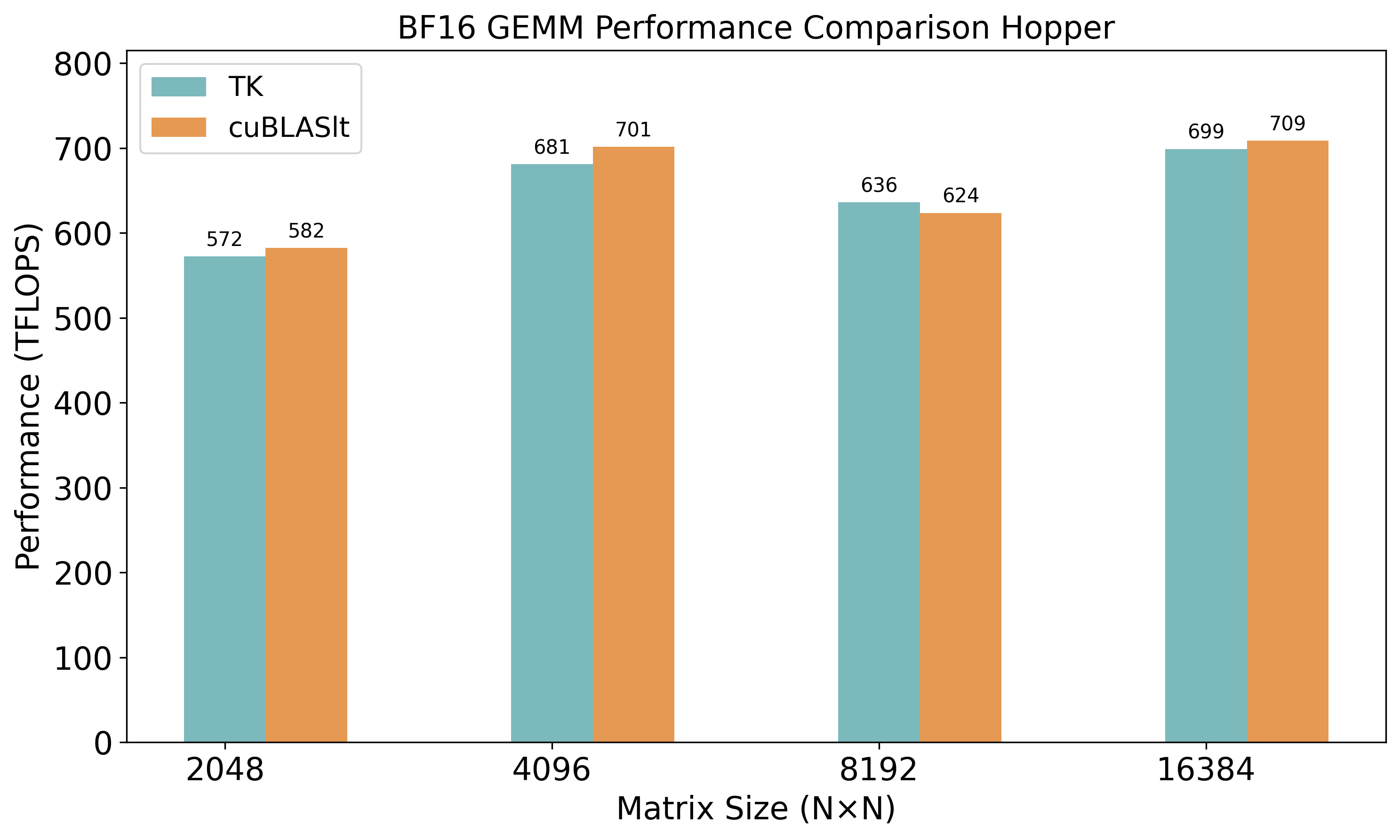}
\hfill
\includegraphics[width=0.48\linewidth]{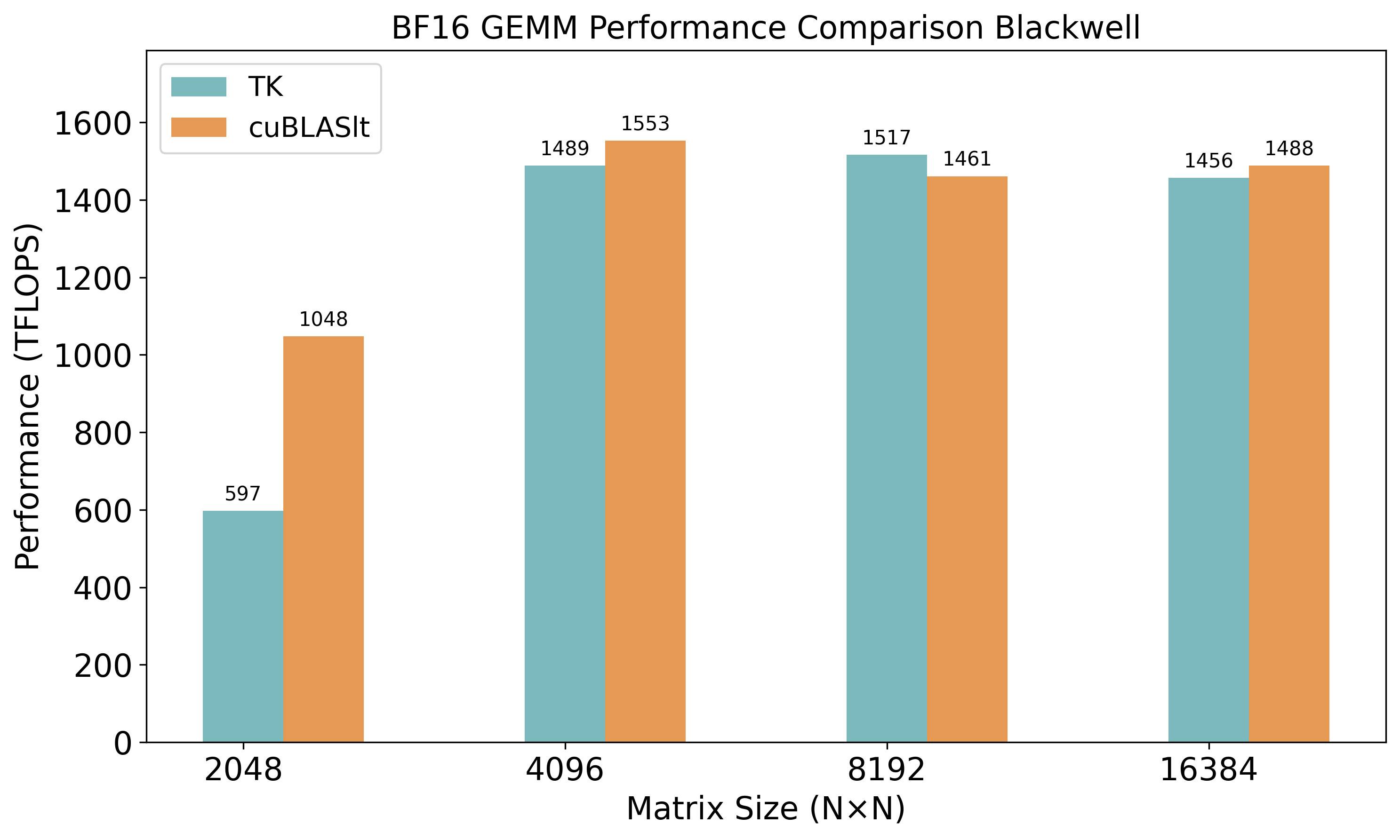}
\caption{\textbf{{\tk} performance.} We compare {\shorttk} to NVIDIA CublasLT BF16 GEMM performance, finding that {\shorttk} kernels offer competitive performance, despite the {\shorttk} kernels being released $\approx$ 8-12 months ago.}
\label{fig:tk_perf}
\end{figure}

%% file: appendix/implementations.tex
\clearpage
\section{Library implementation details}
\label{app:implementation}

\subsection{Shared and Register Memory}
\label{app:layouts}

\paragraph {Register Layouts.} The layout of a register tile dictates whether threads hold elements in a row-major or column-major order. \footnote{\url{https://github.com/ROCm/amd_matrix_instruction_calculator}. This is a useful resource to learn more about the register tile layouts and shapes.} Since threads hold consecutive elements in the reduction dimension of MFMA instructions, the register layout decides whether we are reducing over rows or columns of memory. When loading data from shared memory to registers, we typically use two different types of instructions for BF16:
\vspace{1mm}
\begin{itemize}[itemsep=0.1pt,topsep=0pt,leftmargin=*]
\item \textbf{Row layouts (ds\_read\_b128):} \verb|ds_read*| instructions take in 3 arguments: 1) vector registers to serve as the destination of the data, 2) a shared memory address, and 3) a constant offset from the shared memory address to read data. The \verb|ds_read_b128| instruction is carried out in 4 phases where subsets of the 64 threads execute in each phase (Tab. \ref{tab:phase-bank}). As a result, ensuring that no two threads belonging to the same phase access the same shared memory bank eliminates bank conflicts for this instruction. With 16 threads participating in each phase, each thread accessing 128 bits or 4 banks, all 64 banks should be read and we maximize shared memory throughput.
\item \textbf{Column layouts (ds\_read\_b64\_tr\_b16):} Normally, reading data in a column-major format involves issuing multiple loads for each individual row we access. Using the \verb|ds_read_b64_tr_b16| instruction allows us to perform these column-major loads much more efficiently by having threads access shared memory at a greater granularity. Take the 16x32 register tile for example in \Cref{fig:column_layout}).

In a 16x32 column layout register tile where each thread holds 8 contiguous elements in the reduction dimension (i.e., stride 8), thread 0 holds the first element in rows 0-7. They are shaded in the two tables too. The \verb|ds_read_b64_tr_b16| instruction accomplishes this load by having different threads read data that is placed in another thread's vector register lane. For example, thread 4 technically reads the first element in the second row, but instead of placing it in its own register lane, it puts it into thread 0's register. This instruction executes in two phases where the first 32 threads read during the first phase and the remaining ones read during the second phase. If this SMEM tile only needed to support reads from column-major 16x32 register tiles, an unswizzled pattern would be sufficient to eliminating bank conflicts. However, shown in Figure \ref{fig:swizzle_plot}, a swizzle is necessary to support reads from a row-major 16x32 register tile.
\end{itemize}

\input{figures/layouts/col_layout}

\paragraph{Shared Memory and Register Tile Shapes.} While the previous subsection focused on loads from a 16x32 shared memory tile shape to a 16x32 register tile, different workloads could warrant other shared memory and register tile shapes mapping to different MFMA instructions. HK supports loads and stores between shared memory and register tile shapes as long as one is a multiple of the other. For example, loading from a 16x32 shared memory tile into a 32x16 register tile is not supported, but loading from a 16x16 shared memory tile into a 32x16 register tile is permitted. Each shared memory tile shape is also equipped with a default swizzling pattern that is a best-effort attempt to eliminate bank conflicts for common access patterns. 

\label{app:proof-by-contradiction}
\paragraph{A single swizzle is not possible.} To show why a single swizzling pattern is insufficient across different register tile shapes and layouts on AMD GPUs, consider the following two access patterns that surface in attention backwards:
\begin{enumerate}
    \item A row-layout 16x16 bf16 tile is written to shared memory. For this tile configuration, each thread holds 4 contiguous bf16 values - 64 bits in memory - and the most optimal instruction to issue this write is \verb|ds_write_b64|. Avoiding bank conflicts for this access requires a swizzle pattern that respects the phase ordering and bank behavior as listed in Table \ref{tab:phase-bank}. In this case, a swizzle that abides by these constraints is \verb|offset ^= ((offset % 512) >> 7) << 3|, where 64-bit chunks of memory is shifted around memory using an XOR swizzle.
    \item A row-layout 16x32 bf16 tile is read from shared memory. For this tile, each thread holds 8 contiguous bf16 values - 128 bits in memory - and the most optimal instruction to issue this read is \verb|ds_read_b128|.
\end{enumerate}
Regardless of the swizzling pattern required for \verb|ds_read_b128|, the granularities of these two instructions are in conflict with each other. \verb|ds_read_b128| requires at least 128 bits of memory to be contiguous in shared memory, and the swizzle pattern for \verb|ds_write_b64| breaks apart memory into 64-bit chunks. As a result, different swizzling patterns need to be used for each.

\subsection{Phases and Banks}

Since per-instruction phase and bank behavior is not well documented, we create simple solvers for both. The phase solver iterates over every pair of threads in a wave and performs the shared memory instruction on the same bank. If a shared memory bank occurs, the two threads belong to the same phase. The bank solver takes two threads belonging to the same phase, fixes one thread to access bank zero, and accesses other banks using the other thread. The number of banks between bank zero and the first bank where a bank conflict occurs represents the number of banks accessible by the shared memory instruction.

\begin{table}[h]
    \vspace{-2mm}
    \setlength{\tabcolsep}{4pt}
    \renewcommand{\arraystretch}{1.05}
    \begin{center}
    \begin{small}
    \begin{sc}
    \begin{tabular}{lcc p{0.40\linewidth}}
        \toprule
        \textbf{Instr.} & \textbf{Banks} & \textbf{Phase} & \textbf{Active threads} \\
        \midrule
        \multirow{4}{*}{ds\_read\_b128} & \multirow{4}{*}{64}
          & 0 & 0-3, 12-15, 20-27 \\
          & & 1 & 4-11, 16-19, 28-31 \\
          & & 2 & 32-35, 44-47, 52-59 \\
          & & 3 & 36-43, 48-51, 60-63 \\
        \midrule
        \multirow{8}{*}{ds\_read\_b96} & \multirow{8}{*}{32}
          & 0 & 0-3, 20-23 \\
          & & 1 & 4-7, 16-19 \\
          & & 2 & 8-11, 28-31 \\
          & & 3 & 12-15, 24-27 \\
          & & 4 & 32-35, 52-55 \\
          & & 5 & 36-39, 48-51 \\
          & & 6 & 40-43, 60-63 \\
          & & 7 & 44-47, 56-59 \\
        \midrule
        \multirow{4}{*}{ds\_write\_b64} & \multirow{4}{*}{32}
          & 0 & 0-15 \\
          & & 1 & 16-31 \\
          & & 2 & 32-47 \\
          & & 3 & 48-63 \\
        \midrule
        \multirow{2}{*}{ds\_read\_b64} & \multirow{2}{*}{64}
          & 0 & 0-31 \\
          & & 1 & 32-63 \\
        \bottomrule
    \end{tabular}
    \caption{\textbf{Phase-bank table.} The number of banks available to each shared memory instruction and the number of phases (and participating threads per phase) each instruction requires.}
    \label{tab:phase-bank}
    \end{sc}
    \end{small}
    \end{center}
\end{table}

\subsection{Pinned register tiles}
\label{app::hk-asm}
{\shortname} lets developers control the registers assigned to different register tiles through the concept of register ranges. For example:
\begin{lstlisting}[language=C++, basicstyle=\small\ttfamily]
using Q_ranges = 
  split_many_t<type_list<range<24, 39>>, 4>;
\end{lstlisting}
This defines a list of register ranges where each range contains exactly 4 registers. The register ranges here are \verb|v[24:27]|, \verb|v[28:31]|, \verb|v[32:35]|, and \verb|v[36:39]|. Each register range corresponds to the registers required to hold a single base tile in a register tile, and we specify a list of register ranges when defining a register tile like:

\vspace{2mm}
\begin{lstlisting}[language=C++, basicstyle=\small\ttfamily]
rt<bf16, 16, 128, row_l, 
   rt_16x32_s, Q_ranges> Q_i;
\end{lstlisting}
Developers can call the same functions in {\shortname}, but now have them operate on specific registers instead. As mentioned in Section \ref{sec:hip-warp}, this allowed us to pin AGPRs as the A or B matrix inputs to MFMA instructions when writing our attention backwards kernel.

\vspace{-2mm}

\subsection{Compiler hints}

The LLVM compiler accepts developer-provided hints to guide instruction scheduling on AMD GPUs.\footnote{\url{https://llvm.org/docs/AMDGPUUsage.html}} We use these some of these hints in our kernels to augment the scheduling that we apply at the HIP level. There are three sets of intrinsics that we find useful.

\begin{enumerate}
 \item The \verb|llvm.amdgcn.sched.barrier| intrinsic accepts a mask, which tells the compiler which types of instructions can cross the intrinsic in the compiled schedule. Masks exist for all instructions, VALU (vector ALU) instructions, SALU (scalar ALU) instructions, VMEM (global memory) instructions, MFMA (matrix) instructions, and so on, as described in the documentation. This intrinsic is used to establish hard boundaries between clusters of instructions in our clusters. For instance, see \verb|__builtin_amdgcn_sched_barrier(0)| in our Appendix~\ref{app:kernel_listings} kernel listings. 

 \item The \verb|llvm.amdgcn.sched.group.barrier| intrinsic is used to establish scheduling pipelines. The developer considers a group of instructions and specifies to the compiler precisely how to order them. 
 A call of the builtin accepts a \verb|mask| that specifies the instruction type, \verb|size| indicating the number of instructions of this type that calls the builtin applies to, and a \verb|sync id| serves as an identifier. 

 This builtin creates a ``super group'' of ``instruction groups''. The \verb|sync id| identifies the super group; order is enforced between instruction groups with the same \verb|sync id|, and instruction groups are only scheduled relative to other groups with the same \verb|sync id|. 

The \verb|mask| is a bitmask. Here are some frequently used ones:
\begin{lstlisting}
#define MFMA_MASK 0x08
#define VMEM_MASK 0x20
#define DS_MASK 0x100
\end{lstlisting}

Each call of this intrinsic looks backward and finds the most recent of the corresponding type of instruction which are not already part of a group created by a previous \verb|__builtin_amdgcn_sched_group_barrier|

For example:
\begin{lstlisting}
__builtin_amdgcn_sched_group_barrier(VMEM_MASK, 4, 0);
__builtin_amdgcn_sched_group_barrier(MFMA_MASK, 4, 0);
__builtin_amdgcn_sched_group_barrier(DS_MASK, 8, 0);
__builtin_amdgcn_sched_group_barrier(MFMA_MASK, 4, 0); 
\end{lstlisting}

This finds the last 4 global memory (VMEM) loads and schedules those first, then finds the last 4 matrix (MFMA) instructions and schedules those after the  global memory loads, then finds the last 8 shared to register (DS READ) loads and schedules those after the 4 MFMAs, then finds the previous 4 MFMAs before the last 4 and schedules those last.

 \item The \verb|__builtin_amdgcn_s_setprio| intrinsic lets us specify the priority (0-3) for a wave relative to other waves that are competing for hardware resources. We use this around compute clusters in the \textsc{8-wave} ping-pong schedule as shown in our GEMM and attention forwards kernels.
\end{enumerate}

The limitation of using these hints for scheduling is that any code wrapped in \verb|asm volatile| is black-box to the compiler, and for some instructions (e.g., \verb|v_cvt_pk_bf16_f32| to convert from BF16 to FP32), LLVM builtins are missing. 

It is worth noting that the current Modular AI GEMM kernels (as of October 2025) rely on compiler hints (\verb|sched_group_barrier|). This approach could work since Modular is replacing the compiler as well; however, it requires the developer to think about every single instruction issue rather than providing the option to think about bulk tile primitives. Our opinion is that using scheduling hints at the cluster scope and tile primitives to form the top level kernel schedule (as in our attention forwards kernel) may help simplify programmability and maintain performance.

\subsection{Synchronization}

\paragraph{Loads}
Similar to asynchronous tensor memory acceleration (TMA), AMD CDNA3 and CDNA4 GPUs have direct global memory to LDS (shared memory) load instructions called \verb|buffer_load_dword|. These instructions can load one \verb|dword| (4 bytes, one bank), three \verb|dwordx3| (12 bytes), or four \verb|dwordx4| (16 bytes). The instructions skip the register file and accept constant offsets that also help mitigate address calculation overheads. Once the load is issued, the instruction \verb|vmcnt(x)| specifies to wait until only \verb|x| global memory load instructions remain in flight, and \verb|vmcnt(0)| indicates to wait on all outstanding loads. Ideally, we can separate the distance between load issues and these waits (as shown in our GEMM and attention kernels (Sec.~\ref{app:kernel_listings})).

Similarly, there are asynchronous shared to register memory loads called \verb|ds_read_b32| (or \verb|b64| for 8 bytes, \verb|b96| for 12 bytes, \verb|b128| for 16 bytes). The instruction \verb|lgmkcnt(x)| specifies to wait until \verb|x| shared to register instructions remain in flight, and \verb|lgmkcnt(0)| indicates to wait on all outstanding shared to register loads. 

\paragraph{Execution}  An \verb|__builtin_amdgcn_s_barrier()| functionally matches \verb|syncthreads|. Note that AMD has a SIMD model and NVIDIA follows SIMT, so we do not need to sync the threads within the warp on AMD. As a result, there is no equivalent of \verb|syncwarp| on AMD.

%% file: figures/layouts/col_layout.tex
\begin{figure*}[h]
\centering
\includegraphics[width=0.48\linewidth]{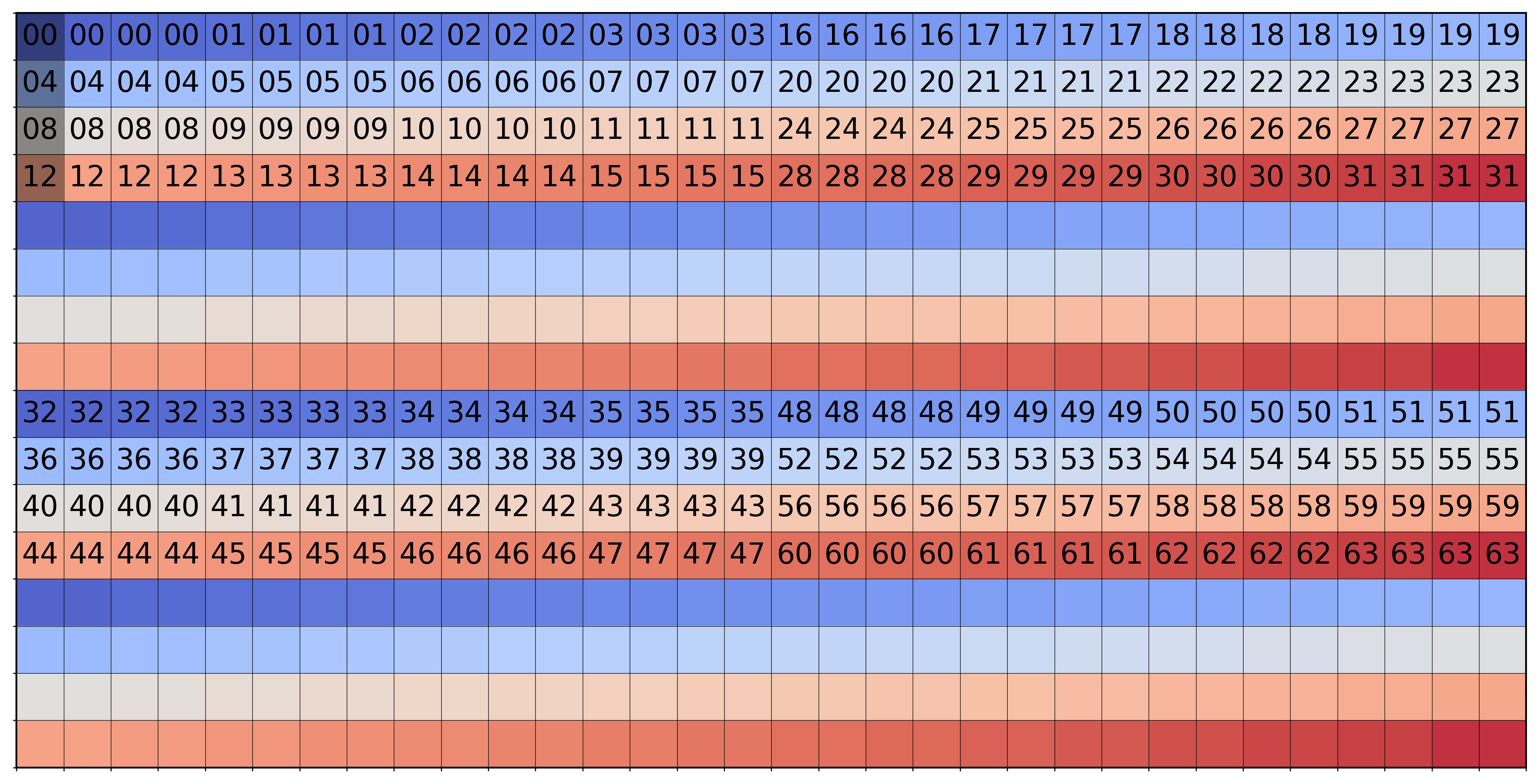}
\includegraphics[width=0.48\linewidth]{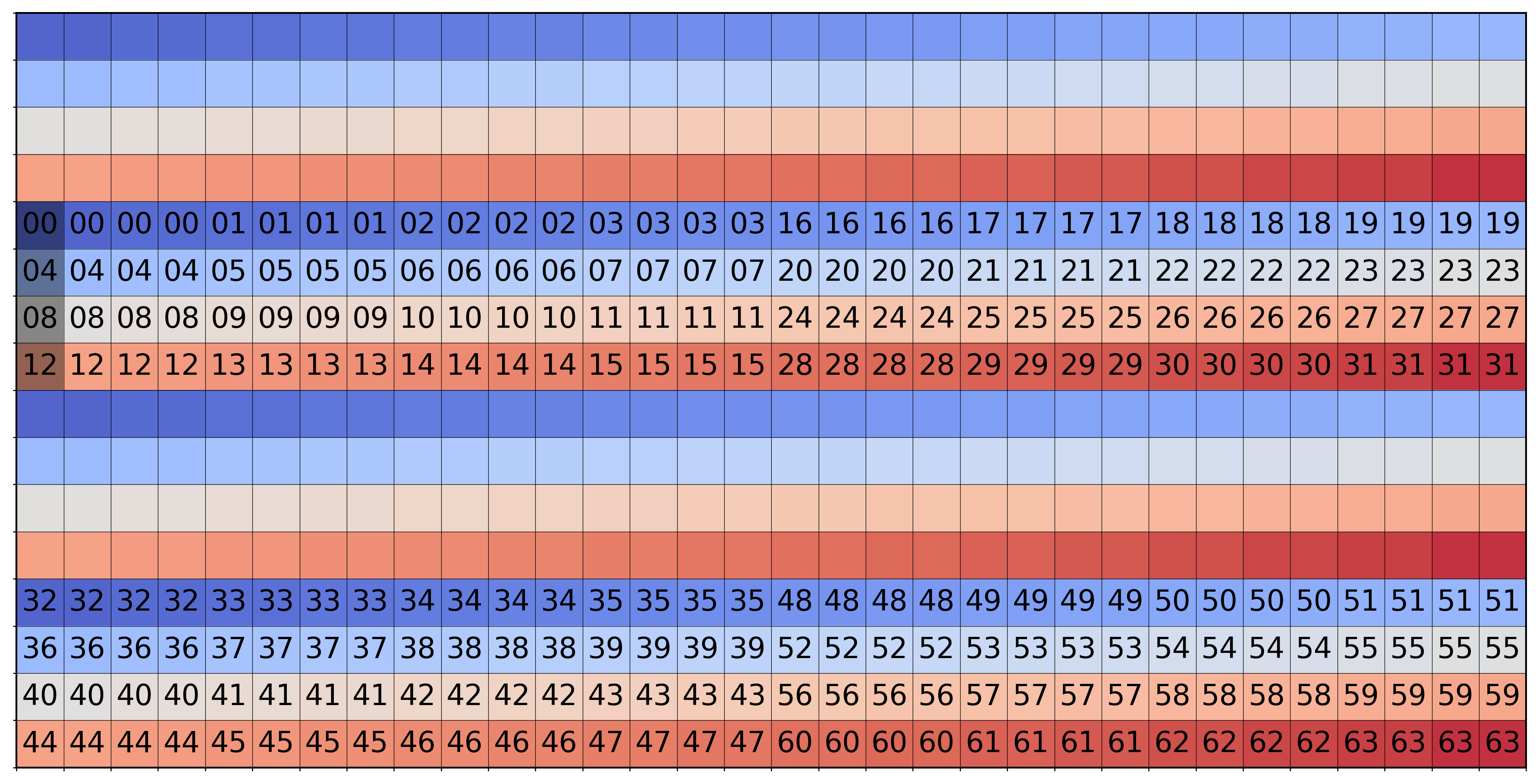}
\caption{Shared memory access pattern for \texttt{ds\_read\_b64\_tr\_b16} for reading in a 16x32 column layout register tile. Each cell represents a single 16 bit values and the numbers represent threads. Different colors represent different shared memory banks (note that a bank spans two cells).}
\label{fig:column_layout}
\vspace{-3mm}
\end{figure*}

%% file: appendix/kernels.tex
\clearpage 
\section{{\name} kernel listings}
\label{app:kernel_listings}

This section demonstrates {\name} kernel examples and discusses the algorithmic details of our kernel implementations. 

\subsection{Matrix Multiply}
The BF16 GEMM kernel (\Cref{fig:matmul-pc-template}) decomposes the problem into computing a $256\times256$ output tile per thread block (denoted by \verb|BLOCK_SIZE|). In the prologue, the kernel pre-loads the A and B input matrices from global to shared memory. The kernel inserts a conditional barrier to stall half the waves (one wave per SIMD) while the other half begins performing additional loads. When this leader wavegroup finishes its additional loads, it unblocks the follower wavegroup through the \verb|s_barrier| invocation. Thereafter, the two wavegroups alternate between the compute and memory clusters shown in the hotloop, where the end of a cluster is always demarked by an \verb|s_barrier|. This represents the \textsc{8-wave ping pong} kernel schedule introduced in ~\Cref{sec:hip-block}.

For the MI325X version of the kernel, we maintain the same \textsc{8-wave} structure, however the hardware only has 65 KB of shared memory so we cannot double buffer in shared memory. Instead, we double buffer using the register file. We do not use the direct HBM to LDS buffer loads, and instead load from HBM to a register buffer, while the waves compute MFMAs on the previously stored register tiles. When the compute completes, the data from the register buffers gets stored down to shared memory using a \verb|ds_write|.

\begin{figure*}[t] 
\begin{lstlisting}[language=CUDACPP, basicstyle=\scriptsize\ttfamily]
constexpr int BLOCK_SIZE       = 256;  
constexpr int HALF_BLOCK_SIZE  = BLOCK_SIZE / 2;
constexpr int K_STEP           = 64;
constexpr int WARPS_M          = 2;
constexpr int WARPS_N          = 4;
constexpr int REG_BLOCK_M      = BLOCK_SIZE / WARPS_M;
constexpr int REG_BLOCK_N      = BLOCK_SIZE / WARPS_N;
constexpr int HALF_REG_BLOCK_M = REG_BLOCK_M / 2;
constexpr int HALF_REG_BLOCK_N = REG_BLOCK_N / 2;
constexpr int DOT_SLICE        = 32;

__global__ void GEMM_BF16(const micro_globals g) {
    // setup
    extern __shared__ alignment_dummy __shm[];
    shared_allocator al((int*)&__shm[0]);
    using ST_A = st_bf<HALF_BLOCK_SIZE, K_STEP, st_16x32_s>;
    using ST_B = st_bf<HALF_BLOCK_SIZE, K_STEP, st_16x32_s>;
    ST_A (&As)[2][2] = al.allocate<ST_A, 2, 2>();
    ST_B (&Bs)[2][2] = al.allocate<ST_B, 2, 2>();

    rt_bf<HALF_REG_BLOCK_M, K_STEP, row_l, rt_16x32_s> A_tile;
    rt_bf<HALF_REG_BLOCK_N, K_STEP, row_l, rt_16x32_s> B_tile_0;
    rt_bf<HALF_REG_BLOCK_N, K_STEP, row_l, rt_16x32_s> B_tile_1;
    rt_fl<HALF_REG_BLOCK_M, HALF_REG_BLOCK_N, col_l, rt_16x16_s> C_accum[2][2];
    zero(C_accum[0][0]);
    zero(C_accum[0][1]);
    zero(C_accum[1][0]);
    zero(C_accum[1][1]);

    // L2 and LLC Cache swizzling
    int wgid = (blockIdx.y * gridDim.x) + blockIdx.x;
    const int NUM_WGS  = gridDim.x * gridDim.y;
    const int WGM = 8;
    // Swizzle chiplet so that wgids are in the same XCD.
    wgid = chiplet_transform_chunked(wgid, NUM_WGS, NUM_XCDS, WGM*WGM);
    // Swizzle for better L2 within the same XCD.
    const int num_pid_m = ceil_div(M, BLOCK_SIZE); 
    const int num_pid_n = ceil_div(N, BLOCK_SIZE); 
    const int num_wgid_in_group = WGM * num_pid_n;
    int group_id = wgid / num_wgid_in_group;
    int first_pid_m = group_id * WGM;
    int group_size_m = min(num_pid_m - first_pid_m, WGM);
    int pid_m = first_pid_m + ((wgid % num_wgid_in_group) % group_size_m);
    int pid_n = (wgid % num_wgid_in_group) / group_size_m;
    // Assign the tile's row/column based on the pid_m and pid_n.
    int row = pid_m; 
    int col = pid_n;  

    const int warp_id = kittens::warpid();
    const int warp_row = warp_id / 4;
    const int warp_col = warp_id % 4;
    const int num_tiles = K / K_STEP;

    int tic = 0;
    int toc = 1;

   // preload
    G::load(Bs[tic][0], g.b, {0, 0, col*2, 0});
    G::load(As[tic][0], g.a, {0, 0, row*2, 0});
    G::load(Bs[tic][1], g.b, {0, 0, col*2 + 1, 0});
    G::load(As[tic][1], g.a, {0, 0, row*2 + 1, 0});

    // conditional stagger
    if (warp_row == 1) {
        __builtin_amdgcn_s_barrier();
    }

    asm volatile("s_waitcnt vmcnt(4)");
    __builtin_amdgcn_s_barrier();

    // preload
    G::load(Bs[toc][0], g.b, {0, 0, col*2, 1});
    G::load(As[toc][0], g.a, {0, 0, row*2, 1});
    G::load(Bs[toc][1], g.b, {0, 0, col*2 + 1, 1});

    asm volatile("s_waitcnt vmcnt(6)");
    __builtin_amdgcn_s_barrier();
\end{lstlisting}
\caption{HK BF16 GEMM, which is competitive with AITER on CDNA4.}
\label{fig:matmul-pc-template}
\end{figure*}

\begin{figure*}[t]  
\begin{lstlisting}[basicstyle=\scriptsize\ttfamily]
    #pragma unroll
    for (int tile = 0; tile < num_tiles - 2; ++tile, tic^=1, toc^=1) {

        auto st_subtile_b = subtile_inplace<HALF_REG_BLOCK_N, K_STEP>(Bs[tic][0], {warp_col, 0});
        load(B_tile_0, st_subtile_b);
        auto st_subtile_a = subtile_inplace<HALF_REG_BLOCK_M, K_STEP>(As[tic][0], {warp_row, 0});
        load(A_tile, st_subtile_a);
        G::load(As[toc][1], g.a, {0, 0, row*2 + 1, tile + 1});
        asm volatile("s_waitcnt lgkmcnt(8)");
        __builtin_amdgcn_s_barrier();

        asm volatile("s_waitcnt lgkmcnt(0)");
        __builtin_amdgcn_s_setprio(1);
        mma_ABt(C_accum[0][0], A_tile, B_tile_0, C_accum[0][0]);
        __builtin_amdgcn_s_setprio(0);
        __builtin_amdgcn_s_barrier();
        __builtin_amdgcn_sched_barrier(0);

        st_subtile_b = subtile_inplace<HALF_REG_BLOCK_N, K_STEP>(Bs[tic][1], {warp_col, 0});
        load(B_tile_1, st_subtile_b);
        G::load(Bs[tic][0], g.b, {0, 0, col*2, tile + 2});
        __builtin_amdgcn_s_barrier();

        asm volatile("s_waitcnt lgkmcnt(0)");
        __builtin_amdgcn_s_setprio(1);
        mma_ABt(C_accum[0][1], A_tile, B_tile_1, C_accum[0][1]);
        __builtin_amdgcn_s_setprio(0);
        __builtin_amdgcn_s_barrier();

        st_subtile_a = subtile_inplace<HALF_REG_BLOCK_M, K_STEP>(As[tic][1], {warp_row, 0});
        load(A_tile, st_subtile_a);
        G::load(As[tic][0], g.a, {0, 0, row*2, tile + 2});
        __builtin_amdgcn_s_barrier();

        asm volatile("s_waitcnt lgkmcnt(0)");
        __builtin_amdgcn_s_setprio(1);
        mma_ABt(C_accum[1][0], A_tile, B_tile_0, C_accum[1][0]);
        __builtin_amdgcn_s_setprio(0);
        __builtin_amdgcn_s_barrier();
        __builtin_amdgcn_sched_barrier(0);

        G::load(Bs[tic][1], g.b, {0, 0, col*2 + 1, tile + 2});
        asm volatile("s_waitcnt vmcnt(6)");
        __builtin_amdgcn_s_barrier();

        __builtin_amdgcn_s_setprio(1);
        mma_ABt(C_accum[1][1], A_tile, B_tile_1, C_accum[1][1]);
        __builtin_amdgcn_s_setprio(0);
        __builtin_amdgcn_s_barrier();
    }

    // Epilogue not shown

    if (warp_row == 0) {
        __builtin_amdgcn_s_barrier();
    }

    // store results
    store(g.c, C_accum[0][0], {0, 0, (row * 2) * WARPS_M + warp_row, col * 2 * WARPS_N + warp_col});
    store(g.c, C_accum[0][1], {0, 0, (row * 2) * WARPS_M + warp_row, col * 2 * WARPS_N + WARPS_N + warp_col});
    store(g.c, C_accum[1][0], {0, 0, (row * 2) * WARPS_M + WARPS_M + warp_row, col * 2 * WARPS_N + warp_col});
    store(g.c, C_accum[1][1], {0, 0, (row * 2) * WARPS_M + WARPS_M + warp_row, col * 2 * WARPS_N + WARPS_N + warp_col});
}
\end{lstlisting}
\end{figure*}

\clearpage
\subsection{Fused Dropout + Residual + Layernorm}
A very simple {\name} kernel that processes a chunk of vectors along the sequence dimension per thread block. This kernel listing demonstrates {\shortname} operators and vectors, which resemble those in PyTorch (e.g., \verb|sum|, \verb|add|, \verb|mul|, \verb|div|).

\begin{figure*}[h] 
\begin{lstlisting}[language=CUDACPP, basicstyle=\scriptsize\ttfamily]
__global__ void fused_layernorm(const norm_globals g) {

    auto warpid = kittens::warpid();
    
    const int batch = blockIdx.y;
    const int seq_start = blockIdx.x*g.n_per_tile;
    constexpr int d_model = 128;
    
    rv_naive<bf16, d_model> residual_s_reg, x_s_reg, norm_weight_s_reg, norm_bias_s_reg;
    load(x_s_reg, g.x, {0, batch, seq_start + warpid, 0});
    asm volatile("s_waitcnt vmcnt(0)");
    load(residual_s_reg, g.residual, {0, batch, seq_start + warpid, 0});
    bf16 mean = __float2bfloat16(0.0f);
    bf16 var  = __float2bfloat16(0.0f);      
    if constexpr (DROPOUT_P > 0.0f) {
        dropout_mask(x_s_reg, DROPOUT_P); 
    }
    if constexpr (DROPOUT_P > 0.0f) {
        constexpr float scale = 1.0f / (1.0f - DROPOUT_P);
        mul(x_s_reg, x_s_reg, __float2bfloat16(scale));
    }
    asm volatile("s_waitcnt vmcnt(0)");
    load(norm_weight_s_reg, g.norm_weight, {0,0,0,0});
    load(norm_bias_s_reg, g.norm_bias, {0,0,0,0});
    add(residual_s_reg, residual_s_reg, x_s_reg);   
    store(g.o_resid, residual_s_reg, {0, batch, seq_start + warpid, 0});

    // mean and variance
    sum(mean, residual_s_reg); 
    constexpr float dim_scale = 1.0f / d_model;
    mean = mean * __float2bfloat16(dim_scale);
    sub(residual_s_reg, residual_s_reg, mean);  
    mul(x_s_reg, residual_s_reg, residual_s_reg);
    sum(var, x_s_reg);
    var = var * __float2bfloat16(dim_scale);
    var = __float2bfloat16(sqrt(__bfloat162float(var + __float2bfloat16(1e-05f))));

    // compute norm
    div(residual_s_reg, residual_s_reg, var);
    asm volatile("s_waitcnt vmcnt(0)");
    mul(residual_s_reg, residual_s_reg, norm_weight_s_reg); 
    add(residual_s_reg, residual_s_reg, norm_bias_s_reg);
    store(g.o, residual_s_reg, {0, batch, seq_start+warpid, 0});
}
\end{lstlisting}
\label{fig:fused-template}
\caption{Fused Dropout + Residual + Layernorm kernel outperforming \texttt{torch.compile}.}
\end{figure*}

\clearpage
\subsection{Attention}
The {\name} attention kernel uses an \textsc{8-wave ping pong} schedule. Each wave computes a $32 \times 128$ tile of the output for a single head and batch. In the prologue, all eight waves first collaboratively load tiles of keys and values, and their own personal tiles of queries. The threads perform the initial query-key matrix multiply and first half of softmax. Then the kernel uses a conditional barrier to stall half the waves (one wave per SIMD). The leader wavegroup proceeds ahead, loading the next tiles of keys and values and upon completion, unlocks the follower wavegroup. In the hotloop, the two wavegroups alternate between compute clusters (each involving matrix multiplies and vector operations) and load clusters (involving global to shared and shared to register loads). 

In compute clusters, the compiler interleaves vector ops and matrix ops. We can also use \verb|sched_barrier| hints to guide the LLVM compiler to interleave in custom patterns. 

\begin{figure*}[t]
\begin{lstlisting}[language=CUDACPP, basicstyle=\scriptsize\ttfamily]
template<int D>
__global__ void attend_ker(const attn_globals<D> g) {

    extern __shared__ alignment_dummy __shm[];
    shared_allocator al((int*)&__shm[0]);
    st_bf<KV_BLOCK_SIZE, ATTN_D, st_32x32_s> (&k_smem)[2] = al.allocate<st_bf<KV_BLOCK_SIZE, ATTN_D, st_32x32_s>, 2>();
    st_bf<KV_BLOCK_SIZE, ATTN_D, st_8x32_s> (&v_smem)[2] = al.allocate<st_bf<KV_BLOCK_SIZE, ATTN_D, st_8x32_s>, 2>();
    
    const int head_idx = (blockIdx.x % GROUP_SIZE) * GROUP_SIZE + (blockIdx.x / GROUP_SIZE);
    const int batch_idx = blockIdx.z;
    const int head_idx_kv = head_idx / GROUP_SIZE;
    const int block_tile_idx = blockIdx.y;
    const int tile_idx = block_tile_idx * NUM_WARPS + warpid();
    const int stagger = warpid() / 4;

    const int num_tiles = ATTN_N / KV_BLOCK_SIZE;
    constexpr float TEMPERATURE_SCALE = (D == 128) ? 0.08838834764f*1.44269504089f : 0.125f*1.44269504089f;

    // Initialize all of the register tiles.
    qo_tile<D, bf16> q_reg; // Q and K are both row layout, as we use mma_ABt.
    qo_tile_transposed<D, bf16> q_reg_transposed;
    kv_tile<D, bf16> k_reg;
    kv_tile_transposed<D, bf16> k_reg_transposed;

    kv_tile<D, bf16, col_l, rt_32x32_s> v_reg;
    qo_tile_transposed<D, float, col_l, rt_32x32_s> o_reg; // Output tile.
    attn_tile<D, float, col_l, rt_32x32_s> att_block[2]; // attention tile, in float.
    attn_tile<D, bf16, col_l, rt_32x32_s> att_block_bf16;
    typename attn_tile<D, float, col_l, rt_32x32_s>::row_vec max_vec, norm_vec, max_vec_prev;

    G::load<1, false>(k_smem[0], g.Kg, {batch_idx, 0, head_idx_kv, 0});
    __builtin_amdgcn_s_waitcnt(0);
    __builtin_amdgcn_sched_barrier(0);
    __builtin_amdgcn_s_barrier();

    qo_tile<D, float> q_reg_fl;
    load<1, qo_tile<D, float>, _gl_QKVO>(q_reg_fl, g.Qg, {batch_idx, tile_idx, head_idx, 0});
    mul(q_reg_fl, q_reg_fl, TEMPERATURE_SCALE);  // Use sqrtf for clarity
    copy(q_reg, q_reg_fl);
    swap_layout_and_transpose(q_reg_transposed, q_reg);

    zero(o_reg);
    zero(norm_vec);
    neg_infty(max_vec_prev);

    // All warps then collaboratively load in the first slice of V (V0) and the second slice of K (K1) into shared memory
    G::load<1, false>(k_smem[1], g.Kg, {batch_idx, 1, head_idx_kv, 0});
    // All warps then load in the first slice of K (K0)
    G::load<1, false>(v_smem[0], g.Vg, {batch_idx, 0, head_idx_kv, 0});
    load(k_reg, k_smem[0]);
    asm volatile("s_waitcnt lgkmcnt(0)");
    asm volatile("s_waitcnt vmcnt(2)");
    __builtin_amdgcn_sched_barrier(0);
    __builtin_amdgcn_s_barrier();

    // each warp performs QK0
    zero(att_block[0]);
    swap_layout_and_transpose(k_reg_transposed, k_reg);
    mma_AtB(att_block[0], k_reg_transposed, q_reg_transposed, att_block[0]);

    // each warp performs a partial softmax of QK0
    col_max(max_vec, att_block[0]);
    sub_col(att_block[0], att_block[0], max_vec);
    exp2(att_block[0], att_block[0]);
    sched_barrier_pairs<8, 6, 1>();
    sched_barrier_exp_pairs<8, 4, 1>();

    // conditional stagger
    if (stagger) {
        __builtin_amdgcn_sched_barrier(0);
        __builtin_amdgcn_s_barrier();
    }
\end{lstlisting}
\end{figure*}

\begin{figure*}[t]
\begin{lstlisting}[language=CUDACPP, basicstyle=\scriptsize\ttfamily]
    // All warps then load in the second slice of K (K1)
    load(k_reg, k_smem[1]);
    // All warps then collaboratively load in the third slice of K (K2) into shared memory
    G::load<1, false>(k_smem[0], g.Kg, {batch_idx, 2, head_idx_kv, 0});
    // All warps then collaboratively load in the second slice of V (V1) into shared memory 
    G::load<1, false>(v_smem[1], g.Vg, {batch_idx, 1, head_idx_kv, 0});
    asm volatile("s_waitcnt lgkmcnt(0)");
    asm volatile("s_waitcnt vmcnt(4)");
    __builtin_amdgcn_sched_barrier(0);
    __builtin_amdgcn_s_barrier();

    // hot loop
    for (int j = 3; j < num_tiles - 1; j += 2) {
        // Cluster 0:
        //      QK1
        zero(att_block[1]);
        swap_layout_and_transpose(k_reg_transposed, k_reg);
        mma_AtB(att_block[1], k_reg_transposed, q_reg_transposed, att_block[1]);
        //      Finish softmax for QK0
        sub(max_vec_prev, max_vec_prev, max_vec); 
        exp2(max_vec_prev, max_vec_prev);  
        mul(norm_vec, norm_vec, max_vec_prev);
        col_sum(norm_vec, att_block[0], norm_vec);
        copy(att_block_bf16, att_block[0]);
        sched_barrier_pairs<16, 3, 2>();
        __builtin_amdgcn_sched_barrier(0);
        __builtin_amdgcn_s_barrier();
        __builtin_amdgcn_sched_barrier(0);

        // Cluster 1:
        //      Load K3 into shared 
        G::load<1, false>(k_smem[1], g.Kg, {batch_idx, j, head_idx_kv, 0});
        //      Load V0 into registers
        load(v_reg, v_smem[0]);
        asm volatile("s_waitcnt lgkmcnt(0)");
        asm volatile("s_waitcnt vmcnt(4)");
        __builtin_amdgcn_sched_barrier(0);
        __builtin_amdgcn_s_barrier();
        __builtin_amdgcn_sched_barrier(0);

        // Cluster 2:
        //      A0V0
        __builtin_amdgcn_s_setprio(1);
        mul_col(o_reg, o_reg, max_vec_prev);
        __builtin_amdgcn_sched_barrier(0);
        mma_AtB(o_reg, v_reg, att_block_bf16, o_reg);
        //      Partial softmax for QK1
        copy(max_vec_prev, max_vec);
        col_max(max_vec, att_block[1], max_vec);
        sub_col(att_block[1], att_block[1], max_vec);
        exp2(att_block[1], att_block[1]);
        sched_barrier_pairs<8, 6, 3>();
        sched_barrier_exp_pairs<8, 4, 3>();
        __builtin_amdgcn_s_setprio(0);
        __builtin_amdgcn_sched_barrier(0);
        __builtin_amdgcn_s_barrier();
        __builtin_amdgcn_sched_barrier(0);

        // Cluster 3:
        //      Load V2 into shared
        G::load<1, false>(v_smem[0], g.Vg, {batch_idx, j - 1, head_idx_kv, 0});
        //      Load K2 into registers
        load(k_reg, k_smem[0]);
        asm volatile("s_waitcnt lgkmcnt(0)");
        asm volatile("s_waitcnt vmcnt(4)");
        __builtin_amdgcn_sched_barrier(0);
        __builtin_amdgcn_s_barrier();
        __builtin_amdgcn_sched_barrier(0);
\end{lstlisting}
\end{figure*}

\begin{figure*}[t]
\begin{lstlisting}[language=CUDACPP, basicstyle=\scriptsize\ttfamily]
        // Cluster 4:
        //      QK2
        zero(att_block[0]);
        swap_layout_and_transpose(k_reg_transposed, k_reg);
        mma_AtB(att_block[0], k_reg_transposed, q_reg_transposed, att_block[0]);
        //      Finish softmax for QK1
        sub(max_vec_prev, max_vec_prev, max_vec); 
        exp2(max_vec_prev, max_vec_prev);  
        mul(norm_vec, norm_vec, max_vec_prev);
        col_sum(norm_vec, att_block[1], norm_vec);
        copy(att_block_bf16, att_block[1]);
        sched_barrier_pairs<16, 3, 4>();
        __builtin_amdgcn_sched_barrier(0);
        __builtin_amdgcn_s_barrier();
        __builtin_amdgcn_sched_barrier(0);
        
        // Cluster 5:
        //      Load K4 into shared
        G::load<1, false>(k_smem[0], g.Kg, {batch_idx, j + 1, head_idx_kv, 0});
        //      Load V1 into registers
        load(v_reg, v_smem[1]);
        asm volatile("s_waitcnt lgkmcnt(0)");
        asm volatile("s_waitcnt vmcnt(4)");
        __builtin_amdgcn_sched_barrier(0);
        __builtin_amdgcn_s_barrier();
        __builtin_amdgcn_sched_barrier(0);
        
        // Cluster 6:
        //      A1V1
        __builtin_amdgcn_s_setprio(1);
        mul_col(o_reg, o_reg, max_vec_prev);
        __builtin_amdgcn_sched_barrier(0);
        mma_AtB(o_reg, v_reg, att_block_bf16, o_reg);
        //      Partial softmax for QK2
        copy(max_vec_prev, max_vec);
        col_max(max_vec, att_block[0], max_vec);
        sub_col(att_block[0], att_block[0], max_vec);
        exp2(att_block[0], att_block[0]);
        sched_barrier_pairs<8, 6, 5>();
        sched_barrier_exp_pairs<8, 4, 5>();
        __builtin_amdgcn_s_setprio(0);
        __builtin_amdgcn_sched_barrier(0);
        __builtin_amdgcn_s_barrier();
        __builtin_amdgcn_sched_barrier(0);

        // Cluster 7:
        //      Load V3 into shared
        G::load<1, false>(v_smem[1], g.Vg, {batch_idx, j, head_idx_kv, 0});
        //      Load K3 into registers
        load(k_reg, k_smem[1]);
        asm volatile("s_waitcnt lgkmcnt(0)");
        asm volatile("s_waitcnt vmcnt(4)");
        __builtin_amdgcn_sched_barrier(0);
        __builtin_amdgcn_s_barrier();
        __builtin_amdgcn_sched_barrier(0);
    }

    // Epilogue not shown

    // Conclusion
    if (!stagger) {
        __builtin_amdgcn_s_barrier();
    }

    qo_tile<D, float, row_l, rt_32x32_s> o_reg_transposed;
    swap_layout_and_transpose(o_reg_transposed, o_reg);
    store<1>(g.Og, o_reg_transposed, {batch_idx, tile_idx, head_idx, 0});

    // multiply by ln(2)
    mul(max_vec, max_vec, 0.69314718056f);
    log(norm_vec, norm_vec);
    add(norm_vec, norm_vec, max_vec);
    store(g.L_vec, norm_vec, {batch_idx, head_idx, 0, tile_idx});
}
\end{lstlisting}
\label{fig:matmul-pc-template}
\caption{{\name} non-causal attention forwards kernel that competes with the assembly kernel provided in AMD's AITER library.}
\end{figure*}

%% file: appendix/case_study.tex
\section{Case study: Preliminary findings for the FP6 GEMM}
\label{app:case_study}

We discuss our \textbf{initial empirical observations} of FP6 hardware behavior on AMD's MI350X and MI355X GPUs. The goal is to characterize how FP6 memory movement and matrix-core operations behave in practice. AMD’s own CK library baselines are \textbf{unoptimized} at the time of writing, and our results should be interpreted as preliminary. FP6 is exciting because AMD matrix cores achieve twice the peak FLOPS for FP6 as NVIDIA devices. 
However, in practice, we incur challenges in loading FP6 values to and from memory, which impact our ability to achieve high utilization.

\paragraph{Memory Loads.}
Our FP6 GEMM kernel uses 4 waves to cooperatively load a 128$\times$128 tile from global memory in each memory cluster. With the FP6 datatype, that tile is $128 \times 128 \times 6 / 8 = 12{,}288$ bytes, or $48$ bytes per thread. We consider the following CDNA4 instructions as options to load from global memory:

\begin{itemize}
    \item \verb|buffer_load_dwordx4| minimizes the number of instructions issued per thread (at 3 per tile). However, naively using this instruction to load to shared memory and then using \verb|ds_read_b128| followed by \verb|ds_read_b64| to read the 24 consecutive bytes owned by each thread from shared memory into registers causes shared memory alignment issues, because \verb|ds_read_b128| must be 16-byte aligned for maximum performance. For example, the second thread on each row of the tile performs a \verb|ds_read_b128| at an offset of 24 bytes into the row. Our kernel becomes shared-memory bound in this configuration. 

    One solution is for every second thread (\verb|laneid % 2 == 1|) to flip the two shared memory load instructions, so \verb|ds_read_b64| loads the first 8 bytes of the thread's data, and \verb|ds_read_b128| loads the next 16 bytes. We find that loading into a register destination that depends on the thread ID requires the use of slow scratch storage. Instead, for threads with flipped \verb|ds_read_*| instructions, we continue reading \verb|ds_read_b128| into the first four registers and \verb|ds_read_b64| into the second two registers, regardless of thread. This approach then requires a shuffle, where we conditionally swap the two regions in registers based on thread ID. This ``breaks the wave,'' resulting in two jump instructions in addition to the VALU instructions required to move the memory. We find that the incurred jump + VALU from shuffling comprises 49\% of the cycles in the kernel's hot loop, resulting in a kernel that achieves only 2430 TFLOPs. 

    Alternatively, we can use two \verb|ds_read_b96| instructions. This approach causes misalignment between the 16-byte chunks read by \verb|buffer_load_dwordx4| from global to shared memory and the 12-byte chunks read by \verb|ds_read_b96|. As a result, we are unable to swizzle the data in shared memory, resulting in 4-way bank conflicts. Given these issues with \verb|buffer_load_dwordx4|, we look at other options for loading FP6 values from global to shared memory.
    
    \item \verb|buffer_load_dwordx3| is appealing for FP6 because it allows a warp to exactly load an 8$\times$128 tile from global memory. It also supports swizzling. Unfortunately, this instruction for FP6 does not outperform FP8 in instruction issue count: each thread issues 4 instructions to load a 128$\times$128 tile, which is the same number used in our FP8 GEMM kernel. This instruction also loads each thread's 12 bytes of data at a stride of 16 bytes, wasting 25\% of the resulting shared memory tile and rendering 8 of the 32 banks available to \verb|ds_read_b96| (Table~\ref{tab:phase-bank}) unused. Despite these drawbacks, we find this is likely the most compelling global-to-shared load instruction for our FP6 GEMM use case. \verb|ds_read_b96| is the natural LDS-to-register load instruction for data loaded to shared memory using \verb|buffer_load_dwordx3|. We find that \verb|ds_read_b96| works well on 16-byte aligned shared memory addresses.

    \item \verb|buffer_load_dwordx1| avoids shared memory waste, alignment issues, and swizzling limitations, but the kernel becomes bottlenecked by the number of instructions issued, resulting in a slower kernel than the one built with \verb|buffer_load_dwordx3|.
\end{itemize}

\paragraph{Matrix Core Operation.}
With the BF16 and FP8 GEMMs, we found the fastest MFMA instruction is the smaller one available for the given dtype (Figure~\ref{fig:pingpong}), which in this case is \verb|v_mfma_f32_16x16x128_f8f6f4|. In this instruction, each thread owns 32 consecutive elements, or 24 consecutive bytes, of each FP6 operand matrix. In our kernel, each thread issues two \verb|ds_read_b96| instructions: one for the first 12 bytes, and one for the latter 12 bytes. \verb|ds_read_b96| constrains the destination base address in the register file to a 16 byte aligned address, so in order to achieve continuity between the 24 bytes of data, we must issue three \verb|v_mov_b32_e32| instructions to shuffle each of the three registers written to by the second \verb|v_mov_b32_e32| down one register. Note that this shuffle is not as expensive as the shuffle required when using \verb|buffer_load_dwordx4| with \verb|ds_read_b128| and \verb|ds_read_b64| because fewer data are shuffled and no waves break due to this shuffle. We note that HIPCC does not handle this shuffle requirement well, and this kernel at size 16384x16384x16384 spills 54 registers to scratch memory, resulting in a slow and incorrect kernel. To remedy this, we rewrote our FP6 GEMM while explicitly scheduling registers, allowing us to intelligently set which registers are used in this shuffle and completely removing register spills. This approach was not without hazards. We need to account for the instruction latency of \verb|v_mov_b32_e32|, so we manually ensure that at least 8 cycles of instructions are between the \verb|v_mov_b32_e32| instructions and the dependent MFMA instruction. In some cases, this involves manually inserting \verb|v_nop| instructions.

\paragraph{Results.}
FP6 matrix core speed is a standout feature of MI350X and MI355X GPUs.
Our FP6 GEMM kernel outperforms AMD's CK implementation and attains performance comparable to our own FP8 GEMM.
These results reflect our initial observations of FP6 hardware behavior; we expect additional improvements in future work.

\begin{figure}[t]
  \centering
  \includegraphics[width=\linewidth,height=0.25\textheight,keepaspectratio]{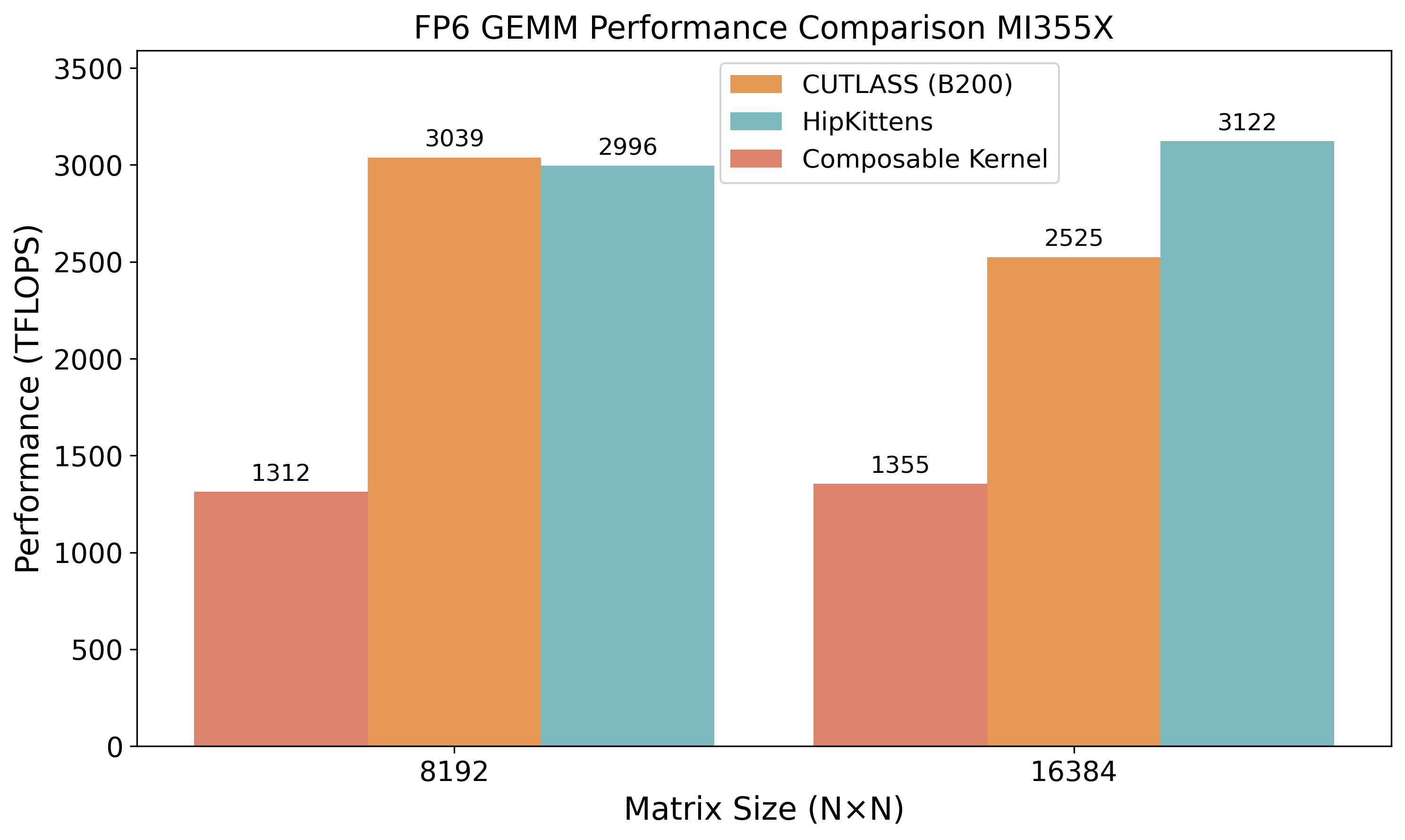}
\caption{\textbf{FP6 GEMM.} We compare performance of FP6 GEMM on square shapes 8192 and 16384 across AMD's CK library, NVIDIA's CUTLASS on B200, and {\name}. We use 500 iterations of warmup and 100 iterations of measurement.}
\end{figure}

%% file: main.bbl
\begin{thebibliography}{38}
\providecommand{\natexlab}[1]{#1}
\providecommand{\url}[1]{\texttt{#1}}
\expandafter\ifx\csname urlstyle\endcsname\relax
  \providecommand{\doi}[1]{doi: #1}\else
  \providecommand{\doi}{doi: \begingroup \urlstyle{rm}\Url}\fi

\bibitem[hel(2025)]{helion2025}
{Helion}, 2025.
\newblock URL \url{https://github.com/pytorch/helion}.

\bibitem[AI@Meta(2024)]{llama3modelcard}
AI@Meta.
\newblock Llama 3 model card.
\newblock 2024.
\newblock URL \url{https://github.com/meta-llama/llama3/blob/main/MODEL_CARD.md}.

\bibitem[{AMD}(2025{\natexlab{a}})]{amd2025aiter}
{AMD}.
\newblock {AITER}, 2025{\natexlab{a}}.
\newblock URL \url{https://github.com/ROCm/aiter}.

\bibitem[{AMD}(2025{\natexlab{b}})]{amd2025ck}
{AMD}.
\newblock {Composable Kernel}, 2025{\natexlab{b}}.
\newblock URL \url{https://github.com/ROCm/composable_kernel}.

\bibitem[AMD(2025)]{amd2025hardware}
AMD.
\newblock Rocm hardware features, 2025.
\newblock URL \url{https://rocm.docs.amd.com/projects/HIP/en/docs-develop/reference/hardware_features.html}.

\bibitem[{AMD}(2025{\natexlab{a}})]{amd2025matrix}
{AMD}.
\newblock {AMD Matrix Instruction Calculator}, 2025{\natexlab{a}}.
\newblock URL \url{https://github.com/ROCm/amd_matrix_instruction_calculator}.

\bibitem[{AMD}(2025{\natexlab{b}})]{amd2025mi355x}
{AMD}.
\newblock {AMD Instinct™ MI355X GPUs}, 2025{\natexlab{b}}.
\newblock URL \url{https://www.amd.com/en/products/accelerators/instinct/mi350/mi355x.html}.

\bibitem[AMD(2025)]{amd2025rocmlib}
AMD.
\newblock Rocm libraries, 2025.
\newblock URL \url{https://github.com/ROCm/rocm-libraries}.

\bibitem[Baronio et~al.(2025)Baronio, Marsella, Pan, Guo, and Alberti]{baronio2025kevinmultiturnrlgenerating}
Carlo Baronio, Pietro Marsella, Ben Pan, Simon Guo, and Silas Alberti.
\newblock Kevin: Multi-turn rl for generating cuda kernels, 2025.
\newblock URL \url{https://arxiv.org/abs/2507.11948}.

\bibitem[Bikshandi \& Shah(2023)Bikshandi and Shah]{bikshandi2023developing}
Ganesh Bikshandi and Jay Shah.
\newblock Developing cuda kernels for accelerated matrix multiplication on nvidia hopper architecture using the cutlass library, 2023.
\newblock URL \url{https://research.colfax-intl.com/wp-content/uploads/2023/12/colfax-gemm-kernels-hopper.pdf}.

\bibitem[Chen et~al.(2018)Chen, Moreau, Jiang, Zheng, Yan, Shen, Cowan, Wang, Hu, Ceze, et~al.]{tvm}
Tianqi Chen, Thierry Moreau, Ziheng Jiang, Lianmin Zheng, Eddie Yan, Haichen Shen, Meghan Cowan, Leyuan Wang, Yuwei Hu, Luis Ceze, et~al.
\newblock $\{$TVM$\}$: An automated $\{$End-to-End$\}$ optimizing compiler for deep learning.
\newblock In \emph{13th USENIX Symposium on Operating Systems Design and Implementation (OSDI 18)}, pp.\  578--594, 2018.

\bibitem[Devlin et~al.(2019)Devlin, Chang, Lee, and Toutanova]{devlin2020transformers}
Jacob Devlin, Ming-Wei Chang, Kenton Lee, and Kristina Toutanova.
\newblock Bert: Pre-training of deep bidirectional transformers for language understanding.
\newblock In \emph{Proceedings of NAACL-HLT 2019}, 2019.

\bibitem[Forum(2023)]{nvidia2023smemforum}
NVIDIA~Developer Forum.
\newblock How to understand the bank conflict of shared memory, 2023.
\newblock URL \url{https://forums.developer.nvidia.com/t/how-to-understand-the-bank-conflict-of-shared-mem/260900}.

\bibitem[Harris et~al.(2020)Harris, Millman, van~der Walt, Gommers, Virtanen, Cournapeau, Wieser, Taylor, Berg, Smith, Kern, Picus, Hoyer, van Kerkwijk, Brett, Haldane, del R{\'{i}}o, Wiebe, Peterson, G{\'{e}}rard-Marchant, Sheppard, Reddy, Weckesser, Abbasi, Gohlke, and Oliphant]{numpy}
Charles~R. Harris, K.~Jarrod Millman, St{\'{e}}fan~J. van~der Walt, Ralf Gommers, Pauli Virtanen, David Cournapeau, Eric Wieser, Julian Taylor, Sebastian Berg, Nathaniel~J. Smith, Robert Kern, Matti Picus, Stephan Hoyer, Marten~H. van Kerkwijk, Matthew Brett, Allan Haldane, Jaime~Fern{\'{a}}ndez del R{\'{i}}o, Mark Wiebe, Pearu Peterson, Pierre G{\'{e}}rard-Marchant, Kevin Sheppard, Tyler Reddy, Warren Weckesser, Hameer Abbasi, Christoph Gohlke, and Travis~E. Oliphant.
\newblock Array programming with {NumPy}.
\newblock \emph{Nature}, September 2020.

\bibitem[Kim et~al.(2022)Kim, Henry, Fahim, and Awadalla]{kim2022sayselephantscantrun}
Young~Jin Kim, Rawn Henry, Raffy Fahim, and Hany~Hassan Awadalla.
\newblock Who says elephants can't run: Bringing large scale moe models into cloud scale production.
\newblock \emph{Proceedings of the Third Workshop on Simple and Efficient Natural Language Processing (SustaiNLP)}, 2022.

\bibitem[Krizhevsky et~al.(2012)Krizhevsky, Sutskever, and Hinton]{krizhevsky2012imagenet}
Alex Krizhevsky, Ilya Sutskever, and Geoffrey~E. Hinton.
\newblock Imagenet classification with deep convolutional neural networks.
\newblock \emph{Advances in Neural Information Processing Systems 25 (NIPS 2012)}, 2012.

\bibitem[Lange et~al.(2025)Lange, Sun, Prasad, Faldor, Tang, and Ha]{lange2025robustagenticcudakernel}
Robert~Tjarko Lange, Qi~Sun, Aaditya Prasad, Maxence Faldor, Yujin Tang, and David Ha.
\newblock Towards robust agentic cuda kernel benchmarking, verification, and optimization, 2025.
\newblock URL \url{https://arxiv.org/abs/2509.14279}.

\bibitem[Lattner \& Adve(2004)Lattner and Adve]{lattner2020llvm}
Chris Lattner and Vikram Adve.
\newblock Llvm: A compilation framework for lifelong program analysis \& transformation, 2004.

\bibitem[Lattner et~al.(2020)Lattner, Amini, Bondhugula, Cohen, Davis, Pienaar, Riddle, Shpeisman, Vasilache, and Zinenko]{lattner2020mlir}
Chris Lattner, Mehdi Amini, Uday Bondhugula, Albert Cohen, Andy Davis, Jacques Pienaar, River Riddle, Tatiana Shpeisman, Nicolas Vasilache, and Oleksandr Zinenko.
\newblock Mlir: A compiler infrastructure for the end of moore's law, 2020.
\newblock URL \url{https://arxiv.org/abs/2002.11054}.

\bibitem[{LLVM Compiler Infrastructure}(2025)]{llvm2025user}
{LLVM Compiler Infrastructure}.
\newblock {User Guide for AMDGPU Backend}, 2025.
\newblock URL \url{https://llvm.org/docs/AMDGPUUsage.html}.

\bibitem[NVIDIA(2017)]{nvidia2017cutlass}
NVIDIA.
\newblock Cuda templates for linear algebra subroutines, 2017.
\newblock URL \url{https://github.com/NVIDIA/cutlass}.

\bibitem[{NVIDIA}(2024)]{nvidia2024cute}
{NVIDIA}.
\newblock {NVIDIA CuTe}, 2024.
\newblock URL \url{https://github.com/ NVIDIA/cutlass/blob/main/media/docs/cute/00_quickstart.md}.

\bibitem[NVIDIA(2025{\natexlab{a}})]{25-blackwell}
NVIDIA.
\newblock Nvidia blackwell architecture technical brief.
\newblock \url{https://resources.nvidia.com/en-us-blackwell-architecture}, 2025{\natexlab{a}}.

\bibitem[NVIDIA(2025{\natexlab{b}})]{nvidia2025cute4}
NVIDIA.
\newblock nvidia-cutlass-dsl, 2025{\natexlab{b}}.
\newblock URL \url{https://pypi.org/project/nvidia-cutlass-dsl/}.

\bibitem[{OpenAI}(2025)]{openai2025amd}
{OpenAI}.
\newblock {AMD and OpenAI announce strategic partnership to deploy 6 gigawatts of AMD GPUs}, 2025.
\newblock URL \url{https://openai.com/index/openai-amd-strategic-partnership/}.

\bibitem[OpenAI et~al.(2025)OpenAI, :, Agarwal, Ahmad, Ai, Altman, Applebaum, Arbus, Arora, Bai, Baker, Bao, Barak, Bennett, Bertao, Brett, Brevdo, Brockman, Bubeck, Chang, Chen, Chen, Cheung, Clark, Cook, Dukhan, Dvorak, Fives, Fomenko, Garipov, Georgiev, Glaese, Gogineni, Goucher, Gross, Guzman, Hallman, Hehir, Heidecke, Helyar, Hu, Huet, Huh, Jain, Johnson, Koch, Kofman, Kundel, Kwon, Kyrylov, Le, Leclerc, Lennon, Lessans, Lezcano-Casado, Li, Li, Lin, Liss, Lily, Liu, Liu, Lu, Lu, Martinovic, McCallum, McGrath, McKinney, McLaughlin, Mei, Mostovoy, Mu, Myles, Neitz, Nichol, Pachocki, Paino, Palmie, Pantuliano, Parascandolo, Park, Pathak, Paz, Peran, Pimenov, Pokrass, Proehl, Qiu, Raila, Raso, Ren, Richardson, Robinson, Rotsted, Salman, Sanjeev, Schwarzer, Sculley, Sikchi, Simon, Singhal, Song, Stuckey, Sun, Tillet, Toizer, Tsimpourlas, Vyas, Wallace, Wang, Wang, Watkins, Weil, Wendling, Whinnery, Whitney, Wong, Yang, Yang, Yasunaga, Ying, Zaremba, Zhan, Zhang, Zhang, Zhang, and
  Zhao]{openai2025gptoss120bgptoss20bmodel}
OpenAI, :, Sandhini Agarwal, Lama Ahmad, Jason Ai, Sam Altman, Andy Applebaum, Edwin Arbus, Rahul~K. Arora, Yu~Bai, Bowen Baker, Haiming Bao, Boaz Barak, Ally Bennett, Tyler Bertao, Nivedita Brett, Eugene Brevdo, Greg Brockman, Sebastien Bubeck, Che Chang, Kai Chen, Mark Chen, Enoch Cheung, Aidan Clark, Dan Cook, Marat Dukhan, Casey Dvorak, Kevin Fives, Vlad Fomenko, Timur Garipov, Kristian Georgiev, Mia Glaese, Tarun Gogineni, Adam Goucher, Lukas Gross, Katia~Gil Guzman, John Hallman, Jackie Hehir, Johannes Heidecke, Alec Helyar, Haitang Hu, Romain Huet, Jacob Huh, Saachi Jain, Zach Johnson, Chris Koch, Irina Kofman, Dominik Kundel, Jason Kwon, Volodymyr Kyrylov, Elaine~Ya Le, Guillaume Leclerc, James~Park Lennon, Scott Lessans, Mario Lezcano-Casado, Yuanzhi Li, Zhuohan Li, Ji~Lin, Jordan Liss, Lily, Liu, Jiancheng Liu, Kevin Lu, Chris Lu, Zoran Martinovic, Lindsay McCallum, Josh McGrath, Scott McKinney, Aidan McLaughlin, Song Mei, Steve Mostovoy, Tong Mu, Gideon Myles, Alexander Neitz, Alex Nichol, Jakub
  Pachocki, Alex Paino, Dana Palmie, Ashley Pantuliano, Giambattista Parascandolo, Jongsoo Park, Leher Pathak, Carolina Paz, Ludovic Peran, Dmitry Pimenov, Michelle Pokrass, Elizabeth Proehl, Huida Qiu, Gaby Raila, Filippo Raso, Hongyu Ren, Kimmy Richardson, David Robinson, Bob Rotsted, Hadi Salman, Suvansh Sanjeev, Max Schwarzer, D.~Sculley, Harshit Sikchi, Kendal Simon, Karan Singhal, Yang Song, Dane Stuckey, Zhiqing Sun, Philippe Tillet, Sam Toizer, Foivos Tsimpourlas, Nikhil Vyas, Eric Wallace, Xin Wang, Miles Wang, Olivia Watkins, Kevin Weil, Amy Wendling, Kevin Whinnery, Cedric Whitney, Hannah Wong, Lin Yang, Yu~Yang, Michihiro Yasunaga, Kristen Ying, Wojciech Zaremba, Wenting Zhan, Cyril Zhang, Brian Zhang, Eddie Zhang, and Shengjia Zhao.
\newblock gpt-oss-120b \& gpt-oss-20b model card, 2025.
\newblock URL \url{https://arxiv.org/abs/2508.10925}.

\bibitem[Ouyang et~al.(2025)Ouyang, Guo, Arora, L.~Zhang, Hu, Ré, and Mirhoseini]{ouyang2025kernelbench}
Anne Ouyang, Simon Guo, Simran Arora, Alex L.~Zhang, William Hu, Christopher Ré, and Azalia Mirhoseini.
\newblock Kernelbench: Can llms write efficient gpu kernels?
\newblock In \emph{International Conference on Machine Learning (ICML)}, 2025.
\newblock \doi{10.48550/arXiv.2502.10517}.
\newblock URL \url{https://arxiv.org/abs/2502.10517}.

\bibitem[Paszke et~al.(2019)Paszke, Gross, Massa, Lerer, Bradbury, Chanan, Killeen, Lin, Gimelshein, Antiga, Desmaison, Köpf, Yang, DeVito, Raison, Tejani, Chilamkurthy, Steiner, Fang, Bai, and Chintala]{paszke2019pytorchimperativestylehighperformance}
Adam Paszke, Sam Gross, Francisco Massa, Adam Lerer, James Bradbury, Gregory Chanan, Trevor Killeen, Zeming Lin, Natalia Gimelshein, Luca Antiga, Alban Desmaison, Andreas Köpf, Edward Yang, Zach DeVito, Martin Raison, Alykhan Tejani, Sasank Chilamkurthy, Benoit Steiner, Lu~Fang, Junjie Bai, and Soumith Chintala.
\newblock Pytorch: An imperative style, high-performance deep learning library, 2019.
\newblock URL \url{https://arxiv.org/abs/1912.01703}.

\bibitem[{Sara Hooker}(2021)]{hooker2020lottery}
{Sara Hooker}.
\newblock {The Hardware Lottery}.
\newblock \emph{Communications of the ACM}, 2021.

\bibitem[{SemiAnalysis}(2024)]{semianalysis2024moat}
{SemiAnalysis}.
\newblock {MI300X vs H100 vs H200 Benchmark Part 1: Training – CUDA Moat Still Alive}, 2024.
\newblock URL \url{https://semianalysis.com/2024/12/22/mi300x-vs-h100-vs-h200-benchmark-part-1-training/}.

\bibitem[Shah et~al.(2024)Shah, Bikshandi, Zhang, Thakkar, Ramani, and Dao]{dao2024flashattention3}
Jay Shah, Ganesh Bikshandi, Ying Zhang, Vijay Thakkar, Pradeep Ramani, and Tri Dao.
\newblock Flashattention-3: Fast and accurate attention with asynchrony and low-precision, 2024.
\newblock URL \url{https://arxiv.org/abs/2407.08608}.

\bibitem[Spector et~al.(2025)Spector, Juravsky, Sul, Dugan, Lim, Fu, Arora, and Ré]{spector2025look}
Benjamin Spector, Jordan Juravsky, Stuart Sul, Owen Dugan, Dylan Lim, Dan Fu, Simran Arora, and Christopher Ré.
\newblock Look ma, no bubbles! designing a low-latency megakernel for llama-1b, 2025.
\newblock URL \url{https://hazyresearch.stanford.edu/blog/2025-05-27-no-bubbles}.

\bibitem[Spector et~al.(2024)Spector, Arora, Singhal, Fu, and Ré]{spector2025tk}
Benjamin~F. Spector, Simran Arora, Aaryan Singhal, Daniel~Y. Fu, and Christopher Ré.
\newblock Thunderkittens: Simple, fast, and adorable ai kernels.
\newblock \emph{International Conference on Learning Representations (ICLR)}, 2024.

\bibitem[Tillet et~al.(2019)Tillet, Kung, and Cox]{triton}
Philippe Tillet, H.~T. Kung, and David Cox.
\newblock Triton: an intermediate language and compiler for tiled neural network computations.
\newblock In \emph{Proceedings of the 3rd ACM SIGPLAN International Workshop on Machine Learning and Programming Languages}, 2019.

\bibitem[{Triton}(2025)]{triton2025gluon}
{Triton}.
\newblock {Gluon?}, 2025.
\newblock URL \url{https://github.com/triton-lang/triton/issues/7392}.

\bibitem[Wang et~al.(2025)Wang, Cheng, Shi, Tang, Mo, Xie, Ma, Xia, Xue, Yang, and Yang]{wang2025tilelang}
Lei Wang, Yu~Cheng, Yining Shi, Zhengju Tang, Zhiwen Mo, Wenhao Xie, Lingxiao Ma, Yuqing Xia, Jilong Xue, Fan Yang, and Zhi Yang.
\newblock Tilelang: A composable tiled programming model for ai systems, 2025.
\newblock URL \url{https://arxiv.org/abs/2504.17577}.

\bibitem[Zhang et~al.(2025)Zhang, Zheng, Lin, Jiang, Bao, Jiang, Hou, Cui, Zheng, Chang, Chen, and Liu]{25-comet}
Shulai Zhang, Ningxin Zheng, Haibin Lin, Ziheng Jiang, Wenlei Bao, Chengquan Jiang, Qi~Hou, Weihao Cui, Size Zheng, Li-Wen Chang, Quan Chen, and Xin Liu.
\newblock {Comet: Fine-grained Computation-communication Overlapping for Mixture-of-Experts}.
\newblock \emph{Proceedings of the 8th MLSys Conference (MLSys)}, March 2025.

\bibitem[Zhou et~al.(2025)Zhou, Lezcano, Goucher, Rakhmati, Niu, Lebar, Szczerbuk, Bell, Tillet, Raoux, and Moudallal]{zhou2025linearlayoutsrobustcode}
Keren Zhou, Mario Lezcano, Adam Goucher, Akhmed Rakhmati, Jeff Niu, Justin Lebar, Pawel Szczerbuk, Peter Bell, Phil Tillet, Thomas Raoux, and Zahi Moudallal.
\newblock Linear layouts: Robust code generation of efficient tensor computation using $\mathbb{F}_2$.
\newblock \emph{Proceedings of the 31st ACM International Conference on Architectural Support for Programming Languages and Operating Systems, Volume (ASPLOS ’26)}, 2025.

\end{thebibliography}
